\documentclass{article}
\usepackage{iclr2027_conference,times}
\iclrfinalcopy
\usepackage{microtype}
\usepackage{url}
\usepackage{amsmath,mathtools,amsfonts,amssymb}
\usepackage{xcolor}
\usepackage{graphicx}
\usepackage{caption}
\usepackage{algorithm}
\usepackage{algpseudocode}
\usepackage{placeins}
\usepackage{float}
\usepackage{booktabs}
\usepackage{enumitem}
\usepackage{pifont}
\usepackage{latexsym}
\usepackage[T1]{fontenc}
\usepackage[utf8]{inputenc}
\usepackage{multirow}
\usepackage{array}
\usepackage{arydshln}
\usepackage{siunitx}
\usepackage{textcomp}
\usepackage{listings}
\usepackage{pgfplots}
\usepgfplotslibrary{groupplots}
\usetikzlibrary{calc,positioning,arrows.meta,fit,backgrounds,decorations.pathreplacing,shapes.symbols,shapes.misc}
\pgfplotsset{compat=1.18}
\usepackage{multicol}
\usepackage[varqu,varl,scaled=0.95]{zi4}
\usepackage{hyperref}
\hypersetup{
    colorlinks=true,       % colored text instead of boxes around links
    linkcolor=teal,        % internal links (sections, figures, equations)
    citecolor=teal,        % citation links
    urlcolor=magenta,      % web links (\url, \href)
    filecolor=teal,        % links to local files
    bookmarks=true,        % PDF sidebar outline
    bookmarksnumbered=true,% show section numbers in the outline
    bookmarksopen=true,    % outline expanded on opening
    pdfstartview=FitH      % open fitted to page width
}

\definecolor{gaincol}{HTML}{1A7F37}
\definecolor{losscol}{HTML}{C0392B}

\newcommand{\entoxx}{\textsc{EN$\rightarrow$XX}}
\newcommand{\xxtoen}{\textsc{XX$\rightarrow$EN}}
\newcommand{\llamai}{\mbox{LLaMA-3.2-3B-Instruct}}
\newcommand{\qweni}{\mbox{Qwen-2.5-3B-Instruct}}
\newcommand{\llama}{\mbox{LLaMA}}
\newcommand{\qwen}{\mbox{Qwen}}
\newcommand{\onesft}{\mbox{1K-SFT}}
\newcommand{\tensft}{\mbox{10K-SFT}}
\newcommand{\fiftysft}{\mbox{50K-SFT}}

\newcommand{\BTtechnique}{\mbox{BackTransLLM}}
\newcommand{\sftbt}{\mbox{\BTtechnique}}
\newcommand{\cptsft}{\mbox{CPT$\rightarrow$SFT}}
\newcommand{\cptrecast}{\mbox{CPT$\rightarrow$\BTtechnique}}
\newcommand{\xxl}{\mbox{X-XXL}}
\newcommand{\xl}{\mbox{X-XL}}
\newcommand{\chrf}{\mbox{ChrF++}}
\newcommand{\sbleu}{\mbox{sacreBLEU}}
\newcommand{\bert}{\mbox{BERTScore}}
\definecolor{gaingreen}{RGB}{0,115,45}
\definecolor{lossred}{RGB}{175,20,20}
\newcommand{\up}[1]{\textcolor{gaingreen}{+#1\%}}
\newcommand{\down}[1]{\textcolor{lossred}{$-$#1\%}}
\newcommand{\gaincell}[2]{\begin{tabular}[t]{@{}c@{}}#1\\[-0.25ex]#2\end{tabular}}

\title{Back-Translation Still Matters:\\Leveraging Monolingual Data for Low-Resource Machine Translation with LLMs}

\author{
Divyanshu Bathla\\
IIT Roorkee\\
\texttt{divyanshu\_b@ee.iitr.ac.in}
\And
Parag Singla\\
IIT Delhi\\
\texttt{parags@cse.iitd.ac.in}
\And
Jitin Singla$^\star$\\
IIT Roorkee\\
\texttt{jsingla@bt.iitr.ac.in}
}

\contribution{$^\star$Corresponding author. Authors acknowledge PARAM Pragya HPC facility IIT Delhi for computational resources used in this work. DB acknowledges the fellowship provided by the IndiaAI Fellowship under the IndiaAI Mission, Ministry of Electronics and Information Technology (MeitY), Government of India. PS acknowledges support from IBM AI Horizon Network. JS acknowledges the compute hardware support from POWERGRID (PCL-2596-ICC) and Dell Seed Program.}

\begin{document}

\maketitle

\begin{abstract}
Pre-trained Large Language Model (LLMs) are effective on many tasks in zero-shot and few-shot settings. For machine translation of low-resource languages, the obvious options are few-shot prompting and fine-tuning on a small parallel set. We ask: what if moderate amounts of low-resource monolingual data are also available? We argue that, rather than continued pretraining, a more effective use of this data is the pre-LLM era technique of back-translation (BT). The current fine-tuned LLM translates monolingual sentences in each language, and the resulting pairs train it in the reverse direction. Updates on the parallel set are interleaved for stable learning. With 1K parallel pairs on Odia and Swahili, BT beats continued pretraining into the low-resource language by 7.7 to 16.4 XCOMET-XXL points in domain. With beam-decoded intermediate translations, it also beats a model with 10$\times$ more parallel data. In-context examples retrieved from the parallel data further improve BT. We also analyze the source, quality and amount of monolingual text. Across five low-resource languages, with Hindi as a high-resource control, and two 3B backbones, BT improves translation into the low-resource language by 16 to 173\% XCOMET-XXL over parallel-only fine-tuning. Blind human evaluation on Odia and Sanskrit agrees with the automatic metrics.
\end{abstract}
% \end{document}
\section{Introduction}
\label{sec:intro}
LLMs are effective on many tasks in zero-shot and few-shot settings \citep{brown2020language}, and they translate well between high-resource languages \citep{hendy2023llmmt}. They are, however, pre-trained overwhelmingly on a handful of high-resource languages, while most of the world's languages appear only sparsely. Open-source instruction-tuned models are therefore fluent in English but have weak or near-zero competence in many low-resource languages. Gold-standard parallel corpora for low-resource pairs are also typically two to three orders of magnitude smaller than those for high-resource pairs, and they are often confined to narrow domains \citep{ramesh2022samanantar, gala2023indictrans2, costajussa2022nllb}.
 
The obvious ways to use such model for a low-resource language are few-shot prompting and supervised fine-tuning (SFT) on a parallel set. Few-shot prompting places a few translated examples in the prompt, and retrieving examples that are similar to the input helps further \citep{vilar2023prompting,zebaze2024incontext,zebaze2025compositional}. SFT on a small, high-quality parallel set adapts an LLM to translation \citep{xu2024paradigm}. Both depend on what the model already knows about the language. But what if we have a moderate amount of monolingual text available at our disposal. So, this paper asks: Can an LLM with a weak prior leverage monolingual text for better translation, and how should that text enter training? 

Two recipes have emerged for using monolingual text with LLMs. The first is continued pretraining (CPT) followed by SFT. ALMA \citep{xu2024paradigm} applies CPT on 7B \& 13B models with 1--20B monolingual tokens before fine-tuning on a small parallel set, and LLaMAX \citep{lu2024llamax} extends CPT, with added parallel data, to over 100 languages. ALMA's CPT optimizes a generic language-modeling objective, so cross-lingual mapping is learned only in SFT. Whether CPT will still pay off with only few million token is unclear. The second recipe is self-distillation. CycleDistill \citep{halder2025cycledistill} translates monolingual text with zero- or few-shot prompts, fine-tunes on its own translations, and repeats the cycle. Its training targets are therefore model-generated, and its authors report no gains if base model is weak.

Before LLMs, monolingual text entered machine translation (MT) mainly through back-translation (BT) \citep{sennrich2016improving}. A reverse model translates a human-written target-language sentence into the source language. The resulting pair has a synthetic source and a human target, and it is added to the training data. Later work sampled or noised the synthetic sources \citep{edunov2018understanding}, tagged them \citep{caswell2019tagged}, iterated the procedure \citep{hoang2018iterative}, and trained one bidirectional model for both directions \citep{niu2018bidirectional}. Dual learning and unsupervised MT learn from round trips over monolingual corpora in a similar way \citep{he2016dual,artetxe2018unsupervised,lample2018unsupervised}. These methods trained encoder--decoder models from scratch, first recurrent ones and later Transformers, and the BT studies used tens of thousands of parallel pairs or more. Their defining property is simple. The human-written sentence stays on the target side, so the model's own errors enter training only as input. Self-distillation does the opposite and trains the model on its own outputs.

We ask whether this property still matters for pretrained instruction-tuned LLMs. To test it, we port BT to such models as \BTtechnique{}. \BTtechnique{} interleaves BT updates on monolingual data with supervised updates on the small parallel set, which keep learning stable. In each BT update, the  model translates a monolingual sentence into the other language (a pivot) and then learns to reconstruct the original sentence from that pivot. A stop-gradient on the pivot means every BT gradient comes from predicting original monolingual text. We call this property anchoring. Like \citet{niu2018bidirectional}, we use one model for both directions, and like the semi-supervised variant of \citet{artetxe2018unsupervised}, we back-translate each batch on the fly with the current model and alternate these updates with supervised steps. While many of these ideas predate LLMs, our contribution is a controlled test of whether they still work for instruction-tuned LLMs with weak priors.

All tested recipes use the same backbone and same monolingual text. CPT and \BTtechnique{} also share the same 1K parallel pairs. We keep ALMA's full-weight CPT but run it on our monolingual pools, and we call the result ALMA-style. We follow CycleDistill's published recipe, with zero- or one-shot prompts and three cycles, and let our 3B model act as its own teacher. XCOMET-XXL \citep{xcomet} does not check the output language, so we report \chrf{} \citep{popovic2017chrfpp} alongside and flag outputs if in wrong language.
 
The results are consistent across Odia and Swahili. Self-distillation fails, which confirms that the prior is weak. After three cycles into Odia, 99\% of the model's outputs contain no Odia script, because training on its own outputs reinforces the errors of a weak prior. CPT helps on 15 of 16 measures, and into English it beats \BTtechnique{} with sampled pivots on XCOMET-XXL. Into the low-resource language, however, it trails \BTtechnique{} by 7.7 to 16.4 XCOMET-XXL points in domain. Pivot decoding also matters, as \citet{edunov2018understanding} found in resource-poor settings. With beam-decoded pivots, an option that only BT has, \BTtechnique{} beats CPT on 15 of 16 measures and a model trained on 10$\times$ more parallel data on all 16. In-context examples retrieved from the parallel data, which every recipe can use, improve both \cptsft{} and \BTtechnique{}. We also analyze the source, quality and amount of monolingual text. Across five low-resource languages and two 3B backbones, \BTtechnique{} improves translation into the low-resource language by 16--173\% XCOMET-XXL over the parallel data alone. Blind human evaluation on two languages agrees with the automatic metrics.  We will make the code publicly available on acceptance. Our primary contributions are:
\begin{itemize}
    \item We study back-translation (BT) as a way to use monolingual data when adapting general-purpose LLMs to low-resource machine translation. We compare it with continued pretraining, the standard LLM-era use of such data.
    \item Our approach builds on pre-LLM BT. The fine-tuned LLM translates monolingual sentences in each language, and the resulting pairs train it in the reverse direction. Updates on small parallel set are interleaved for stable learning.
    \item Experiments on five low-resource languages, Hindi as a high-resource control, and two 3B backbones show that BT improves translation into the low-resource language over parallel-only fine-tuning. On Odia and Swahili, it also beats continued pretraining. Blind human evaluation agrees. Ablations cover decoding of intermediate translations, retrieved in-context examples, and the source, quality and amount of monolingual text.
\end{itemize}

\section{Ways to Leverage Monolingual Data}
\label{sec:background}

\paragraph{Setting:}
We assume a small parallel set $\mathcal{P}=\{(e_i,x_i)\}_{i=1}^{N}$ and two \emph{monolingual} pools $\mathcal{M}_{\textsc{en}}$ and $\mathcal{M}_{\textsc{xx}}$ that are far larger than $\mathcal{P}$, for English (\textsc{en}) and a low-resource language (\textsc{xx}). The backbone $p_\theta$ is an instruction-tuned LLM that is fluent in \textsc{en} but has limited competence in \textsc{xx}. For direction $d\in\{e\!\rightarrow\!x,\,x\!\rightarrow\!e\}$, a template $T_d(\cdot)$ (Appendix~\ref{app:impl}) wraps a source sentence into an instruction prompt, and $p_\theta(\cdot\mid T_d(s))$ is the model's distribution over the target string. Every recipe fine-tunes on $\mathcal{P}$ with the supervised loss
\begin{equation}
\mathcal{L}^{d}_{\mathrm{sft}}(s,t) = -\log p_\theta\!\left(t \mid T_d(s)\right).
\end{equation}
The recipes differ only in how a monolingual sentence $m$ becomes a training example.
% (Table~\ref{tab:taxonomy}).
 
\paragraph{Continued pretraining, then SFT:} CPT treats $m$ as a language-modelling target, $\mathcal{L}_{\mathrm{cpt}}(m)=-\log p_\theta(m)$, and SFT on $\mathcal{P}$ follows.
ALMA \citep{xu2024paradigm} fine-tunes LLaMA-2 7B and 13B with full weights on up to 20B and 12B OSCAR tokens. It then fine-tunes on 58K human-written examples in ten directions.
The loss, either language modeling or SFT loss, falls on human-written text.
% CPT improves how the model reads and writes the language, how much it cannot teach the mapping between languages.
 
\paragraph{Self-distillation:} Self-distillation uses $m$ as the input and the model's own translation as the target:
\begin{equation}
\begin{split}
\hat{t} &\sim p_\theta(\cdot\mid T_d(m)),\\
\mathcal{L}_{\mathrm{sd}}(m) &= -\log p_\theta(\hat{t}\mid T_d(m)),
\end{split}
\end{equation}
repeated over cycles. CycleDistill \citep{halder2025cycledistill} labels Indic sentences with English translations from zero-, one- or four-shot prompts and fine-tunes on these labels for up to three cycles. It distils Gemma-2-9B \citep{gemmateam2024gemma2} and Llama-3.1-8B \citep{grattafiori2024llama3} into same-size and smaller students for Hindi, Bengali and Malayalam. Errors can compound across cycles, and the authors report no gains for Manipuri, which the base model could not translate. We therefore also use self-distillation as a control. If it fails, the model cannot yet label data for itself, which shows that our setting is truly low-resource.
 
\paragraph{Back-translation:}
BT uses $m$ as the target and the model's translation as the input:
\begin{equation}
\begin{split}
\hat{s} &\sim p_\theta(\cdot\mid T_{d^{-1}}(m)),\\
\mathcal{L}_{\mathrm{bt}}(m) &= -\log p_\theta(m\mid T_d(\hat{s})),
\end{split}
\end{equation}
where $d^{-1}$ is the reverse direction \citep{sennrich2016improving}. \citet{edunov2018understanding} found sampled or noised back-translations better than beam search at scale. In their resource-poor setting with 80K pairs, where back-translations are poor, beam search was better. \citet{hoang2018iterative} iterate BT so that forward and backward models regenerate each other's synthetic data. With recurrent models trained from scratch, their gains diminished after the first extra round. \citet{niu2018bidirectional} train one bidirectional model on Tagalog--English and Swahili--English with 24K to 51K parallel pairs, ``always placing the real data on the target side''. Removing target-side synthetic data improved three of their four low-resource directions. Self-distillation and BT therefore build the same kind of synthetic pair. They differ only in which side is human-written.

\paragraph{Other related work:} LLMs typically translate better out of low-resource languages than into them \citep{hendy2023llmmt,zhu2024multilingual,bawden2023investigating}, and \citet{bafna2025translation} localize a late-layer failure for low-resource targets. \citet{pang2024rethinking} analyze when monolingual data helps low-resource MT. These are analyses rather than recipes for our setting. \citet{vamvas2026translation} train an NLLB model on back-translations from a frozen Gemini model, a far stronger external teacher. Round-trip reinforcement learning tunes NLLB to reconstruct English without parallel data \citep{attia2026improving}. NLLB is already trained on large parallel corpora for translation, whereas we adapt a general-purpose LLM with a weak prior on the low-resource language. The reconstruction objective also gives no signal on real target-language text. Neither fits our setting, in which a single LLM learns from 1K parallel pairs and monolingual text alone.

\section{\BTtechnique{}: Back-Translation for Instruction-Tuned LLMs with Weak Priors}
\label{sec:method}
 
% ------------------------------------------------------------------
% Method figure, connected-cycle version (TikZ), compact layout.
% Preamble needs (after \usepackage{pgfplots} or \usepackage{tikz}):
%   \usetikzlibrary{positioning,arrows.meta,calc,fit,backgrounds,shapes.symbols,shapes.misc}
% Shapes: text = wavy box, model = rounded box, loss = cut-corner box, control = plain box.
% Usage:  \input{bt_figure}
% ------------------------------------------------------------------
\usetikzlibrary{shapes.symbols,shapes.misc} % wavy text boxes and cut-corner loss boxes
\providecolor{gaingreen}{RGB}{0,115,45}
\providecolor{lossred}{RGB}{175,20,20}
\begin{figure*}[t]
\centering
\resizebox{\textwidth}{!}{%
\begin{tikzpicture}[
  font=\small,
  >={Stealth[length=2mm]},
  txt/.style={shape=tape, tape bend top=in and out, tape bend bottom=in and out,
              tape bend height=1.1mm, align=center, text width=29mm,
              minimum height=8mm, inner sep=1.5pt},
  data/.style={txt, draw=gaingreen!75!black, fill=gaingreen!8},
  gen/.style={txt, draw=orange!85!black, fill=orange!12, dashed},
  model/.style={rounded corners=5pt, draw=blue!60!black, fill=blue!7, thick,
                align=center, text width=24mm, minimum height=8.5mm, inner sep=2pt},
  loss/.style={chamfered rectangle, chamfered rectangle xsep=1.8mm,
               draw=lossred!80!black, fill=lossred!6, align=center,
               text width=28mm, minimum height=8mm, inner sep=1.5pt},
  ctrl/.style={draw=black!60, fill=black!4, align=center, text width=27mm,
               minimum height=8mm, inner sep=2pt},
  flow/.style={->, thick, black!75},
  grad/.style={->, thick, densely dotted, lossred},
  lab/.style={font=\scriptsize, fill=white, inner sep=1.2pt},
  panel/.style={draw=black!45, dashed, rounded corners=5pt, inner sep=5pt},
  ptitle/.style={font=\small\bfseries, inner sep=1.5pt},
  % ---- icons ----
  flame/.pic={
    \fill[orange!95!red] (0,0) .. controls (-0.20,0.02) and (-0.24,0.22) .. (-0.06,0.44)
      .. controls (-0.04,0.30) and (0.04,0.30) .. (0.05,0.24)
      .. controls (0.12,0.30) and (0.14,0.36) .. (0.12,0.42)
      .. controls (0.26,0.26) and (0.22,0.02) .. (0,0) -- cycle;
    \fill[yellow!85!orange] (0,0.03) .. controls (-0.09,0.05) and (-0.10,0.16) .. (-0.02,0.24)
      .. controls (0.00,0.16) and (0.06,0.16) .. (0.07,0.12)
      .. controls (0.11,0.06) and (0.07,0.02) .. (0,0.03) -- cycle;},
  snow/.pic={
    \foreach \a in {90,150,...,450} {
      \draw[cyan!60!blue, line width=0.9pt] (0,0) -- (\a:0.22);
      \draw[cyan!60!blue, line width=0.7pt] (\a:0.13) -- ++(\a+40:0.07);
      \draw[cyan!60!blue, line width=0.7pt] (\a:0.13) -- ++(\a-40:0.07);}},
]

% ---------------- 1. Supervised step (top-left) ----------------
\node[data]  (par) at (0,0)    {Parallel pair\\ $(s,t)\sim\mathcal{P}$};
\node[model] (m1)  at (3.9,0)  {$p_\theta$, direction $d$\\[-1pt]\scriptsize input $T_d(s)$};
\node[loss]  (l1)  at (3.9,-1.55) {SFT loss\\ $-\log p_\theta(t\mid T_d(s))$};
\pic at ($(m1.north east)+(-0.25,-0.05)$) {flame};
\draw[flow] (par) -- (m1);
\draw[flow] (m1) -- node[right, font=\scriptsize]{predict $t$} (l1);
\draw[grad] (l1.west) to[bend left=40] node[left, font=\scriptsize]{$\nabla_\theta$} (m1.west);
\begin{scope}[on background layer]
  \node[panel, fit=(par)(m1)(l1)] (pa) {};
\end{scope}
\node[ptitle, anchor=south west] at (pa.north west) {Supervised step};

% ---------------- 2. Pivot generation (top-right) ----------------
\node[data]  (mono) at (8.8,0)  {Monolingual sentence\\ $m\sim\mathcal{M}_d$};
\node[model] (m2)   at (13.1,0) {$p_\theta$, direction $d^{-1}$\\[-1pt]\scriptsize input $T_{d^{-1}}(m)$};
\pic at ($(m2.north east)+(0.06,0.04)$) {snow};
\coordinate (snowc) at ($(m2.north east)+(0.3,0.28)$);
\draw[flow] (mono) -- (m2);
\begin{scope}[on background layer]
  \node[panel, fit=(mono)(m2)(snowc)] (pb) {};
\end{scope}
\node[ptitle, anchor=south west] at (pb.north west) {Generate pivot};

% supervised step -> pivot generation
\draw[flow] (pa.east |- m1) -- node[lab, above=1pt]{then} (pb.west |- mono);

% ---------------- pivot (between 2 and 3) ----------------
\node[gen] (piv) at (13.1,-1.55) {Pivot $\hat{s}$\\[-1pt] \scriptsize beam 5 or sampled};
\draw[flow] (m2) -- node[right, font=\scriptsize]{decode} (piv);

% ---------------- 3. Back-translation step (bottom-right) ----------------
\node[model] (m3) at (13.1,-3.1) {$p_\theta$, direction $d$\\[-1pt]\scriptsize input $T_d(\hat{s})$};
\pic at ($(m3.north east)+(-0.25,-0.05)$) {flame};
\draw[flow] (piv) -- (m3) node[pos=0.5, fill=white, draw=black!70, rounded corners=1pt,
       inner sep=1.2pt, font=\scriptsize\bfseries] {sg};
\node[loss] (l2) at (8.8,-3.1) {BT loss\\ $-\log p_\theta(m\mid T_d(\hat{s}))$};
\draw[flow] (m3) -- node[above, font=\scriptsize]{predict $m$} (l2);
\draw[grad] (l2.south east) to[bend right=22] node[below, font=\scriptsize, inner sep=1pt]{$\nabla_\theta$} (m3.south);
% the original monolingual sentence is the target
\draw[flow, gaingreen!70!black, very thick] (mono.south) --
  node[pos=0.5, lab, text=gaingreen!40!black]{target = original $m$} (l2.north);
\begin{scope}[on background layer]
  \coordinate (pcb) at ($(l2.south)+(0,-0.62)$);
  \node[panel, fit=(m3)(l2)(pcb)] (pc) {};
\end{scope}
\node[ptitle, anchor=south west] at ($(pc.south west)+(0.08,0.04)$) {Back-translation step};
% repeat r times
\draw[flow, black!60] (pc.east) -- ++(0.6,0) |-
  node[pos=0.25, right, font=\small\bfseries, text=black!75]{$\times r$} (pb.east);

% ---------------- 4. Switch direction, close the cycle ----------------
\node[ctrl] (sw) at (0,-3.1) {Switch direction\\ $d \leftarrow d^{-1}$};
\draw[flow] (pc.west |- l2) -- node[lab, align=center]{after $r$ BT steps} (sw);
\draw[flow] (sw) -- node[right, font=\scriptsize]{next round} (pa.south -| sw);

% ---------------- legend (one line) ----------------
\node[anchor=north west, font=\scriptsize, text=black!75, inner sep=0pt]
  at ($(pc.south -| sw.west)+(0,-0.22)$)
  {\tikz[baseline=0ex, scale=0.75]\pic{flame}; trainable (gradient to $\theta$)\quad
   \tikz[baseline=-0.55ex, scale=0.7]\pic{snow}; inference only\quad
   \tikz[baseline=-0.6ex]\node[data, text width=4mm, minimum height=3mm, tape bend height=0.6mm, inner sep=0pt]{}; text from the data\quad
   \tikz[baseline=-0.6ex]\node[gen, text width=4mm, minimum height=3mm, tape bend height=0.6mm, inner sep=0pt]{}; model-generated text\quad
   \tikz[baseline=-0.6ex]\node[model, text width=4mm, minimum height=3mm, rounded corners=2pt, inner sep=0pt]{}; model\quad
   \tikz[baseline=-0.6ex]\node[loss, text width=4mm, minimum height=3mm, chamfered rectangle xsep=0.8mm, inner sep=0pt]{}; loss\quad
   \tikz[baseline=-0.6ex]\node[draw=black!70, inner sep=1pt, font=\tiny\bfseries]{sg}; stop-gradient};

% ---------------- outer frame around the whole figure ----------------
\begin{scope}[on background layer]
  \node[draw=black!70, line width=0.6pt, rounded corners=4pt, inner sep=5pt,
        fit=(current bounding box)] {};
\end{scope}

\end{tikzpicture}%
}
\caption{BackTransLLM: One training cycle. After a supervised step in direction $d$, the current model translates a monolingual sentence $m$ in the reverse direction $d^{-1}$ (inference only) to give a pivot $\hat{s}$. Behind a stop-gradient (sg), the model learns to reconstruct the original $m$ from $\hat{s}$. Pivot generation and the back-translation step repeat $r{=}4$ times before the direction switches. All $p_\theta$ boxes share one set of QLoRA parameters $\theta$.}
\label{fig:method}
\end{figure*}

Our question is whether an LLM with a weak prior can leverage monolingual text for better translation, and how that text should enter training. The recipes in \S\ref{sec:background} differ in what the model learns to predict. CPT predicts original text but learns no mapping between languages. Self-distillation learns the mapping, but from its own translations, which are poor when the prior is weak. BT gives both, a translation signal and original text as every target. \BTtechnique{} keeps this anchoring in an instruction-tuned LLM.
 
\paragraph{Training cycle:} Figure~\ref{fig:method} shows one training cycle. Training alternates between the two directions $d\in\{x\!\rightarrow\!e,\,e\!\rightarrow\!x\}$. For each direction, \BTtechnique{} takes one supervised step on a batch of parallel pairs, then $r$ BT steps on batches of monolingual sentences, and then switches direction. The model generates the pivots on the fly with its current parameters, so the BT signal evolves during training. We use $r{=}4$, a robust choice among $r\in\{1,2,4,8\}$ (Appendix~\ref{app:ratio}). An epoch is defined over the monolingual pools, and the parallel data is cycled as needed.
 
\paragraph{Losses:} For a batch $B_{\mathcal{P}}\subset\mathcal{P}$ of parallel pairs and a batch $B_{\mathcal{M}}\subset\mathcal{M}_d$ from the monolingual pool in the target language of $d$, the supervised step and each BT step minimize
\begin{equation}
\begin{aligned}
\mathcal{L}^{d}_{\mathrm{sft}} &= -\tfrac{1}{|B_{\mathcal{P}}|}\textstyle\sum_{(s,t)\in B_{\mathcal{P}}}\log p_\theta(t \mid T_d(s)),\\
\mathcal{L}^{d}_{\mathrm{bt}} &= -\tfrac{1}{|B_{\mathcal{M}}|}\textstyle\sum_{m\in B_{\mathcal{M}}}\log p_\theta(m \mid T_d(\hat{s}_m))
\end{aligned}
\label{eq:steps}
\end{equation}
where $\hat{s}_m=\operatorname{sg}[\operatorname{dec}\,p_\theta(\cdot\mid T_{d^{-1}}(m))]$ is pivot, $\operatorname{sg}$ denotes stop-gradient and $\operatorname{dec}$ is pivot decoder. Per cycle: 1 step on $\mathcal{L}^{d}_{\mathrm{sft}}$ and $r$ on $\mathcal{L}^{d}_{\mathrm{bt}}$.
 
\paragraph{Pivot decoding:} The pivot is the only model-generated text in training, so its quality bounds what reconstruction can teach. Pivots for training into English are written in the low-resource language, which a weak-prior model generates poorly, so sampling adds noise exactly where the model is weakest. We compare sharpened nucleus sampling ($\tau{=}0.3$, top-$p{=}0.9$), which adds diversity \citep{edunov2018understanding}, with beam search over five beams, which \citet{edunov2018understanding} found better in resource-poor settings. Test-time decoding is greedy for every system.

\section{Empirical Results and Discussion}
\label{sec:setup}
 
We compare the recipes of \S\ref{sec:background} on Odia and Swahili with \llama{}, and then analyze \BTtechnique{} on six English-paired languages and two backbones. Our experiments address six questions. \textbf{1)}~Which recipe leverages monolingual data best (\S\ref{sec:recipes})? \textbf{2)}~Does in-context learning help (\S\ref{sec:icl})? \textbf{3)}~Which monolingual text should be used (\S\ref{sec:analysis-domain})? \textbf{4)}~How much monolingual text is needed (\S\ref{subsec:mono-scaling})? \textbf{5)}~Do the gains hold across languages and backbones (\S\ref{sec:crosslang})? \textbf{6)}~Do human judges agree with the automatic metrics (\S\ref{subsec:human_eval})?
 
\paragraph{Experimental setup:} For each language pair, the 1K parallel pairs and both monolingual pools come primarily from one parallel corpus. The corpora are VaqasAI for Odia \citep{ahmed2025odiaeng,haddow2020pmindia}, Samanantar for Assamese \citep{ramesh2022samanantar}, IITB En--Hi for Hindi \citep{kunchukuttan-etal-2018-iit}, Mitrasa\d{m}graha for Sanskrit \citep{nehrdich2026mitrasamgraha}, and MAFAND for both African pairs \citep{adelani-etal-2022-thousand}. CC-100 supplies Amharic monolingual data and augments Swahili \citep{conneau-etal-2020-unsupervised,wenzek-etal-2020-ccnet}. When both monolingual pools come from the same parallel source, they use disjoint subsets, and all pools are disjoint from validation and test data (Appendix~\ref{app:impl}). Hindi serves as a high-resource control. Our primary backbone is \llamai{} \citep{meta2024llama32}, and \qweni{} \citep{yang2024qwen} tests cross-family generality. Both are fine-tuned with 4-bit QLoRA \citep{dettmers2023qlora,lora} under identical settings (LoRA rank 64, peak learning rate $2\times10^{-4}$, maximum sequence length 2{,}048) until the development score converges. Prompts use a simple translation instruction wrapped in the model's chat template. We report XCOMET-XXL (\xxl) \citep{xcomet} and \chrf{} \citep{popovic2017chrfpp} on in-domain held-out test sets of 2K sentences and on Flores-200 \citep{costajussa2022nllb}. Checkpoints are selected on validation \bert{}, so neither primary metric is used for selection.
 
As baselines, \onesft{} fine-tunes on the 1K parallel pairs. \tensft{} uses ten times more pairs as a reference, except for Amharic, which has about 1.9K pairs. Few-shot prompts the base model, without training, with $k{=}5$ examples retrieved from the 1K pairs. \cptsft{} (ALMA-style) applies full-weight continued pretraining to the same 40K + 40K monolingual sentences and then fine-tunes on the parallel data. A variant uses only the first 100 parallel pairs. Self-distillation (CycleDistill-style) runs three cycles with sampled labels, either 0-shot (every prompt is zero-shot) or 1-shot (every prompt contains one gold example). CycleDistill translates into English, so translating into the low-resource language is our extension. Appendices~\ref{app:recipes} and~\ref{app:baselines} give the settings and full results.

\subsection{Back-Translation Uses Monolingual Data Best}
\label{sec:recipes}
 
Table~\ref{tab:recipes} compares three recipes on Odia and Swahili: \cptsft{}, self-distillation and back-translation. Across both languages, approaches that keep original monolingual text as the training target, either CPT (ALMA-style) or back-translation (\BTtechnique{}), are consistently effective. Self-distillation instead deteriorates and can collapse into the wrong output language, especially for Odia, where the base prior is weakest. After three cycles \emph{into Odia}, 99\% of self-distilled outputs contain no Odia script, and most are in English. As a result, \chrf{} drops to 0.54 (0-shot) and 0.53 (1-shot), below the zero-shot model's 7.18. XCOMET-XXL still assigns these outputs 29.60 to 42.27 because it does not check the output language, so we treat the score as invalid and do not report it. Even for Swahili, where the outputs are in the right language, self-distillation with or without a parallel example in the prompt ends below the zero-shot model on all eight measures. Self-distillation therefore cannot use monolingual data when the prior is weak, which confirms that our setting is truly low-resource. We drop it from remaining experiments.
 
\cptsft{} is a strong alternative. It improves on \onesft{} on all eight Odia measures and seven of eight Swahili measures, with a gain of 16.62 \xxl{} points into Odia in domain (Table~\ref{tab:recipes}). Into English, it even exceeds \tensft{} on Flores-200 (51.14 against 47.07 for Odia and 63.90 against 58.46 for Swahili). Even with only 100 parallel pairs, \cptsft{} still beats \onesft{} on all eight Odia measures.
 
With sampled pivots, \BTtechnique{} and \cptsft{} split the two directions. \BTtechnique{} leads \emph{into} Odia and Swahili on both test sets, by 7.67 and 16.39 \xxl{} points in domain and by 1.82 and 14.80 on Flores-200. \cptsft{} leads into English on \xxl{}, by 2.39 and 4.51 points in domain and by 12.27 and 7.75 on Flores-200. \BTtechnique{} without parallel seed stays far below \onesft{}, likely because its pivots come from a model that barely translates. With only 100 pairs, it already gets most of the way to the 1K-pair result (Table~\ref{tab:recipes}).
 
Only back-translation can choose how its pivots are decoded, since \cptsft{} generates none. With beam-decoded pivots, \BTtechnique{} beats \tensft{} on all 16 measures, by 1.53 to 8.29 \xxl{} points on Odia and 2.45 to 11.28 on Swahili. It also beats \cptsft{} on 15 of the 16, by 0.87 to 20.62 \xxl{} points. The exception is Swahili into English on Flores-200, where \cptsft{} leads by 0.50 (63.90 against 63.40). Over sampled pivots, beam pivots gain 5.35 to 13.14 \xxl{} points on Odia and 3.64 to 7.25 on Swahili. In both languages the gain is largest into English on Flores-200 (38.87 to 52.01 and 56.15 to 63.40), exactly where \cptsft{} led by the widest margin. \BTtechnique{} thus uses monolingual data better than CPT into the low-resource language with either decoder, and with beam pivots it also leads into English.

\begin{table*}[t]
\centering
\footnotesize
\setlength{\tabcolsep}{3.8pt}
\resizebox{\textwidth}{!}{%
\begin{tabular}{l c c cccc cccc}
\toprule
 & & & \multicolumn{4}{c}{\textbf{In-Domain (2K)}} & \multicolumn{4}{c}{\textbf{Flores-200}} \\
\cmidrule(lr){4-7}\cmidrule(l){8-11}
 & & & \multicolumn{2}{c}{\entoxx} & \multicolumn{2}{c}{\xxtoen} & \multicolumn{2}{c}{\entoxx} & \multicolumn{2}{c}{\xxtoen} \\
\cmidrule(lr){4-5}\cmidrule(lr){6-7}\cmidrule(lr){8-9}\cmidrule(l){10-11}
\textbf{System} & \textbf{\#Pairs} & \textbf{Pivots} & \xxl & \chrf & \xxl & \chrf & \xxl & \chrf & \xxl & \chrf \\
\midrule
\multicolumn{11}{@{}l}{\textbf{ODIA}} \\
Zero-shot & -- & -- & 20.09 & 7.18 & 16.80 & 17.78 & 20.81 & 4.95 & 16.96 & 19.41 \\

Few-shot ($k{=}5$) & -- & -- & 19.03 & 13.98 & 21.10 & 25.56 & 15.29 & 8.21 & 17.21 & 20.64 \\

\onesft & -- & -- & 42.05 & 25.13 & 45.24 & 46.58 & 30.60 & 18.03 & 40.94 & 37.41 \\
\tensft{} & -- & -- & 64.33 & 40.91 & \underline{61.51} & \underline{59.70} & 37.87 & 21.73 & 47.07 & 39.26 \\
\addlinespace[1.2pt]
\cptsft{} & 100 & -- & 47.71 & 28.50 & 52.13 & 48.88 & 33.55 & 19.06 & 48.27 & 38.98 \\
\cptsft{} & 1K & -- & 58.67 & 34.17 & 57.33 & 54.59 & 37.22 & 20.43 & \underline{51.14} & \underline{41.40} \\
Self-distill (0-shot) & -- & sampled & n/a\textsuperscript{a} & 0.54 & 15.19 & 14.21 & n/a\textsuperscript{a} & 1.08 & 15.27 & 14.21 \\
Self-distill (1-shot) & -- & sampled & n/a\textsuperscript{a} & 0.53 & 14.11 & 14.41 & n/a\textsuperscript{a} & 1.07 & 14.18 & 15.32 \\
\addlinespace[1.2pt]
\BTtechnique{} & 0 & sampled & 23.10 & 19.39 & 22.37 & 27.13 & 20.51 & 13.60 & 18.61 & 21.41 \\

\BTtechnique{} & 100 & sampled & 57.66 & 36.23 & 46.50 & 48.67 & 33.95 & 20.66 & 29.61 & 29.87 \\
\BTtechnique{} & 1K & sampled & \underline{66.34} & \underline{41.32} & 54.94 & 55.19 & \underline{39.04} & \underline{22.25} & 38.87 & 34.92 \\
\BTtechnique{} & 1K & beam 5 & \textbf{72.62} & \textbf{45.36} & \textbf{63.04} & \textbf{61.30} & \textbf{44.39} & \textbf{24.79} & \textbf{52.01} & \textbf{43.10} \\
\midrule
\multicolumn{11}{@{}l}{\textbf{SWAHILI}} \\
Zero-shot & -- & -- & 38.83 & 36.77 & 57.56 & 44.26 & 35.05 & 38.86 & 59.20 & 48.51 \\
Few-shot ($k{=}5$) & -- & -- & 40.27 & 38.02 & 61.35 & 46.35 & 35.29 & 39.37 & 60.54 & \underline{48.68} \\

\onesft & -- & -- & 42.12 & 39.10 & 60.05 & 46.13 & 35.41 & 37.97 & 56.90 & 46.06 \\
\tensft{} & -- & -- & 57.38 & 45.27 & 66.11 & \underline{49.13} & 45.54 & 40.56 & 58.46 & 46.24 \\
\addlinespace[1.2pt]
\cptsft{} & 1K & -- & 46.53 & 39.27 & \underline{67.16} & 49.07 & 38.38 & 36.95 & \textbf{63.90} & 48.66 \\

Self-distill (0-shot) & -- & sampled & 34.39 & 32.64 & 42.03 & 38.32 & 32.44 & 35.09 & 44.44 & 41.64 \\
Self-distill (1-shot) & -- & sampled & 33.14 & 32.89 & 40.61 & 38.19 & 32.07 & 35.98 & 42.17 & 41.44 \\
\addlinespace[1.2pt]
\BTtechnique{} & 1K & sampled & \underline{62.92} & \underline{47.17} & 62.65 & 48.68 & \underline{53.18} & \underline{43.17} & 56.15 & 47.36 \\
\BTtechnique{} & 1K & beam 5 & \textbf{67.15} & \textbf{49.13} & \textbf{68.56} & \textbf{51.07} & \textbf{56.82} & \textbf{45.79} & \underline{63.40} & \textbf{50.64} \\
\bottomrule
\end{tabular}%
}
\caption{Leveraging monolingual data for low-resource MT with \llamai{}. All recipes use 40K English and 40K XX monolingual sentences. \emph{\#Pairs} represent number of parallel pairs used. Self-distillation uses at most one parallel example per prompt and runs three cycles (best checkpoint reported). \emph{Few-shot} prompts the base model with $k{=}5$ examples retrieved from the 1K pairs (\S\ref{sec:icl}). \emph{Pivots}: decoding of the intermediate translations (sampled: $\tau{=}0.3$, top-$p{=}0.9$). \tensft{} uses 10$\times$ more parallel data. \textbf{Bold}/\underline{underline}: best/second-best (higher is better). \textsuperscript{a}Invalid: 99\% of outputs contain no Odia script (87--90\% are English, including 22\% copies of the source), which XCOMET misses.All five metrics, and \qwen{} results for zero-shot, \onesft{}, \tensft{}, \cptsft{} and \BTtechnique{} (sampled pivots), are in Appendix~\ref{app:baselines} and training settings are in Appendix~\ref{app:recipes}.}

\label{tab:recipes}
\end{table*}

\subsection{In-Context Learning Helps Every Recipe}
\label{sec:icl}
 
Beam-decoded pivots are not available to \cptsft{}. However, LLMs offer a second lever that both recipes can use. Examples retrieved from the parallel data can be placed in the prompt, and the model conditions on them without any change to its architecture \citep{brown2020language,vilar2023prompting}. Pre-LLM encoder--decoder systems needed a dedicated memory \citep{gu2018search} or training on inputs augmented with retrieved translations \citep{bulte2019neural}. We prepend $k{=}5$ examples, retrieved from the 1K parallel pairs by source-side similarity, to every training and test prompt, including \BTtechnique{}'s back-translation updates.
 
Table~\ref{tab:icl} compares \onesft{}, \cptsft{} and \BTtechnique{} with and without in-context learning (ICL) on Odia (\llama{}). \onesft{} with ICL outperforms \onesft{} under the same 1K parallel budget and with no additional annotations ($+16.7\%$ \xxl{} into Odia and $+9.2\%$ into English). Without fine-tuning, the same examples help little. Few-shot prompting of the base model stays near its zero-shot quality into Odia (19.03 against 20.09 \xxl{}, Table~\ref{tab:recipes}). ICL also helps both recipes that use monolingual data. \cptsft{} gains 1.61 and 1.29 \xxl{} points, and \BTtechnique{} with sampled pivots gains 3.29 and 5.78. With ICL, \BTtechnique{} leads \cptsft{} on all four measures, by 9.35 \xxl{} points into Odia and 2.10 into English. ICL thus keeps the ranking into Odia and lets \BTtechnique{} overtake \cptsft{} into English. Into English, \BTtechnique{} with ICL nearly matches \tensft{} (60.72 against 61.51 \xxl{}, Table~\ref{tab:recipes}) and exceeds it on \chrf{} (59.87 against 59.70). Amharic shows the same pattern for \onesft{} and \BTtechnique{} (Appendix~\ref{app:icl}).
 
Beam-decoded pivots remain the larger lever. \BTtechnique{} with beam pivots and no ICL beats \cptsft{} with ICL on all four measures, by 4.42 to 12.34 \xxl{} points, and it beats \BTtechnique{} with sampled pivots and ICL (72.62 and 63.04 \xxl{}, Table~\ref{tab:icl}). With both beam pivots and ICL, \BTtechnique{} is best on all four measures, by 3.19 \xxl{} points into Odia and 2.33 into English over beam pivots alone (75.81 and 65.37 \xxl{}). On Flores-200, ICL adds 4.73 \xxl{} points into Odia, and into English it stays 0.56 points below beam pivots alone (Appendix~\ref{app:icl}). \BTtechnique{} leads \cptsft{} into the low-resource language with or without ICL, and only back-translation can add beam-decoded pivots on top.

\begin{table}[t]
\centering
\footnotesize
\setlength{\tabcolsep}{3.5pt}
\begin{tabular}{@{}l c c cc cc@{}}
\toprule
 & & & \multicolumn{2}{c}{\textsc{EN$\rightarrow$OR}} & \multicolumn{2}{c}{\textsc{OR$\rightarrow$EN}} \\
\cmidrule(lr){4-5}\cmidrule(l){6-7}
\textbf{System} & \textbf{Pivots} & \textbf{ICL} & \xxl & \chrf & \xxl & \chrf \\
\midrule
\onesft & -- & No  & 42.05 & 25.13 & 45.24 & 46.58 \\
\onesft & -- & Yes & 49.09 & 29.18 & 49.40 & 49.40 \\
\midrule
\cptsft & -- & No  & 58.67 & 34.17 & 57.33 & 54.59 \\
\cptsft & -- & Yes & 60.28 & 35.03 & 58.62 & 55.19 \\
\midrule
\BTtechnique{} & sampled & No  & 66.34 & 41.32 & 54.94 & 55.19 \\
\BTtechnique{} & sampled & Yes & 69.63 & 43.76 & 60.72 & 59.87 \\
\cdashline{1-7}
\BTtechnique{} & beam 5 & No  & \underline{72.62} & \underline{45.36} & \underline{63.04} & \underline{61.30} \\
\BTtechnique{} & beam 5 & Yes & \textbf{75.81} & \textbf{46.65} & \textbf{65.37} & \textbf{62.33} \\
\bottomrule
\end{tabular}
\caption{ICL on Odia (\llama{}, held-out 2K), with $k{=}5$ semantically similar examples (Appendix~\ref{app:impl}) retrieved from the 1K parallel pairs in training and test prompt. \emph{Pivots}: decoding of the intermediate translations (sampled: $\tau{=}0.3$, top-$p{=}0.9$; beam 5: beam search with five beams), -- for systems without back-translation. \textbf{Bold}/\underline{underline}: best/second-best (higher is better). All five metrics, Flores-200 and Amharic are in Appendix~\ref{app:icl}.}
\label{tab:icl}
\end{table}

\subsection{Monolingual Text Quality Matters as Much as Domain}
\label{sec:analysis-domain}
 
We now analyze \BTtechnique{} further. These analyses use sampled pivots, which are cheaper than beam pivots, so they likely understate it.
 
\paragraph{Monolingual source:} Low-resource languages rarely have curated monolingual corpora, so practitioners rely on web text or other domains. For Odia and Assamese, we replace the in-corpus monolingual pools with a different-domain corpus, OdiEnCorp-2.0 \citep{parida-etal-2020-odiencorp}, and with web text from CC-100, IndicCorp-2.0 \citep{doddapaneni-etal-2023-towards} and WikiMatrix \citep{schwenk-etal-2021-wikimatrix} (Table~\ref{tab:bt-source}, top). Into the low-resource language, every source helps. Odia \xxl{} rises from 42.05 to between 51.82 and 66.34, and web text recovers most of the in-corpus gain (62.21 with IndicCorp-2.0 and CC-100). Assamese shows the same trend with smaller web gains. Into English, web text is less reliable and can fall below the supervised baseline (CC-100 for Odia: 40.96 against 45.24).
 
\paragraph{Quality versus domain:} To separate corpus quality from domain match, we cross the parallel corpus with the BT source over VaqasAI and OdiEnCorp-2.0 and test on both held-out sets (Table~\ref{tab:bt-source}, bottom). With the parallel data fixed, VaqasAI is the better BT source in all eight settings. This includes the OdiEnCorp-2.0 test set itself, where the margins are small (0.6 to 2.5 \xxl{} points). Domain match alone does not explain this, which suggests that corpus quality matters at least as much as domain.

\begin{table}[t]
\centering
\footnotesize
\setlength{\tabcolsep}{2.5pt}
\begin{tabular}{@{}l cc cc@{}}
\toprule
\multicolumn{5}{c}{\textbf{\textit{Monolingual text source}}} \\
\midrule
 & \multicolumn{2}{c}{\entoxx} & \multicolumn{2}{c}{\xxtoen} \\
\cmidrule(lr){2-3}\cmidrule(l){4-5}
\textbf{BT source (\textsc{xx} / \textsc{en})} & \xxl & \chrf & \xxl & \chrf \\
\midrule
\multicolumn{5}{@{}l}{\textbf{ODIA} \emph{(Parallel data: VaqasAI)}} \\
\emph{\onesft{} (no BT)} & \emph{42.05} & \emph{25.13} & \emph{45.24} & \emph{46.58} \\
VaqasAI & \textbf{66.34} & \textbf{41.32} & \textbf{54.94} & \textbf{55.19} \\
OdiEnCorp & 56.38 & 33.20 & 46.04 & 46.63 \\
IndicCorp / CC-100 & \underline{62.21} & \underline{37.83} & 46.30 & 45.21 \\
CC-100 & 53.44 & 34.06 & 40.96 & 42.59 \\
IndicCorp / WikiMatrix & 51.82 & 32.64 & \underline{47.70} & \underline{46.84} \\
\midrule
\multicolumn{5}{@{}l}{\textbf{ASSAMESE} \emph{(Parallel data: Samanantar)}} \\
\emph{\onesft{} (no BT)} & \emph{36.59} & \emph{18.99} & \emph{36.49} & \emph{27.84} \\
Samanantar & \textbf{46.41} & \textbf{27.02} & \textbf{39.69} & \textbf{33.54} \\
CC-100 & \underline{38.61} & \underline{21.73} & \underline{36.63} & \underline{28.29} \\
IndicCorp / CC-100 & 38.11 & 21.24 & 35.48 & 27.95 \\
\midrule
\multicolumn{5}{c}{\textbf{\textit{Monolingual text quality} (Odia, \xxl{})}} \\
\midrule
 & \multicolumn{2}{c}{\entoxx} & \multicolumn{2}{c}{\xxtoen} \\
\cmidrule(lr){2-3}\cmidrule(l){4-5}
\textbf{BT source} & {\scriptsize VaqasAI} & {\scriptsize OEC} & {\scriptsize VaqasAI} & {\scriptsize OEC} \\
\midrule
\multicolumn{5}{@{}l}{\emph{Parallel data: VaqasAI}} \\
VaqasAI & \textbf{66.34} & \textbf{30.37} & \textbf{54.94} & \textbf{30.94} \\
OdiEnCorp & 56.38 & 27.88 & 46.04 & 30.28 \\
\multicolumn{5}{@{}l}{\emph{Parallel data: OdiEnCorp}} \\
VaqasAI & \textbf{56.32} & \textbf{30.19} & \textbf{44.94} & \textbf{28.71} \\
OdiEnCorp & 38.82 & 29.47 & 25.64 & 28.07 \\
\bottomrule
\end{tabular}
\caption{\BTtechnique{} with different monolingual sources (\llama{}, sampled pivots, 40K sentences/language). \emph{Source}: 1K parallel pairs fixed. VaqasAI and Samanantar are in-domain, OdiEnCorp is out-of-domain, and the rest are web text. \emph{Quality}: each Odia parallel corpus crossed with each BT source and scored on both test sets (OEC: OdiEnCorp). \textbf{Bold}/\underline{underline}: best/second-best. All five metrics are in Appendix~\ref{app:datasrc} (source) and Appendix~\ref{app:xdomain} (quality, which adds Flores-200 and \cptsft{} trained on each corpus).}
\label{tab:bt-source}
\end{table}

\begin{table*}[t]
\centering
\tiny
\setlength{\tabcolsep}{3.0pt}
\renewcommand{\arraystretch}{1.1}
\resizebox{\textwidth}{!}{%
\begin{tabular}{@{}c c l *{12}{S[table-format=2.2]}@{}}
\toprule
& &
& \multicolumn{2}{c}{Odia} & \multicolumn{2}{c}{Assamese}
& \multicolumn{2}{c}{Sanskrit} & \multicolumn{2}{c}{Hindi}
& \multicolumn{2}{c}{Amharic} & \multicolumn{2}{c}{Swahili} \\
\cmidrule(lr){4-5}\cmidrule(lr){6-7}\cmidrule(lr){8-9}\cmidrule(lr){10-11}\cmidrule(lr){12-13}\cmidrule(lr){14-15}
& & \multicolumn{1}{l}{Method}
& {X-XXL} & {\chrf} & {X-XXL} & {\chrf} & {X-XXL} & {\chrf}
& {X-XXL} & {\chrf} & {X-XXL} & {\chrf} & {X-XXL} & {\chrf} \\
\midrule
\multirow{8}{*}{\rotatebox[origin=c]{90}{\textbf{LLaMA-3.2-3B}}}
& \multirow{4}{*}{\rotatebox[origin=c]{90}{\textit{\entoxx}}}
& Zero-Shot & {20.09} & {7.18} & {38.81} & {11.61} & {21.72} & {10.91} & {56.65} & {31.85} & {13.03} & {3.08} & {38.83} & {36.77} \\
& & \onesft   & {42.05} & {25.13} & {36.59} & {18.99} & {24.21} & {14.57} & {63.44} & {32.22} & {14.71} & {6.26} & {42.12} & {39.10} \\
& & \sftbt    & {\gaincell{\textbf{66.34}}{\up{58}}} & {\gaincell{\textbf{41.32}}{\up{64}}} & {\gaincell{\textbf{46.41}}{\up{27}}} & {\gaincell{\textbf{27.02}}{\up{42}}} & {\gaincell{\textbf{28.07}}{\up{16}}} & {\gaincell{\textbf{18.09}}{\up{24}}} & {\gaincell{\textbf{65.27}}{\up{3}}} & {\gaincell{\textbf{35.89}}{\up{11}}} & {\gaincell{\textbf{19.34}}{\up{31}}} & {\gaincell{\textbf{10.27}}{\up{64}}} & {\gaincell{\textbf{62.92}}{\up{49}}} & {\gaincell{\textbf{47.17}}{\up{21}}} \\[0.5ex]
\cdashline{3-15}
& \multirow{4}{*}{\rotatebox[origin=c]{90}{\textit{\xxtoen}}}
& Zero-Shot & {16.80} & {17.78} & {31.34} & {23.58} & {22.07} & {18.52} & {64.45} & {38.74} & {17.81} & {14.73} & {57.56} & {44.26} \\
& & \onesft   & {45.24} & {46.58} & {36.49} & {27.84} & {23.60} & {20.72} & {67.78} & {41.65} & {19.37} & {21.92} & {60.05} & {46.13} \\
& & \sftbt    & {\gaincell{\textbf{54.94}}{\up{21}}} & {\gaincell{\textbf{55.19}}{\up{18}}} & {\gaincell{\textbf{39.69}}{\up{9}}} & {\gaincell{\textbf{33.54}}{\up{20}}} & {\gaincell{\textbf{25.67}}{\up{9}}} & {\gaincell{\textbf{23.33}}{\up{13}}} & {\gaincell{\textbf{69.56}}{\up{3}}} & {\gaincell{\textbf{43.97}}{\up{6}}} & {\gaincell{\textbf{20.18}}{\up{4}}} & {\gaincell{\textbf{23.57}}{\up{8}}} & {\gaincell{\textbf{62.65}}{\up{4}}} & {\gaincell{\textbf{48.68}}{\up{6}}} \\[0.5ex]
\midrule
\multirow{8}{*}{\rotatebox[origin=c]{90}{\textbf{Qwen-2.5-3B}}}
& \multirow{4}{*}{\rotatebox[origin=c]{90}{\textit{\entoxx}}}
& Zero-Shot & {16.96} & {7.05} & {24.04} & {9.62} & {20.60} & {11.82} & {23.89} & {21.20} & {15.16} & {2.68} & {17.97} & {20.44} \\
& & \onesft   & {25.55} & {21.20} & {27.14} & {14.75} & {19.44} & {11.41} & {27.27} & {21.48} & {13.70} & {7.17} & {21.14} & {26.99} \\
& & \sftbt    & {\gaincell{\textbf{69.73}}{\up{173}}} & {\gaincell{\textbf{46.51}}{\up{119}}} & {\gaincell{\textbf{41.58}}{\up{53}}} & {\gaincell{\textbf{24.12}}{\up{64}}} & {\gaincell{\textbf{25.91}}{\up{33}}} & {\gaincell{\textbf{16.84}}{\up{48}}} & {\gaincell{\textbf{46.79}}{\up{72}}} & {\gaincell{\textbf{30.82}}{\up{43}}} & {\gaincell{\textbf{18.90}}{\up{38}}} & {\gaincell{\textbf{10.86}}{\up{51}}} & {\gaincell{\textbf{46.62}}{\up{121}}} & {\gaincell{\textbf{43.16}}{\up{60}}} \\[0.5ex]
\cdashline{3-15}
& \multirow{4}{*}{\rotatebox[origin=c]{90}{\textit{\xxtoen}}}
& Zero-Shot & {22.70} & {28.33} & {31.08} & {24.06} & {24.33} & {19.56} & {\textbf{59.35}} & {37.08} & {17.75} & {18.97} & {23.87} & {28.61} \\
& & \onesft   & {41.05} & {43.31} & {35.43} & {26.87} & {\textbf{25.80}} & {\textbf{21.88}} & {56.00} & {35.95} & {19.75} & {22.64} & {32.94} & {34.77} \\
& & \sftbt    & {\gaincell{\textbf{51.72}}{\up{26}}} & {\gaincell{\textbf{54.86}}{\up{27}}} & {\gaincell{\textbf{36.87}}{\up{4}}} & {\gaincell{\textbf{31.27}}{\up{16}}} & {\gaincell{23.77}{\down{8}}} & {\gaincell{21.74}{\down{1}}} & {\gaincell{56.48}{\up{1}}} & {\gaincell{\textbf{38.98}}{\up{8}}} & {\gaincell{\textbf{20.17}}{\up{2}}} & {\gaincell{\textbf{23.36}}{\up{3}}} & {\gaincell{\textbf{52.14}}{\up{58}}} & {\gaincell{\textbf{44.66}}{\up{28}}} \\[0.5ex]
\bottomrule
\end{tabular}%
}
\caption{Held-out results for six English-paired languages and two Instruct backbones (1K parallel pairs, 40K monolingual sentences, sampled pivots). Under each \BTtechnique{} score, \textcolor{gaingreen}{green}/\textcolor{lossred}{red} is the relative gain/loss over \onesft{}. All five metrics, with \tensft{} included, are in Appendix~\ref{app:heldout}. Flores-200 results are in Appendix~\ref{app:flores}.}
\label{tab:main-crosslang}
\end{table*}

\subsection{More Monolingual Text Helps, with Diminishing Returns}
\label{subsec:mono-scaling}
 
Holding the parallel data at 1K pairs, we vary only the monolingual budget (20K, 40K and 80K sentences per language), so any change in quality comes from the added monolingual text. Appendix~\ref{app:scaling} gives the full grid for Odia. Into Odia, \xxl{} rises from 42.05 with no monolingual data to 61.0 at 20K, 66.3 at 40K and 67.9 at 80K, and \chrf{} rises in step from 25.1 to 42.8. Into English, \xxl{} rises from 45.2 to 56.2 and \chrf{} from 46.6 to 56.7. Going from 40K to 80K adds only 1.6 and 1.2 \xxl{} points, so we use 40K as the default. The pattern holds at other parallel budgets. With 10K pairs, BT raises \xxl{} into Odia from 64.33 to 72.04 at 40K and 74.17 at 80K.

\subsection{Gains Hold Across Languages and Backbones}
\label{sec:crosslang}
 
Table~\ref{tab:main-crosslang} tests whether the gains generalize to six English-paired languages from Indic and African families and to a second backbone. \BTtechnique{} is the best of the three systems in 45 of the 48 cells (selected example outputs are in Appendix~\ref{app:examples}). Into the low-resource language, it improves on \onesft{} by 16\% to 173\% \xxl{} across the five low-resource languages and both backbones (median 44\%). Into English, the gains are smaller (median 7\%), and the only loss is Sanskrit on \qwen{} ($-8\%$ \xxl{} and $-1\%$ \chrf{}). This asymmetry partly reflects headroom. Conditional perplexity into English is low at the outset and barely moves, while into the low-resource language it starts high and drops sharply for five of the six pairs (Appendix~\ref{app:ppl}). The Hindi control gains only 3\% on \llama{}, whose Hindi is already strong, but 72\% into Hindi on \qwen{}, whose Hindi prior is weak. With one tenth of the parallel data, \BTtechnique{} also matches or exceeds \tensft{} into the low-resource language in eight of ten comparisons, while \tensft{} stays ahead into English (Appendix~\ref{app:heldout}). Multi-seed experiments show that Odia results are stable across three random seeds (Appendix~\ref{app:seeds}).

\subsection{Human Judges Agree with the Metrics}
\label{subsec:human_eval}
 
In a blind study, three independent annotators ranked the overall quality of \entoxx{} translations for Odia and Sanskrit (protocol and agreement in Appendix~\ref{app:human}). They compared \onesft{}, \tensft{} and \BTtechnique{} trained on \llama{} on 110 held-out source sentences. The rankings mirror the automatic metrics (Table~\ref{tab:human_eval}). Every annotator ranks \BTtechnique{} best for both languages, \tensft{} second and \onesft{} last.

\begin{table}[t]
\centering
\footnotesize
\setlength{\tabcolsep}{3.2pt}
\begin{tabular}{@{}l cccc cccc@{}}
\toprule
 & \multicolumn{4}{c}{\textbf{Odia}} & \multicolumn{4}{c}{\textbf{Sanskrit}} \\
\cmidrule(lr){2-5}\cmidrule(l){6-9}
\textbf{System} & A1 & A2 & A3 & All & A1 & A2 & A3 & All \\
\midrule
\onesft & 2.49 & 2.36 & 2.52 & 2.46 & 2.27 & 2.11 & 2.43 & 2.27 \\
\tensft & \underline{1.81} & \underline{1.87} & \underline{1.85} & \underline{1.84} & \underline{1.93} & \underline{2.04} & \underline{1.86} & \underline{1.94} \\
\sftbt & \textbf{1.70} & \textbf{1.76} & \textbf{1.64} & \textbf{1.70} & \textbf{1.80} & \textbf{1.85} & \textbf{1.71} & \textbf{1.79} \\
\bottomrule
\end{tabular}%
\caption{Human evaluation of \entoxx{} translation: average rank (lower is better) from three blind annotators (A1 to A3) on 110 sentences per language. \BTtechnique{} uses sampled pivots. Protocol and inter-annotator agreement are in Appendix~\ref{app:human}.}
\label{tab:human_eval}
\end{table}

\section{Conclusions and Limitations}
\label{sec:conclusion}
We asked how an instruction-tuned LLM with a weak prior should use scarce parallel data and bounded monolingual text in a low-resource language. With identical data on Odia and Swahili, self-distillation failed, and continued pretraining helped but trailed back-translation into the low-resource language. Back-translation keeps original text on the target side. With beam-decoded pivots, it beat continued pretraining and ten times more parallel data, and retrieved in-context examples also improved it. Across five low-resource languages and two 3B backbones, its gains into the low-resource language were large and consistent, and human judges agreed. In practice, we recommend beam-decoded pivots and monolingual text chosen for quality before domain. In this regime the pre-LLM recipe remains the right default. Future work includes more languages and backbones, larger models and pairs that do not involve English.

We study 3B instruction-tuned backbones with about 1K parallel pairs, translating only to and from English. The recipe comparison, the in-context learning study and the human evaluation each cover two languages on one backbone. Our ALMA-style and CycleDistill-style baselines are re-implementations at this scale, and self-distillation may work with stronger priors or larger models. We combine ICL with beam-decoded pivots only for Odia and with a single seed, and we do not yet apply back-translation after CPT.

\section*{Author Contributions}
\noindent D.B. implemented the system, ran the experiments, and performed the analysis. All authors contributed to study design and interpretation of results, provided feedback throughout the project, and participated in writing, reviewing, and approving the final manuscript. J.S. and P.S. arranged the computational resources required for the project.

\section*{AI Use Statement}
In this work, we used generative AI tools for two purposes. First, they assisted with writing and debugging code for experiment scripts, data processing and figure generation. Second, they helped with language editing of the manuscript, including grammar, wording and sentence flow. We did not use generative AI tools to generate research ideas, design the method or experiments, produce or alter experimental results, or write the related-work analysis. All AI-assisted code was reviewed, tested and verified for correctness by the authors. All AI-edited text was reviewed by the authors for accuracy and consistency with our results. We take full responsibility for the final content of this work, including text, claims and artifacts produced with the aid of generative AI.

\bibliography{main}

@inproceedings{brown2020language,
  title     = {Language Models are Few-Shot Learners},
  author    = {Brown, Tom and others},
  booktitle = {Advances in Neural Information Processing Systems},
  volume    = {33},
  pages     = {1877--1901},
  publisher = {Curran Associates, Inc.},
  year      = {2020},
  url       = {https://proceedings.neurips.cc/paper_files/paper/2020/file/1457c0d6bfcb4967418bfb8ac142f64a-Paper.pdf}
}

@inproceedings{xu2024paradigm,
  title     = {A Paradigm Shift in Machine Translation: Boosting Translation Performance of Large Language Models},
  author    = {Xu, Haoran and others},
  booktitle = {The Twelfth International Conference on Learning Representations},
  year      = {2024},
  url       = {https://openreview.net/forum?id=farT6XXntP}
}

@misc{costajussa2022nllb,
  title         = {No Language Left Behind: Scaling Human-Centered Machine Translation},
  author        = {{NLLB Team} and others},
  year          = {2022},
  eprint        = {2207.04672},
  archivePrefix = {arXiv},
  primaryClass  = {cs.CL},
  url           = {https://arxiv.org/abs/2207.04672}
}

@inproceedings{sennrich2016improving,
  title     = {Improving Neural Machine Translation Models with Monolingual Data},
  author    = {Sennrich, Rico and others},
  booktitle = {Proceedings of the 54th Annual Meeting of the Association for Computational Linguistics (Volume 1: Long Papers)},
  month     = aug,
  year      = {2016},
  address   = {Berlin, Germany},
  publisher = {Association for Computational Linguistics},
  url       = {https://aclanthology.org/P16-1009/},
  doi       = {10.18653/v1/P16-1009},
  pages     = {86--96}
}

@inproceedings{edunov2018understanding,
  title     = {Understanding Back-Translation at Scale},
  author    = {Edunov, Sergey and others},
  booktitle = {Proceedings of the 2018 Conference on Empirical Methods in Natural Language Processing},
  month     = oct # "-" # nov,
  year      = {2018},
  address   = {Brussels, Belgium},
  publisher = {Association for Computational Linguistics},
  url       = {https://aclanthology.org/D18-1045/},
  doi       = {10.18653/v1/D18-1045},
  pages     = {489--500}
}

@inproceedings{hoang2018iterative,
  title     = {Iterative Back-Translation for Neural Machine Translation},
  author    = {Hoang, Vu Cong Duy and others},
  booktitle = {Proceedings of the 2nd Workshop on Neural Machine Translation and Generation},
  month     = jul,
  year      = {2018},
  address   = {Melbourne, Australia},
  publisher = {Association for Computational Linguistics},
  url       = {https://aclanthology.org/W18-2703/},
  doi       = {10.18653/v1/W18-2703},
  pages     = {18--24}
}

@misc{grattafiori2024llama3,
  title         = {The {Llama} 3 Herd of Models},
  author        = {Grattafiori, Aaron and others},
  year          = {2024},
  eprint        = {2407.21783},
  archivePrefix = {arXiv},
  url           = {https://arxiv.org/abs/2407.21783}
}

@misc{yang2024qwen,
  title         = {{Qwen2.5} Technical Report},
  author        = {Yang, An and others},
  year          = {2024},
  eprint        = {2412.15115},
  archivePrefix = {arXiv},
  primaryClass  = {cs.CL},
  url           = {https://arxiv.org/abs/2412.15115}
}

@inproceedings{papineni2002bleu,
  title     = {{BLEU}: a Method for Automatic Evaluation of Machine Translation},
  author    = {Papineni, Kishore and others},
  booktitle = {Proceedings of the 40th Annual Meeting of the Association for Computational Linguistics},
  month     = jul,
  year      = {2002},
  address   = {Philadelphia, Pennsylvania, USA},
  publisher = {Association for Computational Linguistics},
  pages     = {311--318},
  doi       = {10.3115/1073083.1073135},
  url       = {https://aclanthology.org/P02-1040/}
}

@inproceedings{zhang2020bertscore,
  title     = {{BERTScore}: Evaluating Text Generation with {BERT}},
  author    = {Zhang, Tianyi and others},
  booktitle = {International Conference on Learning Representations},
  year      = {2020},
  url       = {https://openreview.net/forum?id=SkeHuCVFDr}
}

@inproceedings{vilar2023prompting,
  title     = {Prompting {P}a{LM} for Translation: Assessing Strategies and Performance},
  author    = {Vilar, David and others},
  booktitle = {Proceedings of the 61st Annual Meeting of the Association for Computational Linguistics (Volume 1: Long Papers)},
  month     = jul,
  year      = {2023},
  address   = {Toronto, Canada},
  publisher = {Association for Computational Linguistics},
  pages     = {15406--15427},
  doi       = {10.18653/v1/2023.acl-long.859},
  url       = {https://aclanthology.org/2023.acl-long.859/}
}

@article{gala2023indictrans2,
  title   = {{IndicTrans2}: Towards High-Quality and Accessible Machine Translation Models for all 22 Scheduled {Indian} Languages},
  author  = {Gala, Jay and others},
  journal = {Transactions on Machine Learning Research},
  issn    = {2835-8856},
  year    = {2023},
  url     = {https://openreview.net/forum?id=vfT4YuzAYA}
}

@misc{hendy2023llmmt,
  title         = {How Good Are {GPT} Models at Machine Translation? A Comprehensive Evaluation},
  author        = {Hendy, Amr and others},
  year          = {2023},
  eprint        = {2302.09210},
  archivePrefix = {arXiv},
  primaryClass  = {cs.CL},
  url           = {https://arxiv.org/abs/2302.09210}
}

@inproceedings{dettmers2023qlora,
  title     = {{QLoRA}: Efficient Finetuning of Quantized {LLMs}},
  author    = {Dettmers, Tim and others},
  booktitle = {Advances in Neural Information Processing Systems},
  volume    = {36},
  pages     = {10088--10115},
  publisher = {Curran Associates, Inc.},
  year      = {2023},
  doi       = {10.52202/075280-0441}
}

@misc{meta2024llama32,
  title        = {{Llama} 3.2: Revolutionizing edge {AI} and vision with open, customizable models},
  author       = {{Meta AI}},
  year         = {2024},
  month        = sep,
  howpublished = {\url{https://ai.meta.com/blog/llama-3-2-connect-2024-vision-edge-mobile-devices/}},
  note         = {Accessed: 2026-10-12}
}

@inproceedings{parida-etal-2020-odiencorp,
  title     = {{OdiEnCorp} 2.0: {Odia-English} Parallel Corpus for Machine Translation},
  author    = {Parida, Shantipriya and others},
  booktitle = {Proceedings of the WILDRE5--5th Workshop on Indian Language Data: Resources and Evaluation},
  month     = may,
  year      = {2020},
  pages     = {14--19},
  address   = {Marseille, France},
  publisher = {European Language Resources Association (ELRA)},
  isbn      = {979-10-95546-67-2},
  url       = {https://aclanthology.org/2020.wildre-1.3/}
}

@misc{haddow2020pmindia,
  title         = {{PMIndia} -- A Collection of Parallel Corpora of Languages of {India}},
  author        = {Haddow, Barry and Kirefu, Faheem},
  year          = {2020},
  eprint        = {2001.09907},
  archivePrefix = {arXiv},
  primaryClass  = {cs.CL},
  url           = {https://arxiv.org/abs/2001.09907}
}

@article{ramesh2022samanantar,
  title     = {Samanantar: The Largest Publicly Available Parallel Corpora Collection for 11 {Indic} Languages},
  author    = {Ramesh, Gowtham and others},
  journal   = {Transactions of the Association for Computational Linguistics},
  volume    = {10},
  pages     = {145--162},
  year      = {2022},
  publisher = {MIT Press},
  address   = {Cambridge, MA},
  doi       = {10.1162/tacl_a_00452},
  url       = {https://aclanthology.org/2022.tacl-1.9/}
}

@inproceedings{adelani-etal-2022-thousand,
  title     = {A Few Thousand Translations Go a Long Way! Leveraging Pre-trained Models for {A}frican News Translation},
  author    = {Adelani, David and others},
  booktitle = {Proceedings of the 2022 Conference of the North American Chapter of the Association for Computational Linguistics: Human Language Technologies},
  month     = jul,
  year      = {2022},
  address   = {Seattle, United States},
  publisher = {Association for Computational Linguistics},
  url       = {https://aclanthology.org/2022.naacl-main.223},
  doi       = {10.18653/v1/2022.naacl-main.223},
  pages     = {3053--3070}
}

@misc{nehrdich2026mitrasamgraha,
  title         = {Mitrasa{\d{m}}graha: A Comprehensive Classical {Sanskrit} Machine Translation Dataset},
  author        = {Nehrdich, Sebastian and others},
  year          = {2026},
  eprint        = {2601.07314},
  archivePrefix = {arXiv},
  primaryClass  = {cs.CL},
  url           = {https://arxiv.org/abs/2601.07314}
}

@inproceedings{kunchukuttan-etal-2018-iit,
  title     = {The {IIT} {Bombay} {English-Hindi} Parallel Corpus},
  author    = {Kunchukuttan, Anoop and others},
  booktitle = {Proceedings of the Eleventh International Conference on Language Resources and Evaluation ({LREC} 2018)},
  month     = may,
  year      = {2018},
  address   = {Miyazaki, Japan},
  publisher = {European Language Resources Association (ELRA)},
  url       = {https://aclanthology.org/L18-1548/}
}

@inproceedings{conneau-etal-2020-unsupervised,
  title     = {Unsupervised Cross-lingual Representation Learning at Scale},
  author    = {Conneau, Alexis and others},
  booktitle = {Proceedings of the 58th Annual Meeting of the Association for Computational Linguistics},
  month     = jul,
  year      = {2020},
  address   = {Online},
  publisher = {Association for Computational Linguistics},
  pages     = {8440--8451},
  doi       = {10.18653/v1/2020.acl-main.747},
  url       = {https://aclanthology.org/2020.acl-main.747/}
}

@inproceedings{wenzek-etal-2020-ccnet,
  title     = {{CCNet}: Extracting High Quality Monolingual Datasets from Web Crawl Data},
  author    = {Wenzek, Guillaume and others},
  booktitle = {Proceedings of the Twelfth Language Resources and Evaluation Conference},
  month     = may,
  year      = {2020},
  address   = {Marseille, France},
  publisher = {European Language Resources Association},
  pages     = {4003--4012},
  isbn      = {979-10-95546-34-4},
  url       = {https://aclanthology.org/2020.lrec-1.494/}
}

@inproceedings{doddapaneni-etal-2023-towards,
  title     = {Towards Leaving No {I}ndic Language Behind: Building Monolingual Corpora, Benchmark and Models for {I}ndic Languages},
  author    = {Doddapaneni, Sumanth and others},
  booktitle = {Proceedings of the 61st Annual Meeting of the Association for Computational Linguistics (Volume 1: Long Papers)},
  month     = jul,
  year      = {2023},
  address   = {Toronto, Canada},
  publisher = {Association for Computational Linguistics},
  pages     = {12402--12426},
  doi       = {10.18653/v1/2023.acl-long.693},
  url       = {https://aclanthology.org/2023.acl-long.693/}
}

@inproceedings{schwenk-etal-2021-wikimatrix,
  title     = {{WikiMatrix}: Mining 135{M} Parallel Sentences in 1620 Language Pairs from {Wikipedia}},
  author    = {Schwenk, Holger and others},
  booktitle = {Proceedings of the 16th Conference of the European Chapter of the Association for Computational Linguistics: Main Volume},
  month     = apr,
  year      = {2021},
  address   = {Online},
  publisher = {Association for Computational Linguistics},
  pages     = {1351--1361},
  doi       = {10.18653/v1/2021.eacl-main.115},
  url       = {https://aclanthology.org/2021.eacl-main.115/}
}

@inproceedings{lora,
  title     = {{LoRA}: Low-Rank Adaptation of Large Language Models},
  author    = {Hu, Edward J. and others},
  booktitle = {International Conference on Learning Representations},
  year      = {2022},
  url       = {https://openreview.net/forum?id=nZeVKeeFYf9}
}

@article{xcomet,
  title     = {{xCOMET}: Transparent Machine Translation Evaluation through Fine-grained Error Detection},
  author    = {Guerreiro, Nuno M. and others},
  journal   = {Transactions of the Association for Computational Linguistics},
  volume    = {12},
  pages     = {979--995},
  year      = {2024},
  publisher = {MIT Press},
  doi       = {10.1162/tacl_a_00683},
  url       = {https://aclanthology.org/2024.tacl-1.54/}
}

@inproceedings{post2018sacrebleu,
  title     = {A Call for Clarity in Reporting {BLEU} Scores},
  author    = {Post, Matt},
  booktitle = {Proceedings of the Third Conference on Machine Translation: Research Papers},
  month     = oct,
  year      = {2018},
  address   = {Brussels, Belgium},
  publisher = {Association for Computational Linguistics},
  pages     = {186--191},
  doi       = {10.18653/v1/W18-6319},
  url       = {https://aclanthology.org/W18-6319/}
}

@inproceedings{popovic2017chrfpp,
  title     = {{chrF++}: words helping character n-grams},
  author    = {Popovi{\'c}, Maja},
  booktitle = {Proceedings of the Second Conference on Machine Translation},
  month     = sep,
  year      = {2017},
  address   = {Copenhagen, Denmark},
  publisher = {Association for Computational Linguistics},
  pages     = {612--618},
  doi       = {10.18653/v1/W17-4770},
  url       = {https://aclanthology.org/W17-4770/}
}

@misc{khanuja2021muril,
  title         = {{MuRIL}: Multilingual Representations for {Indian} Languages},
  author        = {Khanuja, Simran and others},
  year          = {2021},
  eprint        = {2103.10730},
  archivePrefix = {arXiv},
  primaryClass  = {cs.CL},
  url           = {https://arxiv.org/abs/2103.10730}
}

@inproceedings{caswell2019tagged,
  title     = {Tagged Back-Translation},
  author    = {Caswell, Isaac and others},
  booktitle = {Proceedings of the Fourth Conference on Machine Translation (Volume 1: Research Papers)},
  month     = aug,
  year      = {2019},
  address   = {Florence, Italy},
  publisher = {Association for Computational Linguistics},
  url       = {https://aclanthology.org/W19-5206/},
  doi       = {10.18653/v1/W19-5206},
  pages     = {53--63}
}

@article{pang2024rethinking,
  title     = {Rethinking the Exploitation of Monolingual Data for Low-Resource Neural Machine Translation},
  author    = {Pang, Jianhui and others},
  journal   = {Computational Linguistics},
  volume    = {50},
  number    = {1},
  month     = mar,
  year      = {2024},
  address   = {Cambridge, MA},
  publisher = {MIT Press},
  url       = {https://aclanthology.org/2024.cl-1.2/},
  doi       = {10.1162/coli_a_00496},
  pages     = {25--47}
}

@inproceedings{he2016dual,
  title     = {Dual Learning for Machine Translation},
  author    = {He, Di and others},
  booktitle = {Advances in Neural Information Processing Systems},
  volume    = {29},
  pages     = {820--828},
  publisher = {Curran Associates, Inc.},
  year      = {2016},
  url       = {https://proceedings.neurips.cc/paper_files/paper/2016/file/5b69b9cb83065d403869739ae7f0995e-Paper.pdf}
}

@inproceedings{artetxe2018unsupervised,
  title     = {Unsupervised Neural Machine Translation},
  author    = {Artetxe, Mikel and others},
  booktitle = {International Conference on Learning Representations},
  year      = {2018},
  url       = {https://openreview.net/forum?id=Sy2ogebAW}
}

@misc{attia2026improving,
  title         = {Improving Low-Resource Machine Translation via Round-Trip Reinforcement Learning},
  author        = {Attia, Ahmed and Aji, Alham Fikri},
  year          = {2026},
  eprint        = {2601.12535},
  archivePrefix = {arXiv},
  primaryClass  = {cs.CL},
  url           = {https://arxiv.org/abs/2601.12535}
}

@inproceedings{zhu2024multilingual,
  title     = {Multilingual Machine Translation with Large Language Models: Empirical Results and Analysis},
  author    = {Zhu, Wenhao and others},
  booktitle = {Findings of the Association for Computational Linguistics: NAACL 2024},
  month     = jun,
  year      = {2024},
  address   = {Mexico City, Mexico},
  publisher = {Association for Computational Linguistics},
  pages     = {2765--2781},
  doi       = {10.18653/v1/2024.findings-naacl.176},
  url       = {https://aclanthology.org/2024.findings-naacl.176/}
}

@inproceedings{bawden2023investigating,
  title     = {Investigating the Translation Performance of a Large Multilingual Language Model: the case of {BLOOM}},
  author    = {Bawden, Rachel and Yvon, Fran{\c{c}}ois},
  booktitle = {Proceedings of the European Association for Machine Translation (EAMT)},
  year      = {2023}
}

@inproceedings{lu2024llamax,
  title     = {{LL}a{MAX}: Scaling Linguistic Horizons of {LLM} by Enhancing Translation Capabilities Beyond 100 Languages},
  author    = {Lu, Yinquan and others},
  booktitle = {Findings of the Association for Computational Linguistics: EMNLP 2024},
  month     = nov,
  year      = {2024},
  address   = {Miami, Florida, USA},
  publisher = {Association for Computational Linguistics},
  pages     = {10748--10772},
  doi       = {10.18653/v1/2024.findings-emnlp.631},
  url       = {https://aclanthology.org/2024.findings-emnlp.631/}
}

@inproceedings{bafna2025translation,
  title     = {The Translation Barrier Hypothesis: Multilingual Generation with Large Language Models Suffers from Implicit Translation Failure},
  author    = {Bafna, Niyati and others},
  booktitle = {Proceedings of the 14th International Joint Conference on Natural Language Processing and the 4th Conference of the Asia-Pacific Chapter of the Association for Computational Linguistics},
  month     = dec,
  year      = {2025},
  address   = {Mumbai, India},
  publisher = {The Asian Federation of Natural Language Processing and The Association for Computational Linguistics},
  pages     = {1541--1568},
  doi       = {10.18653/v1/2025.ijcnlp-long.83},
  url       = {https://aclanthology.org/2025.ijcnlp-long.83/}
}

@misc{vamvas2026translation,
  title         = {Translation Asymmetry in {LLM}s as a Data Augmentation Factor: A Case Study for 6 {R}omansh Language Varieties},
  author        = {Vamvas, Jannis and others},
  year          = {2026},
  eprint        = {2603.25489},
  archivePrefix = {arXiv},
  primaryClass  = {cs.CL},
  url           = {https://arxiv.org/abs/2603.25489}
}

@inproceedings{zebaze2024incontext,
  title     = {In-Context Example Selection via Similarity Search Improves Low-Resource Machine Translation},
  author    = {Zebaze, Armel Randy and others},
  booktitle = {Findings of the Association for Computational Linguistics: NAACL 2025},
  month     = apr,
  year      = {2025},
  address   = {Albuquerque, New Mexico},
  publisher = {Association for Computational Linguistics},
  pages     = {1222--1252},
  doi       = {10.18653/v1/2025.findings-naacl.68},
  url       = {https://aclanthology.org/2025.findings-naacl.68/}
}

@inproceedings{zebaze2025compositional,
  title     = {Compositional Translation: A Novel {LLM}-based Approach for Low-resource Machine Translation},
  author    = {Zebaze, Armel Randy and others},
  booktitle = {Findings of the Association for Computational Linguistics: EMNLP 2025},
  month     = nov,
  year      = {2025},
  address   = {Suzhou, China},
  publisher = {Association for Computational Linguistics},
  pages     = {22328--22357},
  doi       = {10.18653/v1/2025.findings-emnlp.1216},
  url       = {https://aclanthology.org/2025.findings-emnlp.1216/}
}

@misc{ahmed2025odiaeng,
  author       = {Ahmed, Vaqas},
  title        = {{Odia--English} Sentence-Aligned Precision Corpus},
  year         = {2025},
  howpublished = {Hugging Face dataset \texttt{vaqasai/Odia\_English\_Sentences\_Aligned\_81k} (Vaqas AI)},
  note         = {MIT License. Accessed: 2026-06-15},
  url          = {https://huggingface.co/datasets/vaqasai/Odia_English_Sentences_Aligned_81k}
}

@inproceedings{halder2025cycledistill,
  title     = {{C}ycle{D}istill: Bootstrapping Machine Translation using {LLM}s with Cyclical Distillation},
  author    = {Halder, Deepon and others},
  booktitle = {Proceedings of the Twelfth Workshop on Asian Translation (WAT 2025)},
  month     = dec,
  year      = {2025},
  address   = {Mumbai, India},
  publisher = {Association for Computational Linguistics},
  pages     = {78--92},
  doi       = {10.18653/v1/2025.wat-1.7},
  url       = {https://aclanthology.org/2025.wat-1.7/}
}

@article{wilcoxon1945individual,
  title   = {Individual Comparisons by Ranking Methods},
  author  = {Wilcoxon, Frank},
  journal = {Biometrics Bulletin},
  volume  = {1},
  number  = {6},
  pages   = {80--83},
  year    = {1945},
  month   = dec,
  doi     = {10.2307/3001968}
}

@inproceedings{feng2022language,
  title     = {Language-agnostic {BERT} Sentence Embedding},
  author    = {Feng, Fangxiaoyu and others},
  booktitle = {Proceedings of the 60th Annual Meeting of the Association for Computational Linguistics (Volume 1: Long Papers)},
  month     = may,
  year      = {2022},
  address   = {Dublin, Ireland},
  publisher = {Association for Computational Linguistics},
  pages     = {878--891},
  doi       = {10.18653/v1/2022.acl-long.62},
  url       = {https://aclanthology.org/2022.acl-long.62/}
}

@misc{liu2019roberta,
  title         = {{RoBERTa}: A Robustly Optimized {BERT} Pretraining Approach},
  author        = {Liu, Yinhan and others},
  year          = {2019},
  eprint        = {1907.11692},
  archivePrefix = {arXiv},
  primaryClass  = {cs.CL},
  url           = {https://arxiv.org/abs/1907.11692}
}

@inproceedings{chen2014smoothing,
  title     = {A Systematic Comparison of Smoothing Techniques for Sentence-Level {BLEU}},
  author    = {Chen, Boxing and Cherry, Colin},
  booktitle = {Proceedings of the Ninth Workshop on Statistical Machine Translation},
  pages     = {362--367},
  year      = {2014},
  month     = jun,
  address   = {Baltimore, Maryland, USA},
  publisher = {Association for Computational Linguistics},
  doi       = {10.3115/v1/W14-3346},
  url       = {https://aclanthology.org/W14-3346/}
}

@inproceedings{niu2018bidirectional,
  title     = {Bi-Directional Neural Machine Translation with Synthetic Parallel Data},
  author    = {Niu, Xing and others},
  booktitle = {Proceedings of the 2nd Workshop on Neural Machine Translation and Generation},
  month     = jul,
  year      = {2018},
  address   = {Melbourne, Australia},
  publisher = {Association for Computational Linguistics},
  url       = {https://aclanthology.org/W18-2710/},
  doi       = {10.18653/v1/W18-2710},
  pages     = {84--91}
}

@inproceedings{gu2018search,
  title     = {Search Engine Guided Neural Machine Translation},
  author    = {Gu, Jiatao and others},
  booktitle = {Proceedings of the AAAI Conference on Artificial Intelligence},
  volume    = {32},
  year      = {2018},
  url       = {https://ojs.aaai.org/index.php/AAAI/article/view/12013},
  doi       = {10.1609/aaai.v32i1.12013}
}

@inproceedings{bulte2019neural,
  title     = {Neural Fuzzy Repair: Integrating Fuzzy Matches into Neural Machine Translation},
  author    = {Bulte, Bram and Tezcan, Arda},
  booktitle = {Proceedings of the 57th Annual Meeting of the Association for Computational Linguistics},
  month     = jul,
  year      = {2019},
  address   = {Florence, Italy},
  publisher = {Association for Computational Linguistics},
  url       = {https://aclanthology.org/P19-1175/},
  doi       = {10.18653/v1/P19-1175},
  pages     = {1800--1809}
}

@misc{gemmateam2024gemma2,
  title         = {{Gemma} 2: Improving Open Language Models at a Practical Size},
  author        = {{Gemma Team} and others},
  year          = {2024},
  eprint        = {2408.00118},
  archivePrefix = {arXiv},
  url           = {https://arxiv.org/abs/2408.00118}
}

@inproceedings{wolf-etal-2020-transformers,
  title     = {Transformers: State-of-the-Art Natural Language Processing},
  author    = {Wolf, Thomas and others},
  booktitle = {Proceedings of the 2020 Conference on Empirical Methods in Natural Language Processing: System Demonstrations},
  month     = oct,
  year      = {2020},
  address   = {Online},
  publisher = {Association for Computational Linguistics},
  pages     = {38--45},
  doi       = {10.18653/v1/2020.emnlp-demos.6},
  url       = {https://aclanthology.org/2020.emnlp-demos.6/}
}

@inproceedings{ansel2024pytorch2,
  title     = {{PyTorch} 2: Faster Machine Learning Through Dynamic {Python} Bytecode Transformation and Graph Compilation},
  author    = {Ansel, Jason and others},
  booktitle = {Proceedings of the 29th ACM International Conference on Architectural Support for Programming Languages and Operating Systems, Volume 2 (ASPLOS '24)},
  month     = apr,
  year      = {2024},
  publisher = {ACM},
  doi       = {10.1145/3620665.3640366}
}

@misc{mangrulkar2022peft,
  title        = {{PEFT}: State-of-the-art Parameter-Efficient Fine-Tuning methods},
  author       = {Mangrulkar, Sourab and others},
  year         = {2022},
  howpublished = {\url{https://github.com/huggingface/peft}}
}

@misc{gugger2022accelerate,
  title        = {{Accelerate}: Training and inference at scale made simple, efficient and adaptable},
  author       = {Gugger, Sylvain and others},
  year         = {2022},
  howpublished = {\url{https://github.com/huggingface/accelerate}}
}

@inproceedings{lample2018unsupervised,
  title     = {Unsupervised Machine Translation Using Monolingual Corpora Only},
  author    = {Lample, Guillaume and others},
  booktitle = {International Conference on Learning Representations},
  year      = {2018},
  url       = {https://openreview.net/forum?id=rkYTTf-AZ}
}

@inproceedings{koehn2004statistical,
  title     = {Statistical Significance Tests for Machine Translation Evaluation},
  author    = {Koehn, Philipp},
  booktitle = {Proceedings of the 2004 Conference on Empirical Methods in Natural Language Processing},
  pages     = {388--395},
  year      = {2004},
  address   = {Barcelona, Spain},
  publisher = {Association for Computational Linguistics},
  url       = {https://aclanthology.org/W04-3250/}
}

@article{holm1979simple,
  title   = {A Simple Sequentially Rejective Multiple Test Procedure},
  author  = {Holm, Sture},
  journal = {Scandinavian Journal of Statistics},
  volume  = {6},
  number  = {2},
  pages   = {65--70},
  year    = {1979}
}
\bibliographystyle{plainnat}

\newpage
\appendix
\lstdefinestyle{promptstyle}{
  basicstyle=\ttfamily\scriptsize,
  breaklines=true,
  breakindent=0pt,
  postbreak=\mbox{$\hookrightarrow$\space},
  columns=fullflexible,
  keepspaces=true,
  upquote=true,
  frame=single,
  framesep=4pt,
  xleftmargin=4pt,
  escapeinside={(*@}{@*)},
  literate={,}{{,}\allowbreak}1
}

% \begin{center}
% \Large 
%     \textbf{Appendix\\Back-Translation Still Matters: Leveraging Monolingual Data for Low-Resource Machine Translation with LLMs}
% \end{center}

% --- Float control in appendix ---
% (Disabled) Previously, we inserted \FloatBarrier before each (sub)section to prevent floats
% from moving across appendix sections.

\section{Implementation and Reproducibility Details}
\label{app:impl}

\paragraph{Corpora:} Table~\ref{tab:corpora} lists the parallel and monolingual corpora for each language pair. Unless noted otherwise, the monolingual pools come from the same source as the parallel seed, after the 1K seed pairs are removed. Table~\ref{tab:corpora1} lists the alternative monolingual corpora used in the source analysis of \S\ref{sec:analysis-domain}. The VaqasAI corpus card states that its translations were produced with Gemini from English source text, so the Odia side of the VaqasAI seed, pools and held-out test set is machine-translated. Flores-200 is our human-translated Odia test set.

\begin{table*}[tp]
\centering
{\small
\setlength{\tabcolsep}{3pt}
\renewcommand{\arraystretch}{1.1}
\begin{tabular}{@{}l>{\raggedright\arraybackslash}p{0.55\linewidth}>{\raggedright\arraybackslash}p{0.2\linewidth}l@{}}
\toprule
Pair & Parallel corpus & Monolingual Corpus & \tensft~\\
\midrule
Odia--En     & VaqasAI (+20K from PMIndia) \citep{ahmed2025odiaeng,haddow2020pmindia} & same-corpus & yes \\
Odia--En     & OdiEnCorp-2.0 \citep{parida-etal-2020-odiencorp} Used for domain-transfer analysis (\S\ref{sec:analysis-domain}) & same-corpus & yes \\
Assamese--En & Samanantar \citep{ramesh2022samanantar} & same-corpus & yes \\
Hindi--En    & IIT Bombay \citep{kunchukuttan-etal-2018-iit} & same-corpus & yes \\
Sanskrit--En & Mitrasa\d{m}graha \citep{nehrdich2026mitrasamgraha} & same-corpus & yes \\
Amharic--En  & MAFAND \citep{adelani-etal-2022-thousand}. No training split: we merge its dev and test sets ($\sim$1.9K pairs) and re-split them into the 1K seed, 450 development and 486 test pairs & CC-100 & no \\
Swahili--En  & MAFAND \citep{adelani-etal-2022-thousand} & MAFAND + CC-100 & yes \\
\bottomrule
\end{tabular}
}
\caption{Data sources per language pair. ``Same-corpus'' = monolingual pools are disjoint slices of the seed corpus excluding the 1K SFT pairs.}
\label{tab:corpora}
\end{table*}

\begin{table*}[tp]
\centering
{\small
\begin{tabular}{@{}llll@{}}
\toprule
Pair & Parallel corpus & Monolingual Corpus & in/out-domain \\
\midrule
Odia--En     & VaqasAI & VaqasAI & in-domain\\
Odia--En     & VaqasAI & OdiEnCorp-2.0 & out-of-domain\\
Odia--En     & VaqasAI & CC-100 & out-of-domain\\
Odia--En     & VaqasAI & IndicCorp-2.0 (Odia), CC-100 (En) & out-of-domain\\
Odia--En     & VaqasAI & IndicCorp-2.0 (Odia), WikiMatrix (En) & out-of-domain\\
\midrule
Assamese--En & Samanantar & Samanantar & in-domain\\
Assamese--En & Samanantar & CC-100 & out-of-domain\\
Assamese--En & Samanantar & IndicCorp-2.0 (Assamese), CC-100 (En) & out-of-domain\\
\bottomrule
\end{tabular}
}
\caption{Additional data source used for off-the-shelf monolingual analysis for Odia and Assamese languages. VaqasAI was appended with additional 20K sentence from PMIndia data source. OdiEnCorp-2.0 \citep{parida-etal-2020-odiencorp}, CC-100 \citep{conneau-etal-2020-unsupervised,wenzek-etal-2020-ccnet}, IndicCorp-2.0 \citep{doddapaneni-etal-2023-towards} and WikiMatrix \citep{schwenk-etal-2021-wikimatrix}.}
\label{tab:corpora1}
\end{table*}

\paragraph{Monolingual token budget:}
Each language has a pool of 40K monolingual sentences. We count the tokens of the pool text as the training loader reads it, without special tokens or the chat template. With the \llama{} tokenizer, the Odia--English pools contain 9.69M tokens (8.75M Odia and 0.94M English), and the Swahili--English pools contain 3.51M (2.36M Swahili and 1.15M English). With the \qwen{} tokenizer, the totals are 8.01M (7.06M Odia and 0.95M English) and 3.57M (2.41M Swahili and 1.16M English). Under \llama{}, an Odia pool sentence averages 219 tokens, a Swahili sentence 59 and an English sentence 24 to 29. Both totals are at least two orders of magnitude below the 1B to 20B monolingual tokens used by ALMA \citep{xu2024paradigm}.

\paragraph{Training configuration:}
Both backbones use identical settings. The causal language-modelling loss is computed on target tokens only, under the model's chat template wrapped in a translation instruction. Every system uses the data, prompt template and metrics described in this section, and is decoded greedily at test time. Each configuration is trained and evaluated once and Appendix~\ref{app:seeds} repeats the Odia runs with three seeds. The recipes differ in what is trained, on which data and for how long, and Appendix~\ref{app:recipes} gives the settings of each recipe (supervised fine-tuning and \BTtechnique{} in Table~\ref{tab:hyperparams}, \cptsft{} in Table~\ref{tab:hyperparams-cpt}, self-distillation in Table~\ref{tab:hyperparams-sd}).

\paragraph{Prompt template:}
All models are prompted with the instruction below, wrapped in the backbone's chat template, with the loss computed on target tokens only. \texttt{\{SRC\_LANG\}} and \texttt{\{TGT\_LANG\}} are replaced by the language names (e.g.\ English, Odia), and \texttt{\{SRC\_SEN\}} by the sentence to translate. Under in-context learning (Appendix~\ref{app:icl}), up to $k{=}5$ retrieved demonstrations are prepended before that line.

\medskip
\noindent\textbf{Listing 1: Base prompt (no ICL).}
\nopagebreak
\begin{lstlisting}
Translate {SRC_LANG} to {TGT_LANG} without any explanation: {SRC_SEN}
\end{lstlisting}
\medskip

\noindent\textbf{Listing 2: Prompt with in-context learning ($k=5$).}
\nopagebreak
\begin{lstlisting}
Here are some similar translation examples:

{SRC_LANG}: {retrieved_source_1}
{TGT_LANG}: {retrieved_target_1}

{SRC_LANG}: {retrieved_source_2}
{TGT_LANG}: {retrieved_target_2}
... (up to 5 examples) ...

Now translate {SRC_LANG} to {TGT_LANG} without any explanation: {SRC_SEN}
\end{lstlisting}
\medskip

\paragraph{Automatic metrics:}
We report five automatic metrics, computed identically for both directions and both evaluation sets. For learned evaluation we use XCOMET \citep{xcomet} in its reference-based setting, scoring each (source, hypothesis, reference) triple with the \texttt{Unbabel/XCOMET-XXL} and \texttt{Unbabel/XCOMET-XL} checkpoints. XCOMET does not check whether the hypothesis is in the requested language, so we always report \chrf{} alongside it. \chrf~\citep{popovic2017chrfpp} uses the \sbleu~\citep{post2018sacrebleu} CHRF implementation with \texttt{word\_order=2} (character $n$-grams plus word unigrams and bigrams). BLEU \citep{papineni2002bleu} uses the NLTK sentence-level implementation with uniform $4$-gram weights, Chen and Cherry smoothing method~2 \citep{chen2014smoothing} and whitespace tokenisation. Both are averaged over sentences and scaled to $0$--$100$. For \bert~F1 \citep{zhang2020bertscore} we use a target-language encoder, namely \texttt{roberta-large} \citep{liu2019roberta} into English, \texttt{google/muril-large-cased} \citep{khanuja2021muril} for the four Indic targets, and \texttt{xlm-roberta-large} \citep{conneau-etal-2020-unsupervised} for Amharic and Swahili. We report the baseline-normalised value $(\mathrm{F1}-b)/(1-b)$. Into English we use the built-in baseline. For the other targets, $b$ is the mean \bert~F1 over 100K random sentence pairs from the corresponding monolingual corpus (Table~\ref{tab:bertscore-baselines}).

\begin{table}[t]
\centering
{\footnotesize
\setlength{\tabcolsep}{2.5pt}
\renewcommand{\arraystretch}{1.1}
\begin{tabular}{@{}ll c S[table-format=1.6]@{}}
\toprule
Target & Encoder & Layer & {Baseline $b$} \\
\midrule
English & \texttt{roberta-large} & default & {built-in} \\
\midrule
Hindi    & \texttt{muril-large-cased} & 18 & 0.473602 \\
Assamese & \texttt{muril-large-cased} & 18 & 0.503943 \\
Odia     & \texttt{muril-large-cased} & 18 & 0.499375 \\
Sanskrit & \texttt{muril-large-cased} & 18 & 0.496216 \\
\midrule
Amharic  & \texttt{xlm-roberta-large} & 17 & 0.819427 \\
Swahili  & \texttt{xlm-roberta-large} & 17 & 0.829090 \\
\bottomrule
\end{tabular}
}
\caption{Per-language \bert~baselines $b$ used for the normalisation with the encoder and layer for each target.}
\label{tab:bertscore-baselines}
\end{table}

\paragraph{In-context retrieval:}
For the in-context learning experiments (Appendix~\ref{app:icl}), we retrieve up to $k{=}5$ nearest demonstrations by cosine similarity from the same 1K gold seed used for supervised fine-tuning, so no new annotation is needed. Retrieval is monolingual on the source side. Each query is embedded with a language-specific sentence encoder and searched against a per-language index, and the corresponding source--target pairs serve as demonstrations. We embed Amharic and Swahili with LaBSE \citep{feng2022language} (\texttt{sentence-transformers/LaBSE}, \texttt{pooler\_output}, $L_2$ normalised), Odia, Assamese, Sanskrit and Hindi with MuRIL-large \citep{khanuja2021muril} (\texttt{google/muril-large-cased}), and English with RoBERTa-large \citep{liu2019roberta}. For MuRIL and RoBERTa we mean-pool the last four layers. Inputs are truncated to $512$ tokens. We exclude the query's exact match after NFC normalisation and whitespace collapsing. Retrieved examples are prepended to the instruction as a few-shot block under the chat template. If the prompt exceeds $2{,}048$ tokens, we reduce the number of shots ($5\!\rightarrow\!4\!\rightarrow\!3\dots$) until it fits.

\paragraph{Software and hardware:}
Experiments were implemented in PyTorch \citep{ansel2024pytorch2} with HuggingFace Transformers \citep{wolf-etal-2020-transformers}, PEFT \citep{mangrulkar2022peft}, Accelerate \citep{gugger2022accelerate}, and bitsandbytes. Training used DistributedDataParallel on NVIDIA A100 (80GB) GPUs. Table~\ref{tab:software} lists the versions of all packages. All experiments were tracked with Weights and Biases. Unless noted otherwise, each run used eight GPUs. All times below are approximate wall-clock times for a single run. A \BTtechnique{} run took about 48 hours with sampled pivots and 72 hours with beam-decoded pivots, and an ICL run took about 72 hours. A \cptsft{} run took about 12 hours on two GPUs, and a self-distillation run (three cycles) took about 20 hours on four GPUs.

\begin{table}[t]
\centering
{\footnotesize
\setlength{\tabcolsep}{4pt}
\begin{tabular}{@{}ll@{}}
\toprule
Package & Version \\
\midrule
Python                     & 3.12 \\
PyTorch (CUDA 12.8)        & 2.10.0 (for \xxl: 2.11.0) \\
Transformers               & 5.3.0 (for \xxl: 4.57.6) \\
PEFT                       & 0.18.1 \\
Accelerate                 & 1.12.0 \\
bitsandbytes               & 0.49.1 \\
Datasets                   & 4.5.0 \\
Weights \& Biases          & 0.25.0 \\
sacreBLEU                  & 2.6.0 \\
NLTK                       & 3.9.2 \\
bert-score                 & 0.3.13 \\
unbabel-comet      & 2.2.7 \\
PyTorch Lightning & 2.6.5 \\
\bottomrule
\end{tabular}
}
\caption{Package versions used for training and evaluation. For the XCOMET calculation we used the versions in parentheses.}
\label{tab:software}
\end{table}

\section{Training Recipes}
\label{app:recipes}

This section gives the settings of the recipes compared in \S\ref{sec:recipes}. All of them use the data, prompt template and metrics of Appendix~\ref{app:impl}, the same settings on both backbones, and the quantisation and LoRA configuration of Table~\ref{tab:hyperparams}.

\paragraph{\BTtechnique{}:}
One epoch is one pass over the back-translation data, which with 40K monolingual sentences per language is 780 optimizer steps (156 supervised and 624 back-translation). Each run is trained until its development score converges. \cptrecast{} starts \BTtechnique{} from the stage-1 model of \cptsft{}.

\begin{table}[t]
\centering
{\footnotesize
\begin{tabular}{@{}l>{\raggedright\arraybackslash}p{0.57\columnwidth}@{}}
\toprule
Quantization         & 4-bit NF4 (double quantization), bf16 compute \\
LoRA                 & rank $64$, $\alpha{=}64$, dropout $0.05$ \\
LoRA target          & all attention + feed-forward projections \\
Trainable parameters & 97.3M (\llama{}), 119.7M (\qwen{}) \\
Optimizer            & PagedAdamW (32-bit), gradient clipping at norm 1.0 \\
Peak learning rate   & $2\times10^{-4}$, cosine decay, $5\%$ warmup \\
Interleaving         & per direction, 1 supervised step then $r{=}4$ BT steps \\
Training length      & until the development \bert~converges \\
Max sequence length  & $2{,}048$ tokens \\
Effective Batch size           &  128 \\
Validation           & every quarter-epoch, development \bert~\\
BT pivot decoding    & nucleus, $\tau{=}0.3$, top-$p{=}0.9$ \\
                     & or beam search, 5 beams \\
                     & at most 256 new tokens \\
Eval decoding        & greedy, at most 256 new tokens \\
\bottomrule
\end{tabular}
}
\caption{\BTtechnique{} training and decoding hyperparameters (shared across both backbones). \onesft{} and \tensft{} use the same settings without back-translation steps.}
\label{tab:hyperparams}
\end{table}

\paragraph{ALMA-style \cptsft{}:}
We follow ALMA's 7B-LoRA setup \citep{xu2024paradigm}: full-weight continued pretraining on the same monolingual pools that \BTtechnique{} back-translates, then LoRA fine-tuning on the 1K pairs in both directions with mixed-direction batches (Table~\ref{tab:hyperparams-cpt}). The same stage 2 without CPT is the matched control, so the two differ only in stage 1. Our setting is much smaller than ALMA's: a 3B instead of a 7B or 13B model, about 9M instead of up to 20B monolingual tokens, and 1K instead of 58K parallel pairs.

\begin{table}[t]
\centering
{\footnotesize
\begin{tabular}{@{}l>{\raggedright\arraybackslash}p{0.57\columnwidth}@{}}
\toprule
\multicolumn{2}{@{}l}{\textit{Stage 1: continued pretraining}} \\
Trained weights      & all weights, bf16 \\
Data                 & 40K XX + 40K En monolingual sentences, packed into 512-token blocks \\
Effective Batch size           & 256 blocks \\
Peak learning rate   & $2\times10^{-5}$, inverse-square-root decay to 10\% of the peak, $1\%$ warmup \\
Weight decay         & 0.01 \\
Epochs               & 8 \\
Validation           & held-out LM loss (1K + 1K sentences) before training and after each epoch, lowest loss kept \\
\midrule
\multicolumn{2}{@{}l}{\textit{Stage 2: supervised fine-tuning}} \\
Trained weights      & LoRA on the 4-bit CPT model (as Table~\ref{tab:hyperparams}) \\
Data                 & 1K gold pairs in both directions (2{,}000 rows), mixed batches \\
Peak learning rate   & $2\times10^{-4}$, cosine decay, $5\%$ warmup \\

Max sequence length  & $2{,}048$ tokens \\
Effective Batch size           & 128 \\
Validation           & every quarter-epoch, development \bert~in both directions (highest mean kept, ties broken by BLEU) \\
Eval decoding        & greedy, at most 256 new tokens \\
\bottomrule
\end{tabular}
}
\caption{\cptsft{} hyperparameters (ALMA-style, shared across both backbones). Stage 2 without stage 1 is the matched control.}
\label{tab:hyperparams-cpt}
\end{table}

\paragraph{CycleDistill-style self-distillation:}
CycleDistill \citep{halder2025cycledistill} does not state its epochs, learning rate, decoding or checkpoint selection (its released code sets defaults for some of these), so we use our shared settings (Table~\ref{tab:hyperparams-sd}).
Self-distill (0-shot) uses zero-shot prompts throughout, and self-distill (1-shot) places one gold example from the 1K seed in every labelling, training, validation and test prompt. Unlike CycleDistill, we use 40K instead of 20K pool sentences, and let a 3B model act as its own teacher instead of an 8B or 9B teacher.
\begin{table}[t]
\centering
{\footnotesize
\begin{tabular}{@{}l>{\raggedright\arraybackslash}p{0.57\columnwidth}@{}}
\toprule
Cycles               & 3, each starting from the previous cycle's selected model, with a fresh optimizer and schedule \\
Labels               & the current model translates the 40K source-language pool with nucleus sampling ($\tau{=}0.3$, top-$p{=}0.9$), at most 768 new tokens \\
Training data        & 40K (source, own translation) pairs, no gold pairs \\
Quantization, LoRA   & as Table~\ref{tab:hyperparams} \\
Peak learning rate   & $2\times10^{-4}$, cosine decay, $5\%$ warmup \\
Epochs               & 1 per cycle (312 steps) \\
Max sequence length  & $4{,}096$ tokens \\
Effective Batch size           & 128 \\
Validation           & every quarter-epoch, development \bert~of the trained direction (sampled decoding) \\
Selection            & highest development \bert{}, ties broken by BLEU. The untrained model cannot be kept in cycle 1, and later cycles keep their starting model unless a checkpoint improves on it \\
Eval decoding        & greedy, at most 256 new tokens, with the training prompt (0-shot or 1-shot) \\
\bottomrule
\end{tabular}
}
\caption{Self-distillation hyperparameters (CycleDistill-style, \llamai). Self-distill (0-shot) and self-distill (1-shot) share these settings and differ only in the prompts.}
\label{tab:hyperparams-sd}
\end{table}

\section{Recipe Comparison: Full Results}
\label{app:baselines}

Tables~\ref{tab:recipes-full-id} and~\ref{tab:recipes-full-flores} give all five metrics for the recipe comparison of \S\ref{sec:recipes} on both backbones (\qwen{} has no beam-pivot, self-distillation or few-shot runs).

\begin{table*}[tp]
\centering
\footnotesize
\setlength{\tabcolsep}{3pt}
\begin{tabular}{@{}l l *{10}{r}@{}}
\toprule
 & & \multicolumn{5}{c}{\entoxx} & \multicolumn{5}{c}{\xxtoen} \\
\cmidrule(lr){3-7}\cmidrule(l){8-12}
\textbf{System} & \textbf{Pivots} & \xxl & \xl & \chrf & BS & BLEU & \xxl & \xl & \chrf & BS & BLEU \\
\midrule
\multicolumn{12}{@{}l}{\textbf{\llamai{}, Odia}} \\
Zero-shot & -- & 20.09 & 26.43 & 7.18 & -7.06 & 1.37 & 16.80 & 27.97 & 17.78 & 7.58 & 6.35 \\
Few-shot ($k{=}5$) & -- & 19.03 & 24.85 & 13.98 & 1.33 & 4.27 & 21.10 & 34.41 & 25.56 & 31.06 & 10.32 \\

\onesft{} & -- & 42.05 & 43.33 & 25.13 & 28.87 & 10.68 & 45.24 & 64.13 & 46.58 & 56.56 & 24.34 \\
\tensft{} & -- & 64.33 & 61.05 & 40.91 & 50.32 & 20.03 & \underline{61.51} & \underline{78.21} & \underline{59.70} & \underline{69.50} & \underline{38.03} \\
\addlinespace[1.2pt]
\cptsft{} (100 pairs) & -- & 47.71 & 49.74 & 28.50 & 36.02 & 12.27 & 52.13 & 71.61 & 48.88 & 61.00 & 25.61 \\
\cptsft{} (1K pairs) & -- & 58.67 & 57.09 & 34.17 & 43.72 & 15.96 & 57.33 & 74.81 & 54.59 & 65.01 & 31.35 \\
Self-distill (0-shot) & sampled & n/a\textsuperscript{a} & n/a\textsuperscript{a} & 0.54 & n/a\textsuperscript{a} & 0.14 & 15.19 & 22.43 & 14.21 & -4.53 & 4.78 \\
Self-distill (1-shot) & sampled & n/a\textsuperscript{a} & n/a\textsuperscript{a} & 0.53 & n/a\textsuperscript{a} & 0.15 & 14.11 & 20.76 & 14.41 & 11.18 & 4.95 \\
\addlinespace[1.2pt]
\BTtechnique{} (0 pairs) & sampled & 23.10 & 29.70 & 19.39 & 20.01 & 8.05 & 22.37 & 35.78 & 27.13 & 31.66 & 11.20 \\

\BTtechnique{} (100 pairs) & sampled & 57.66 & 58.12 & 36.23 & 46.28 & 17.56 & 46.50 & 65.46 & 48.67 & 58.71 & 27.59 \\
\BTtechnique{} (1K pairs) & sampled & \underline{66.34} & \underline{64.25} & \underline{41.32} & \underline{52.18} & \underline{20.43} & 54.94 & 73.29 & 55.19 & 65.72 & 33.70 \\
\BTtechnique{} (1K pairs) & beam 5 & \textbf{72.62} & \textbf{68.67} & \textbf{45.36} & \textbf{57.02} & \textbf{22.96} & \textbf{63.04} & \textbf{78.82} & \textbf{61.30} & \textbf{70.14} & \textbf{38.69} \\
\midrule
\multicolumn{12}{@{}l}{\textbf{\llamai{}, Swahili}} \\
Zero-shot & -- & 38.83 & 49.82 & 36.77 & 39.17 & 15.61 & 57.56 & 72.52 & 44.26 & 55.30 & 22.38 \\
Few-shot ($k{=}5$) & -- & 40.27 & 51.52 & 38.02 & 39.98 & 15.95 & 61.35 & 75.23 & 46.35 & 57.89 & 23.90 \\

\onesft{} & -- & 42.12 & 52.19 & 39.10 & 41.66 & 17.08 & 60.05 & 76.08 & 46.13 & 60.37 & 25.23 \\
\tensft{} & -- & 57.38 & 66.24 & 45.27 & 49.18 & 21.45 & 66.11 & 80.20 & \underline{49.13} & \underline{63.59} & \underline{27.82} \\
\addlinespace[1.2pt]
\cptsft{} (1K pairs) & -- & 46.53 & 56.82 & 39.27 & 42.82 & 17.33 & \underline{67.16} & \textbf{81.89} & 49.07 & 63.45 & 27.45 \\
Self-distill (0-shot) & sampled & 34.39 & 45.70 & 32.64 & 33.92 & 13.05 & 42.03 & 60.20 & 38.32 & 45.12 & 17.62 \\
Self-distill (1-shot) & sampled & 33.14 & 43.78 & 32.89 & 33.19 & 13.23 & 40.61 & 57.94 & 38.19 & 44.86 & 18.00 \\
\addlinespace[1.2pt]
\BTtechnique{} (1K pairs) & sampled & \underline{62.92} & \underline{71.23} & \underline{47.17} & \underline{51.79} & \underline{23.32} & 62.65 & 77.40 & 48.68 & 61.72 & 27.34 \\
\BTtechnique{} (1K pairs) & beam 5 & \textbf{67.15} & \textbf{74.76} & \textbf{49.13} & \textbf{53.63} & \textbf{24.67} & \textbf{68.56} & \underline{81.14} & \textbf{51.07} & \textbf{64.20} & \textbf{29.11} \\
\midrule
\multicolumn{12}{@{}l}{\textbf{\qweni{}, Odia}} \\
Zero-shot & -- & 16.96 & 24.16 & 7.05 & -1.12 & 1.33 & 22.70 & 36.73 & 28.33 & 30.22 & 10.51 \\
\onesft{} & -- & 25.55 & 31.08 & 21.20 & 16.79 & 8.37 & 41.05 & 57.70 & 43.31 & 51.91 & 21.11 \\
\tensft{} & -- & \underline{66.21} & \underline{63.93} & \underline{42.69} & \underline{51.94} & \underline{20.84} & \textbf{60.77} & \textbf{76.27} & \textbf{59.37} & \textbf{68.10} & \textbf{37.53} \\
\cptsft{} (1K pairs) & -- & 32.04 & 35.82 & 25.25 & 25.50 & 10.55 & 50.20 & \underline{68.52} & 49.21 & 59.93 & 26.32 \\
\BTtechnique{} (1K pairs) & sampled & \textbf{69.73} & \textbf{67.96} & \textbf{46.51} & \textbf{56.97} & \textbf{24.86} & \underline{51.72} & 68.32 & \underline{54.86} & \underline{63.25} & \underline{34.61} \\
\midrule
\multicolumn{12}{@{}l}{\textbf{\qweni{}, Swahili}} \\
Zero-shot & -- & 17.97 & 26.45 & 20.44 & 3.22 & 7.35 & 23.87 & 37.27 & 28.61 & 30.34 & 11.28 \\

\onesft{} & -- & 21.14 & 30.73 & 26.99 & 19.50 & 10.67 & 32.94 & 49.06 & 34.77 & 44.45 & 17.05 \\
\tensft{} & -- & \textbf{47.48} & \underline{55.56} & \textbf{43.59} & \underline{45.90} & \textbf{20.56} & \textbf{57.94} & \textbf{72.23} & \textbf{46.45} & \textbf{59.00} & \textbf{25.41} \\
\cptsft{} (1K pairs) & -- & 24.38 & 33.87 & 31.06 & 30.79 & 12.28 & 43.57 & 59.89 & 39.86 & 51.78 & 20.50 \\
\BTtechnique{} (1K pairs) & sampled & \underline{46.62} & \textbf{55.61} & \underline{43.16} & \textbf{46.53} & \underline{20.19} & \underline{52.14} & \underline{68.00} & \underline{44.66} & \underline{56.31} & \underline{24.19} \\
\bottomrule
\end{tabular}
\caption{Leveraging monolingual data for low-resource MT: full results for Table~\ref{tab:recipes} on the in-domain test sets (2K), all metrics: \xxl, \xl, \chrf, \bert{} (BS) and BLEU. We compare CPT-based SFT (ALMA-style), self-distillation (CycleDistill-style), and back-translation (BT) with 100 or 1K parallel pairs plus 40K English and 40K XX monolingual sentences, using \llamai{} and \qweni{} as base models (\qwen{} has no beam-pivot, self-distillation or few-shot runs). Self-distillation uses no parallel data. \emph{Few-shot} prompts the base model, without training, with $k$ in-context examples retrieved from the 1K seed, where $k$ is the number of examples (\S\ref{sec:icl}). \emph{Pivots} indicate decoding for intermediate translations (sampled: $\tau{=}0.3$, top-$p{=}0.9$). \textbf{Bold}/\underline{underline} mark best/second-best per column per language and backbone. \tensft{} uses 10$\times$ more parallel data. Self-distill runs three cycles, and we report the best checkpoint. \textsuperscript{a}Invalid: 99\% of outputs contain no Odia script (87--90\% English, incl.\ 22\% copies, with the rest in other scripts such as Gujarati/Bengali), which XCOMET and the MuRIL-based \bert{} miss.}

\label{tab:recipes-full-id}
\end{table*}

\begin{table*}[tp]
\centering
\footnotesize
\setlength{\tabcolsep}{3pt}
\begin{tabular}{@{}l l *{10}{r}@{}}
\toprule
 & & \multicolumn{5}{c}{\entoxx} & \multicolumn{5}{c}{\xxtoen} \\
\cmidrule(lr){3-7}\cmidrule(l){8-12}
\textbf{System} & \textbf{Pivots} & \xxl & \xl & \chrf & BS & BLEU & \xxl & \xl & \chrf & BS & BLEU \\
\midrule
\multicolumn{12}{@{}l}{\textbf{\llamai{}, Odia}} \\
Zero-shot & -- & 20.81 & 23.61 & 4.95 & -11.11 & 0.65 & 16.96 & 24.76 & 19.41 & 4.18 & 5.54 \\

Few-shot ($k{=}5$) & -- & 15.29 & 19.37 & 8.21 & -10.01 & 1.72 & 17.21 & 22.68 & 20.64 & 11.79 & 5.58 \\

\onesft{} & -- & 30.60 & 29.41 & 18.03 & 17.80 & 5.84 & 40.94 & 54.67 & 37.41 & 46.19 & 14.27 \\
\tensft{} & -- & 37.87 & 33.48 & 21.73 & 26.17 & 7.20 & 47.07 & 61.54 & 39.26 & 49.77 & 16.19 \\
\addlinespace[1.2pt]
\cptsft{} (100 pairs) & -- & 33.55 & 31.26 & 19.06 & 21.14 & 5.96 & 48.27 & 61.70 & 38.98 & 49.63 & 15.05 \\
\cptsft{} (1K pairs) & -- & 37.22 & 33.04 & 20.43 & 25.01 & 6.82 & \underline{51.14} & \underline{63.75} & \underline{41.40} & \underline{51.86} & \underline{16.70} \\
Self-distill (0-shot) & sampled & n/a\textsuperscript{a} & n/a\textsuperscript{a} & 1.08 & n/a\textsuperscript{a} & 0.60 & 15.27 & 20.38 & 14.21 & -7.20 & 4.16 \\
Self-distill (1-shot) & sampled & n/a\textsuperscript{a} & n/a\textsuperscript{a} & 1.07 & n/a\textsuperscript{a} & 0.50 & 14.18 & 19.98 & 15.32 & 7.26 & 4.72 \\
\addlinespace[1.2pt]

\BTtechnique{} (0 pairs) & sampled & 20.51 & 22.02 & 13.60 & 9.91 & 4.07 & 18.61 & 27.58 & 21.41 & 20.14 & 7.19 \\

\BTtechnique{} (100 pairs) & sampled & 33.95 & 32.14 & 20.66 & 24.68 & 7.10 & 29.61 & 44.77 & 29.87 & 38.03 & 11.33 \\
\BTtechnique{} (1K pairs) & sampled & \underline{39.04} & \underline{35.91} & \underline{22.25} & \underline{26.61} & \underline{7.70} & 38.87 & 54.56 & 34.92 & 45.29 & 13.79 \\
\BTtechnique{} (1K pairs) & beam 5 & \textbf{44.39} & \textbf{38.39} & \textbf{24.79} & \textbf{30.86} & \textbf{8.59} & \textbf{52.01} & \textbf{63.95} & \textbf{43.10} & \textbf{52.08} & \textbf{17.78} \\
\midrule
\multicolumn{12}{@{}l}{\textbf{\llamai{}, Swahili}} \\
Zero-shot & -- & 35.05 & 39.94 & 38.86 & 43.63 & 14.05 & 59.20 & 70.09 & 48.51 & 60.34 & 25.09 \\
Few-shot ($k{=}5$) & -- & 35.29 & 41.02 & 39.37 & 43.34 & 14.24 & 60.54 & 71.08 & \underline{48.68} & 60.60 & 24.78 \\

\onesft{} & -- & 35.41 & 40.21 & 37.97 & 41.76 & 13.61 & 56.90 & 71.21 & 46.06 & 60.55 & 23.34 \\
\tensft{} & -- & 45.54 & 50.01 & 40.56 & 46.87 & 14.94 & 58.46 & 72.20 & 46.24 & 60.70 & 23.86 \\
\addlinespace[1.2pt]
\cptsft{} (1K pairs) & -- & 38.38 & 44.49 & 36.95 & 41.76 & 13.39 & \textbf{63.90} & \textbf{77.22} & 48.66 & \textbf{63.79} & \underline{25.61} \\

Self-distill (0-shot) & sampled & 32.44 & 37.78 & 35.09 & 39.95 & 12.63 & 44.44 & 57.98 & 41.64 & 49.90 & 19.55 \\
Self-distill (1-shot) & sampled & 32.07 & 37.14 & 35.98 & 40.60 & 12.54 & 42.17 & 55.09 & 41.44 & 48.16 & 18.93 \\
\addlinespace[1.2pt]
\BTtechnique{} (1K pairs) & sampled & \underline{53.18} & \underline{56.80} & \underline{43.17} & \underline{50.67} & \underline{17.49} & 56.15 & 69.54 & 47.36 & 59.38 & 24.82 \\
\BTtechnique{} (1K pairs) & beam 5 & \textbf{56.82} & \textbf{61.38} & \textbf{45.79} & \textbf{52.51} & \textbf{18.89} & \underline{63.40} & \underline{74.88} & \textbf{50.64} & \underline{63.64} & \textbf{27.68} \\
\midrule
\multicolumn{12}{@{}l}{\textbf{\qweni{}, Odia}} \\
Zero-shot & -- & 16.90 & 20.20 & 7.37 & -0.32 & 1.60 & 24.65 & 34.49 & 29.77 & 30.94 & 9.07 \\
\onesft{} & -- & 17.96 & 21.20 & 13.69 & 3.70 & 3.78 & 30.99 & 43.55 & 30.90 & 37.34 & 11.02 \\
\tensft{} & -- & \textbf{33.47} & \textbf{31.88} & \textbf{21.46} & \underline{22.90} & \underline{7.13} & \textbf{40.23} & \textbf{53.57} & \textbf{35.51} & \underline{44.30} & \textbf{13.69} \\
\cptsft{} (1K pairs) & -- & 20.94 & 22.65 & 15.06 & 8.96 & 4.24 & \underline{38.14} & \underline{52.58} & \underline{34.18} & \textbf{44.42} & \underline{12.99} \\
\BTtechnique{} (1K pairs) & sampled & \underline{31.73} & \underline{31.70} & \underline{21.21} & \textbf{23.63} & \textbf{7.21} & 24.55 & 37.28 & 28.66 & 33.36 & 10.67 \\
\midrule
\multicolumn{12}{@{}l}{\textbf{\qweni{}, Swahili}} \\
Zero-shot & -- & 14.74 & 19.38 & 17.45 & -1.29 & 4.56 & 19.78 & 29.89 & 27.23 & 27.43 & 9.78 \\
\onesft{} & -- & 14.50 & 19.85 & 21.82 & 12.01 & 6.26 & 23.55 & 36.18 & 30.33 & 36.29 & 12.09 \\
\tensft{} & -- & \underline{32.10} & \underline{36.18} & \underline{36.74} & \underline{40.51} & \underline{12.86} & \textbf{42.81} & \textbf{57.22} & \textbf{39.74} & \textbf{49.97} & \underline{18.19} \\
\cptsft{} (1K pairs) & -- & 16.23 & 22.00 & 26.28 & 25.48 & 7.42 & 31.71 & 45.19 & 34.90 & 43.92 & 15.58 \\
\BTtechnique{} (1K pairs) & sampled & \textbf{33.53} & \textbf{38.94} & \textbf{37.26} & \textbf{42.96} & \textbf{13.74} & \underline{38.25} & \underline{54.34} & \underline{39.62} & \underline{48.71} & \textbf{18.36} \\
\bottomrule
\end{tabular}
\caption{Recipe comparison on Flores-200, all metrics. Notation as in Table~\ref{tab:recipes-full-id}.}
\label{tab:recipes-full-flores}
\end{table*}

\section{Interleaving Ratio $r$: SFT vs.\ BT Steps}
\label{app:ratio}
This section studies the interleaving schedule of \S\ref{sec:method}, in which each supervised step is followed by $r$ back-translation steps in the same direction. We compare $r\in\{1,2,4,8\}$ on the held-out (2K) test set for Odia and Sanskrit in both directions (Table~\ref{tab:ratio-all}). Quality follows an inverted U in $r$. Odia peaks at $r{=}2$, which is the best ratio on all ten metric and direction columns. Sanskrit peaks at $r{=}4$, which is the best ratio on seven of ten. $r{=}8$ is below each language's best ratio on every column, and $r{=}1$ on all but one. We use $r{=}4$ throughout: Sanskrit peaks there, and it is within 2.7 points of the best ratio on every Odia column. Tuning $r$ per language could add further gains, for example $2.70$ \xxl{} points into Odia at $r{=}2$.

\begin{table*}[tp]
\centering
{\small
\setlength{\tabcolsep}{5pt}
\begin{tabular}{@{}l ccccc@{}}
\toprule
Configuration & \xxl~& \xl~& \chrf~& \bert~F1 & BLEU \\
\midrule
\multicolumn{6}{@{}l}{\textbf{Sanskrit}} \\
\addlinespace[0.2em]
\multicolumn{6}{@{}l}{\quad\textit{\entoxx}} \\
\addlinespace[0.1em]
% Zero-Shot & 21.72 & 25.16 & 10.91 & -3.56 & 0.94 \\
% \onesft~ & 24.21 & 26.21 & 14.57 & 11.14 & 6.38 \\
Interleave $r=1$ & 26.97 & 28.03 & 16.30 & \underline{16.47} & \textbf{8.63} \\
Interleave $r=2$ & \textbf{28.25} & \textbf{28.31} & \underline{17.92} & 15.92 & \underline{8.44} \\
Interleave $r=4$ (default) & \underline{28.07} & \underline{28.23} & \textbf{18.09} & \textbf{17.61} & 8.32 \\
Interleave $r=8$ & 26.85 & 27.60 & 16.63 & 13.95 & 7.78 \\
% \cmidrule(lr){1-6}
% \tensft~ & 27.43 & 27.71 & 16.44 & 13.20 & 7.76 \\
\addlinespace[0.35em]
\multicolumn{6}{@{}l}{\quad\textit{\xxtoen}} \\
\addlinespace[0.1em]
% Zero-Shot & 22.07 & 25.75 & 18.52 & 7.01 & 5.08 \\
% \onesft~ & 23.60 & 27.55 & 20.72 & 13.32 & 5.92 \\
Interleave $r=1$ & 23.70 & 26.24 & 22.66 & 10.94 & 5.99 \\
Interleave $r=2$ & \underline{25.10} & \underline{28.22} & \underline{22.86} & \underline{13.21} & \underline{6.58} \\
Interleave $r=4$ (default) & \textbf{25.67} & \textbf{28.57} & \textbf{23.33} & \textbf{13.93} & \textbf{6.66} \\
Interleave $r=8$ & 24.69 & 27.92 & 21.71 & 12.02 & 6.55 \\
% \cmidrule(lr){1-6}
% \tensft~ & \textbf{29.09} & \textbf{32.35} & \textbf{23.75} & \textbf{20.77} & \textbf{7.49} \\
\midrule
\multicolumn{6}{@{}l}{\textbf{Odia}} \\
\addlinespace[0.2em]
\multicolumn{6}{@{}l}{\quad\textit{\entoxx}} \\
\addlinespace[0.1em]
% Zero-Shot & 20.09 & 26.43 & 7.18 & -7.06 & 1.37 \\
% \onesft~ & 42.05 & 43.33 & 25.13 & 28.87 & 10.68 \\
Interleave $r=1$ & \underline{66.51} & 63.79 & \underline{42.18} & \underline{52.45} & \underline{21.13} \\
Interleave $r=2$ & \textbf{69.04} & \textbf{66.72} & \textbf{43.37} & \textbf{54.58} & \textbf{22.46} \\
Interleave $r=4$ (default) & 66.34 & \underline{64.25} & 41.32 & 52.18 & 20.43 \\
Interleave $r=8$ & 63.72 & 62.74 & 39.57 & 50.69 & 20.01 \\
% \cmidrule(lr){1-6}
% \tensft~ & 64.33 & 61.05 & 40.91 & 50.32 & 20.03 \\
\addlinespace[0.35em]
\multicolumn{6}{@{}l}{\quad\textit{\xxtoen}} \\
\addlinespace[0.1em]
% Zero-Shot & 16.80 & 27.97 & 17.78 & 7.58 & 6.35 \\
% \onesft~ & 45.24 & 64.13 & 46.58 & 56.56 & 24.34 \\
Interleave $r=1$ & 54.89 & \underline{73.40} & \underline{55.88} & \underline{66.10} & \underline{34.68} \\
Interleave $r=2$ & \textbf{56.85} & \textbf{73.87} & \textbf{57.05} & \textbf{66.34} & \textbf{34.97} \\
Interleave $r=4$ (default) & \underline{54.94} & 73.29 & 55.19 & 65.72 & 33.70 \\
Interleave $r=8$ & 52.21 & 70.39 & 54.22 & 63.91 & 33.13 \\
% \cmidrule(lr){1-6}
% \tensft~ & \textbf{61.51} & \textbf{78.21} & \textbf{59.70} & \textbf{69.50} & \textbf{38.03} \\
\bottomrule
\end{tabular}
}
\caption{Interleaving ratio $r$ (\llamai, held-out 2K, sampled pivots). One supervised step is followed by $r$ back-translation steps in the same direction. The main experiments use $r{=}4$. Odia (top) and Sanskrit (bottom), \entoxx~above \xxtoen. Bold and underline denote the best and second per metric within each language and direction.}
\label{tab:ratio-all}
\end{table*}

\section{In-Context Learning}
\label{app:icl}
We also ask whether the same 1K seed can be used more effectively at no annotation cost. With in-context learning (ICL), we prepend up to $k{=}5$ semantically similar examples, retrieved from the 1K seed, to each training and test sentence (Appendix~\ref{app:impl}). Examples are dropped when the prompt would exceed 2,048 tokens. This happens mainly on Odia Flores-200 into English, where prompts contain 3.6 examples on average. The retrieved examples are used consistently in supervised and back-translation updates and at inference time.

Table~\ref{tab:icl-full} gives all five metrics for the systems of Table~\ref{tab:icl}, for Odia and Amharic, in domain and on Flores-200. In domain, \onesft{} with ICL outperforms \onesft{} under the same budget ($+16.7\%$ \xxl{} \textsc{EN$\rightarrow$OR} and $+9.2\%$ \textsc{OR$\rightarrow$EN} for Odia), and \BTtechnique{} with ICL improves over \BTtechnique{} (both with sampled pivots) on every metric in both directions for both languages. \cptsft{} gains less from ICL (1.6 and 1.3 \xxl{} points on Odia). On Odia Flores-200, ICL changes the \xxl{} of \onesft{} and \cptsft{} by less than one point, and \BTtechnique{} with ICL gains 3.9 points into English and none into Odia. On Amharic Flores-200, ICL changes the \xxl{} of both systems by less than one point, and \BTtechnique{} with ICL is below \BTtechnique{} on all five metrics into Amharic and on four of the five into English. With beam-decoded pivots (Odia only), \BTtechnique{} with ICL improves over \BTtechnique{} on every in-domain metric in both directions (3.19 and 2.33 \xxl{} points). On Odia Flores-200, it improves all five metrics into Odia (4.73 \xxl{} points), and into English it is below \BTtechnique{} on three of the five metrics, by at most 0.56 points.

\begin{table*}[tp]
\centering
\footnotesize
\setlength{\tabcolsep}{3pt}
\begin{tabular}{@{}l c c *{10}{r}@{}}
\toprule
 & & & \multicolumn{5}{c}{\entoxx} & \multicolumn{5}{c}{\xxtoen} \\
\cmidrule(lr){4-8}\cmidrule(l){9-13}
\textbf{System} & \textbf{Pivots} & \textbf{ICL} & \xxl & \xl & \chrf & BS & BLEU & \xxl & \xl & \chrf & BS & BLEU \\
\midrule
\multicolumn{13}{@{}l}{\textbf{Odia, in-domain (2K)}} \\
\onesft{} & -- & No & 42.05 & 43.33 & 25.13 & 28.87 & 10.68 & 45.24 & 64.13 & 46.58 & 56.56 & 24.34 \\
\onesft{} & -- & Yes & 49.09 & 49.12 & 29.18 & 36.08 & 13.32 & 49.40 & 67.06 & 49.40 & 59.44 & 25.89 \\
\cptsft{} & -- & No & 58.67 & 57.09 & 34.17 & 43.72 & 15.96 & 57.33 & 74.81 & 54.59 & 65.01 & 31.35 \\
\cptsft{} & -- & Yes & 60.28 & 58.94 & 35.03 & 45.42 & 16.45 & 58.62 & 76.78 & 55.19 & 66.70 & 31.93 \\
\BTtechnique{} & sampled & No & 66.34 & 64.25 & 41.32 & 52.18 & 20.43 & 54.94 & 73.29 & 55.19 & 65.72 & 33.70 \\
\BTtechnique{} & sampled & Yes & 69.63 & 66.78 & 43.76 & 55.24 & 22.24 & 60.72 & 77.84 & 59.87 & 69.27 & 37.71 \\
\BTtechnique{} & beam 5 & No & \underline{72.62} & \underline{68.67} & \underline{45.36} & \underline{57.02} & \underline{22.96} & \underline{63.04} & \underline{78.82} & \underline{61.30} & \underline{70.14} & \underline{38.69} \\
\BTtechnique{} & beam 5 & Yes & \textbf{75.81} & \textbf{72.05} & \textbf{46.65} & \textbf{59.18} & \textbf{24.03} & \textbf{65.37} & \textbf{80.69} & \textbf{62.33} & \textbf{71.48} & \textbf{39.68} \\
\midrule
\multicolumn{13}{@{}l}{\textbf{Odia, Flores-200}} \\
\onesft{} & -- & No & 30.60 & 29.41 & 18.03 & 17.80 & 5.84 & 40.94 & 54.67 & 37.41 & 46.19 & 14.27 \\
\onesft{} & -- & Yes & 31.36 & 29.38 & 18.50 & 19.41 & 5.92 & 40.82 & 53.57 & 38.18 & 45.80 & 14.30 \\
\cptsft{} & -- & No & 37.22 & 33.04 & 20.43 & 25.01 & 6.82 & 51.14 & 63.75 & 41.40 & 51.86 & 16.70 \\
\cptsft{} & -- & Yes & 37.50 & 33.87 & 21.06 & 25.91 & 6.99 & 51.01 & \textbf{64.70} & 41.33 & \textbf{52.82} & 16.88 \\
\BTtechnique{} & sampled & No & 39.04 & 35.91 & 22.25 & 26.61 & 7.70 & 38.87 & 54.56 & 34.92 & 45.29 & 13.79 \\
\BTtechnique{} & sampled & Yes & 39.05 & 35.34 & 22.92 & 27.42 & 7.70 & 42.75 & 57.19 & 37.84 & 46.56 & 14.64 \\
\BTtechnique{} & beam 5 & No & \underline{44.39} & \underline{38.39} & \underline{24.79} & \underline{30.86} & \underline{8.59} & \textbf{52.01} & 63.95 & \textbf{43.10} & 52.08 & \textbf{17.78} \\
\BTtechnique{} & beam 5 & Yes & \textbf{49.12} & \textbf{41.46} & \textbf{26.66} & \textbf{35.00} & \textbf{9.26} & \underline{51.45} & \underline{64.26} & \underline{42.55} & \underline{52.79} & \underline{17.72} \\
\midrule
\multicolumn{13}{@{}l}{\textbf{Amharic, in-domain}} \\
\onesft{} & -- & No & 14.71 & 22.25 & 6.26 & -6.75 & 2.12 & 19.37 & 29.20 & 21.92 & 25.98 & 7.97 \\
\onesft{} & -- & Yes & 16.11 & 24.56 & 8.94 & 4.84 & 4.21 & \underline{21.41} & \underline{33.92} & \underline{24.58} & \textbf{29.72} & \underline{8.80} \\
\BTtechnique{} & sampled & No & \underline{19.34} & \underline{24.74} & \underline{10.27} & \underline{15.13} & \underline{4.40} & 20.18 & 31.21 & 23.57 & 27.05 & 8.10 \\
\BTtechnique{} & sampled & Yes & \textbf{20.63} & \textbf{26.75} & \textbf{12.13} & \textbf{18.88} & \textbf{5.45} & \textbf{21.97} & \textbf{34.36} & \textbf{25.56} & \underline{29.23} & \textbf{8.84} \\
\midrule
\multicolumn{13}{@{}l}{\textbf{Amharic, Flores-200}} \\
\onesft{} & -- & No & 13.12 & 20.18 & 4.38 & -11.54 & 1.29 & 15.19 & 21.23 & 19.63 & \underline{17.54} & \underline{6.65} \\
\onesft{} & -- & Yes & 13.36 & 20.24 & 4.98 & -6.05 & 2.02 & 15.30 & 21.42 & 20.00 & 17.08 & 6.52 \\
\BTtechnique{} & sampled & No & \textbf{16.95} & \textbf{20.60} & \textbf{6.71} & \textbf{7.36} & \textbf{2.90} & \underline{15.74} & \textbf{22.95} & \textbf{20.55} & \textbf{17.96} & \textbf{6.66} \\
\BTtechnique{} & sampled & Yes & \underline{16.13} & \underline{20.26} & \underline{5.83} & \underline{1.40} & \underline{2.12} & \textbf{15.88} & \underline{22.26} & \underline{20.41} & 17.47 & 6.51 \\

\bottomrule
\end{tabular}
\caption{In-context learning, all metrics (\llamai): \xxl, \xl, \chrf, \bert{} (BS) and BLEU. ICL applies $k{=}5$ demonstrations retrieved from the 1K seed in every training and test prompt. \emph{Pivots}: decoding of the intermediate translations (sampled: $\tau{=}0.3$, top-$p{=}0.9$; beam 5: beam search with five beams), -- for systems without back-translation. Amharic has no run with beam pivots. The Amharic in-domain test set has 486 sentences, and Amharic has no \cptsft{} run.  \textbf{Bold} and \underline{underline}: best and second-best system per column within each block. Ties at two decimals are marked alike.}
\label{tab:icl-full}
\end{table*}

\section{Monolingual Source: Full Results}
\label{app:datasrc}
This section completes the monolingual-source comparison of \S\ref{sec:analysis-domain} (Table~\ref{tab:bt-source}, top) with all five metrics (Table~\ref{tab:btsrc-all}). The parallel data is fixed at 1K pairs from VaqasAI (Odia) or Samanantar (Assamese), only the source of the 40K monolingual sentences per language changes, and each language is scored on its own held-out test set. The in-corpus monolingual text is best on every metric in both directions for both languages. Into the low-resource language, every other source also improves over \onesft{} on all five metrics. Into English, the other sources give mixed results: CC-100 falls below \onesft{} on all five metrics for Odia, and only IndicCorp-2.0 with WikiMatrix (Odia) improves on all five.

\begin{table*}[tp]
\centering
\footnotesize
\setlength{\tabcolsep}{3pt}
\begin{tabular}{@{}l *{10}{r}@{}}
\toprule
 & \multicolumn{5}{c}{\entoxx} & \multicolumn{5}{c}{\xxtoen} \\
\cmidrule(lr){2-6}\cmidrule(l){7-11}
\textbf{BT source (\textsc{xx} / \textsc{en})} & \xxl & \xl & \chrf & BS & BLEU & \xxl & \xl & \chrf & BS & BLEU \\
\midrule
\multicolumn{11}{@{}l}{\textbf{Odia} \emph{(seed: VaqasAI)}} \\
\emph{\onesft{} (no BT)} & \emph{42.05} & \emph{43.33} & \emph{25.13} & \emph{28.87} & \emph{10.68} & \emph{45.24} & \emph{64.13} & \emph{46.58} & \emph{56.56} & \emph{24.34} \\
VaqasAI & \textbf{66.34} & \textbf{64.25} & \textbf{41.32} & \textbf{52.18} & \textbf{20.43} & \textbf{54.94} & \textbf{73.29} & \textbf{55.19} & \textbf{65.72} & \textbf{33.70} \\
OdiEnCorp & 56.38 & 55.03 & 33.20 & 41.54 & 15.23 & 46.04 & 63.90 & 46.63 & 55.46 & 24.40 \\
IndicCorp / CC-100 & \underline{62.21} & \underline{59.96} & \underline{37.83} & \underline{47.98} & \underline{17.56} & 46.30 & 65.26 & 45.21 & 55.94 & 24.03 \\
CC-100 & 53.44 & 54.48 & 34.06 & 42.12 & 15.96 & 40.96 & 59.75 & 42.59 & 51.88 & 22.44 \\
IndicCorp / WikiMatrix & 51.82 & 51.53 & 32.64 & 40.03 & 14.31 & \underline{47.70} & \underline{67.16} & \underline{46.84} & \underline{57.95} & \underline{24.74} \\
\midrule
\multicolumn{11}{@{}l}{\textbf{Assamese} \emph{(seed: Samanantar)}} \\
\emph{\onesft{} (no BT)} & \emph{36.59} & \emph{41.92} & \emph{18.99} & \emph{24.65} & \emph{9.74} & \emph{36.49} & \emph{56.40} & \emph{27.84} & \emph{42.12} & \emph{15.97} \\
Samanantar & \textbf{46.41} & \textbf{49.95} & \textbf{27.02} & \textbf{35.76} & \textbf{16.01} & \textbf{39.69} & \textbf{59.63} & \textbf{33.54} & \textbf{45.38} & \textbf{19.98} \\
CC-100 & \underline{38.61} & \underline{43.29} & \underline{21.73} & 28.23 & \underline{12.26} & \underline{36.63} & \underline{55.69} & \underline{28.29} & \underline{39.70} & \underline{15.62} \\
IndicCorp / CC-100 & 38.11 & 42.16 & 21.24 & \underline{28.26} & 11.39 & 35.48 & 54.33 & 27.95 & 38.26 & 15.25 \\
\bottomrule
\end{tabular}
\caption{Back-translation source sweep, all metrics (\llamai, held-out 2K, sampled pivots, fixed 1K seed): \xxl, \xl, \chrf, \bert{} (BS) and BLEU. VaqasAI and Samanantar are in-domain, OdiEnCorp is another domain, and the rest are web text. One name means both sides come from it. \textbf{Bold} and \underline{underline}: best and second-best back-translation source per column and language. \onesft{} (italic) is the supervised reference and is not ranked.}
\label{tab:btsrc-all}
\end{table*}

\section{Monolingual Text Quality versus Domain: Full Results}
\label{app:xdomain}

This section completes the quality-versus-domain comparison of \S\ref{sec:analysis-domain} (Table~\ref{tab:bt-source}, bottom). We cross the Odia parallel corpus with the 40K + 40K monolingual corpus over VaqasAI and OdiEnCorp-2.0, and use the monolingual corpus either as the back-translation source of \BTtechnique{} or for the continued pretraining of \cptsft{}. \onesft{} and \tensft{} are included as references. Table~\ref{tab:domain-all} evaluates both directions on the held-out test set of each corpus and on Flores-200. With the parallel data fixed, VaqasAI is the better back-translation source on \xxl{} and \xl{} in all eight held-out settings, including on the OdiEnCorp-2.0 test set itself. The \xxl{} margins on that test set are small (0.6 to 2.5 points) and have not been tested for significance, and there \chrf{} favours OdiEnCorp-2.0 back-translation in all four settings, and \bert{} and BLEU in three of four. On Flores-200, VaqasAI back-translation is ahead on every metric for both parallel corpora and both directions. For back-translation, domain match alone therefore does not explain the results, which suggests that corpus quality matters at least as much as domain. Continued pretraining behaves differently: with VaqasAI parallel data, continued pretraining on OdiEnCorp-2.0 beats continued pretraining on VaqasAI on every metric on the OdiEnCorp-2.0 test set and on Flores-200.

\begin{table*}[tp]
\centering
\footnotesize
\setlength{\tabcolsep}{3pt}
\begin{tabular}{@{}l l *{10}{r}@{}}
\toprule
 & & \multicolumn{5}{c}{\entoxx} & \multicolumn{5}{c}{\xxtoen} \\
\cmidrule(lr){3-7}\cmidrule(l){8-12}
\textbf{System} & \textbf{Monolingual} & \xxl & \xl & \chrf & BS & BLEU & \xxl & \xl & \chrf & BS & BLEU \\
\midrule
\multicolumn{12}{@{}l}{\textbf{VaqasAI test (2K)}} \\
\multicolumn{12}{@{}l}{\quad\textit{Parallel data from VaqasAI}} \\
\onesft{} & -- & 42.05 & 43.33 & 25.13 & 28.87 & 10.68 & 45.24 & 64.13 & 46.58 & 56.56 & 24.34 \\
\tensft{} & -- & \underline{64.33} & \underline{61.05} & \underline{40.91} & \underline{50.32} & \underline{20.03} & \textbf{61.51} & \textbf{78.21} & \textbf{59.70} & \textbf{69.50} & \textbf{38.03} \\
\cptsft{} & VaqasAI & 58.67 & 57.09 & 34.17 & 43.72 & 15.96 & \underline{57.33} & \underline{74.81} & 54.59 & 65.01 & 31.35 \\
\cptsft{} & OdiEnCorp & 58.68 & 56.56 & 33.03 & 42.80 & 15.02 & 55.39 & 72.92 & 52.40 & 62.87 & 28.64 \\
\BTtechnique{} & VaqasAI & \textbf{66.34} & \textbf{64.25} & \textbf{41.32} & \textbf{52.18} & \textbf{20.43} & 54.94 & 73.29 & \underline{55.19} & \underline{65.72} & \underline{33.70} \\
\BTtechnique{} & OdiEnCorp & 56.38 & 55.03 & 33.20 & 41.54 & 15.23 & 46.04 & 63.90 & 46.63 & 55.46 & 24.40 \\
\addlinespace[1.2pt]
\multicolumn{12}{@{}l}{\quad\textit{Parallel data from OdiEnCorp-2.0}} \\
\onesft{} & -- & 28.50 & 32.20 & 18.34 & 15.22 & 6.68 & 28.86 & 46.11 & 30.31 & 37.41 & 13.49 \\
\tensft{} & -- & \underline{49.23} & \underline{48.72} & \underline{25.86} & \underline{31.65} & \underline{10.55} & 36.65 & 56.38 & 35.26 & 45.60 & 16.14 \\
\cptsft{} & VaqasAI & 45.45 & 47.12 & 25.52 & 31.03 & 10.49 & \textbf{44.96} & \textbf{66.68} & \underline{41.85} & \underline{55.45} & \underline{21.12} \\
\cptsft{} & OdiEnCorp & 43.33 & 44.78 & 23.72 & 28.83 & 9.53 & 39.47 & 59.26 & 36.97 & 47.92 & 17.40 \\
\BTtechnique{} & VaqasAI & \textbf{56.32} & \textbf{56.88} & \textbf{35.52} & \textbf{44.45} & \textbf{16.80} & \underline{44.94} & \underline{64.76} & \textbf{47.16} & \textbf{57.74} & \textbf{26.48} \\
\BTtechnique{} & OdiEnCorp & 38.82 & 43.00 & 23.82 & 27.69 & 10.25 & 25.63 & 41.52 & 28.69 & 33.28 & 12.94 \\
\midrule
\multicolumn{12}{@{}l}{\textbf{OdiEnCorp-2.0 test (2K)}} \\
\multicolumn{12}{@{}l}{\quad\textit{Parallel data from VaqasAI}} \\
\onesft{} & -- & 23.97 & 24.19 & 14.98 & 10.60 & 3.75 & 30.85 & 38.23 & 28.01 & 25.59 & 9.52 \\
\tensft{} & -- & 29.50 & 27.57 & 17.80 & 17.28 & 5.43 & 35.17 & 43.57 & 29.51 & \underline{29.74} & \underline{11.92} \\
\cptsft{} & VaqasAI & 28.70 & 27.78 & 17.09 & 16.20 & 5.11 & \underline{37.02} & \underline{44.86} & \underline{30.40} & 29.49 & 11.13 \\
\cptsft{} & OdiEnCorp & \textbf{30.38} & \underline{28.90} & 17.99 & \underline{19.72} & 5.51 & \textbf{47.16} & \textbf{53.60} & \textbf{34.50} & \textbf{38.43} & \textbf{12.37} \\
\BTtechnique{} & VaqasAI & \underline{30.37} & \textbf{29.27} & \underline{18.35} & 17.37 & \underline{5.64} & 30.94 & 39.83 & 26.83 & 25.58 & 10.32 \\
\BTtechnique{} & OdiEnCorp & 27.88 & 26.72 & \textbf{19.47} & \textbf{21.41} & \textbf{6.04} & 30.28 & 36.04 & 29.86 & 25.88 & 10.60 \\
\addlinespace[1.2pt]
\multicolumn{12}{@{}l}{\quad\textit{Parallel data from OdiEnCorp-2.0}} \\
\onesft{} & -- & 21.75 & 22.24 & 15.03 & 10.01 & 3.90 & 27.95 & 34.23 & 28.31 & 24.66 & 9.43 \\
\tensft{} & -- & \textbf{33.97} & \textbf{30.28} & \textbf{22.90} & \textbf{28.63} & \textbf{7.16} & \textbf{35.71} & \textbf{42.32} & \textbf{33.43} & \textbf{33.96} & \textbf{12.47} \\
\cptsft{} & VaqasAI & 29.04 & 27.33 & 19.39 & 23.08 & 5.84 & 33.80 & 41.12 & 30.56 & 30.26 & 10.95 \\
\cptsft{} & OdiEnCorp & 29.30 & 27.65 & 20.30 & \underline{25.18} & 6.25 & \underline{34.36} & \underline{41.13} & \underline{31.57} & \underline{31.04} & \underline{11.69} \\
\BTtechnique{} & VaqasAI & \underline{30.19} & \underline{28.68} & 20.58 & 24.91 & 6.43 & 28.71 & 36.28 & 28.24 & 27.76 & 10.37 \\
\BTtechnique{} & OdiEnCorp & 29.47 & 27.45 & \underline{20.97} & 25.07 & \underline{6.69} & 28.07 & 34.88 & 28.67 & 26.58 & 10.20 \\
\midrule
\multicolumn{12}{@{}l}{\textbf{Flores-200}} \\
\multicolumn{12}{@{}l}{\quad\textit{Parallel data from VaqasAI}} \\
\onesft{} & -- & 30.60 & 29.41 & 18.03 & 17.80 & 5.84 & 40.94 & 54.67 & 37.41 & 46.19 & 14.27 \\
\tensft{} & -- & 37.87 & 33.48 & \underline{21.73} & 26.17 & \underline{7.20} & 47.07 & 61.54 & 39.26 & 49.77 & 16.19 \\
\cptsft{} & VaqasAI & 37.22 & 33.04 & 20.43 & 25.01 & 6.82 & \underline{51.14} & \underline{63.75} & \underline{41.40} & \underline{51.86} & \underline{16.70} \\
\cptsft{} & OdiEnCorp & \textbf{41.28} & \textbf{36.35} & 21.01 & \textbf{27.58} & 7.07 & \textbf{54.01} & \textbf{66.00} & \textbf{43.01} & \textbf{53.12} & \textbf{17.75} \\
\BTtechnique{} & VaqasAI & \underline{39.04} & \underline{35.91} & \textbf{22.25} & \underline{26.61} & \textbf{7.70} & 38.87 & 54.56 & 34.92 & 45.29 & 13.79 \\
\BTtechnique{} & OdiEnCorp & 31.70 & 30.30 & 19.87 & 22.42 & 6.88 & 33.11 & 47.30 & 33.26 & 38.56 & 12.10 \\
\addlinespace[1.2pt]
\multicolumn{12}{@{}l}{\quad\textit{Parallel data from OdiEnCorp-2.0}} \\
\onesft{} & -- & 23.96 & 24.67 & 15.17 & 11.90 & 4.63 & 29.86 & 43.92 & 29.79 & 33.09 & 11.13 \\
\tensft{} & -- & \textbf{38.47} & \textbf{34.38} & \underline{20.72} & \underline{25.67} & \underline{7.13} & 39.14 & 55.19 & 34.63 & 43.43 & 13.08 \\
\cptsft{} & VaqasAI & 32.95 & 31.19 & 18.54 & 22.86 & 6.04 & \textbf{43.15} & \textbf{60.12} & \textbf{36.06} & \textbf{47.44} & \textbf{14.67} \\
\cptsft{} & OdiEnCorp & 33.62 & 31.96 & 18.48 & 23.39 & 6.29 & \underline{42.06} & \underline{58.06} & \underline{35.89} & \underline{45.66} & \underline{14.29} \\
\BTtechnique{} & VaqasAI & \underline{36.61} & \underline{34.22} & \textbf{21.42} & \textbf{26.30} & \textbf{7.46} & 27.84 & 43.11 & 28.54 & 33.98 & 10.87 \\
\BTtechnique{} & OdiEnCorp & 29.20 & 28.78 & 18.75 & 20.87 & 6.43 & 21.80 & 34.30 & 25.19 & 26.13 & 8.82 \\
\bottomrule
\end{tabular}
\caption{Cross-domain transfer (Odia, \llamai, sampled pivots), all metrics: \xxl, \xl, \chrf, \bert{} (BS) and BLEU. The parallel data (1K pairs, or 10K for \tensft{}) and the 40K + 40K monolingual sentences each come from VaqasAI or OdiEnCorp-2.0. The monolingual sentences are used for continued pretraining (\cptsft{}) or as the back-translation source (\BTtechnique{}). \textbf{Bold} and \underline{underline}: best and second-best system per column within each block. Models trained on different corpora sit on different score scales, so compare rows within a test set.}
\label{tab:domain-all}
\end{table*}

\section{Scaling Parallel and Monolingual Data}
\label{app:scaling}

This section gives the full per-metric results behind \S\ref{subsec:mono-scaling}, and Figure~\ref{fig:mono-scaling} plots the two primary metrics with 1K parallel pairs. Using \llamai{} on Odia, we train \BTtechnique{} with 100 parallel pairs and 40K monolingual sentences per language, with 1K pairs and 20K, 40K or 80K, and with 10K pairs and 40K or 80K. For reference, we fine-tune on 1K, 10K and 50K pairs without monolingual data, train \cptsft{} with the same 100 pairs, and include the zero-shot model. We evaluate on the held-out (Table~\ref{tab:scaling-heldout}) and Flores-200 (Table~\ref{tab:scaling-flores}) test sets.

\pgfplotsset{
  scalingaxis/.style={
    width=0.54\columnwidth, height=4.3cm,
    xmin=-6, xmax=86, xtick={0,20,40,80},
    xticklabels={0,20K,40K,80K},
    ymin=20, ymax=72, ytick={20,30,40,50,60,70},
    tick label style={font=\scriptsize},
    label style={font=\scriptsize},
    title style={font=\scriptsize, yshift=-2pt},
    legend style={font=\scriptsize, draw=none, fill=none,
                  cells={anchor=west}, at={(0.97,0.03)}, anchor=south east},
    grid=major, grid style={dotted, gray!40},
    every axis plot/.append style={line width=0.7pt, mark size=1.6pt},
  }
}
\begin{figure}[t]
\centering
\begin{tikzpicture}
\begin{groupplot}[
  group style={group size=2 by 1, horizontal sep=6pt},
  scalingaxis,
  scale only axis,
  width=0.40\columnwidth, height=0.44\columnwidth,
  xtick={0,20,40,80},
  enlarge x limits=0.14, enlarge y limits=0.16,
  clip=false,
  tick label style={font=\scriptsize},
  label style={font=\scriptsize},
  title style={font=\small\itshape, yshift=-2pt},
  ylabel style={yshift=-3pt},
  legend style={font=\scriptsize, draw=none, fill=none,
                at={(0.97,0.04)}, anchor=south east,
                inner sep=1pt, /tikz/every even column/.append style={column sep=2pt}},
  legend cell align=left,
  every node near coord/.append style={font=\tiny},
]
% ---------------- en -> xx ----------------
\nextgroupplot[title={\entoxx}, ylabel={Score}]
\addplot[color=black!70!blue, mark=*, mark size=1.5pt, thick,
  nodes near coords={\pgfmathprintnumber[fixed, fixed zerofill, precision=1]{\pgfplotspointmeta}},
  every node near coord/.append style={anchor=south, yshift=2pt, text=black!70!blue}]
  coordinates {(0,42.05) (20,61.00) (40,66.34) (80,67.90)};
\addplot[color=black!55!red, mark=square*, mark size=1.3pt, thick, dashed,
  nodes near coords={\pgfmathprintnumber[fixed, fixed zerofill, precision=1]{\pgfplotspointmeta}},
  every node near coord/.append style={anchor=north, yshift=-2pt, text=black!55!red}]
  coordinates {(0,25.13) (20,36.65) (40,41.32) (80,42.79)};
% ---------------- xx -> en ----------------
\nextgroupplot[title={\xxtoen}, yticklabels={}]
\addplot[color=black!70!blue, mark=*, mark size=1.5pt, thick,
  nodes near coords={\pgfmathprintnumber[fixed, fixed zerofill, precision=1]{\pgfplotspointmeta}},
  every node near coord/.append style={anchor=north, yshift=-2pt, text=black!70!blue}]
  coordinates {(0,45.24) (20,51.16) (40,54.94) (80,56.16)};
\addplot[color=black!55!red, mark=square*, mark size=1.3pt, thick, dashed,
  nodes near coords={\pgfmathprintnumber[fixed, fixed zerofill, precision=1]{\pgfplotspointmeta}},
  every node near coord/.append style={anchor=south, yshift=2pt, text=black!55!red}]
  coordinates {(0,46.58) (20,52.23) (40,55.19) (80,56.71)};
\legend{\xxl, \chrf}
\end{groupplot}
\end{tikzpicture}
\vspace{-4pt}
\caption{Scaling the monolingual corpus under a fixed 1K parallel seed (Odia, \llamai, held-out 2K, sampled pivots). The x-axis gives monolingual sentences per language.}
\label{fig:mono-scaling}
\vspace{-6pt}
\end{figure}

In domain, more monolingual data improves translation in both directions, with larger gains into the low-resource language than into English. The gains are steepest up to 40K monolingual sentences and flatten beyond it. For Odia \entoxx, \xxl~rises by $5.3$ points from 20K to 40K but only by $1.6$ from 40K to 80K, which motivates the 40K default. On Flores-200, \xxl{} rises from 20K to 40K and falls slightly from 40K to 80K. The gain into Odia is not confined to the 1K seed. Even with only 100 parallel pairs, 40K monolingual sentences lift \entoxx~\xxl~to $57.66$, well above the $42.05$ of \onesft{} on 1K pairs and the $47.71$ of \cptsft{} with the same 100 pairs. Into English, \cptsft{} is ahead ($52.13$ against $46.50$). Into Odia, more parallel data yields further gains ($57.66$, $66.34$ and $72.04$ \xxl{} with 100, 1K and 10K pairs at 40K). Into English, adding 40K monolingual sentences to 10K pairs gives lower \xxl{} than \tensft{} ($57.48$ against $61.51$ in domain, $35.93$ against $47.07$ on Flores-200).

\begin{table*}[tp]
\centering
\scriptsize
\setlength{\tabcolsep}{2.5pt}
\renewcommand{\arraystretch}{1.0}
\resizebox{\textwidth}{!}{%
\begin{tabular}{@{}l cc cc cc cc cc@{}}
\toprule
& \multicolumn{2}{c}{\xxl} & \multicolumn{2}{c}{\xl}
& \multicolumn{2}{c}{\chrf} & \multicolumn{2}{c}{\bert~F1}
& \multicolumn{2}{c}{BLEU} \\
\cmidrule(lr){2-3}\cmidrule(lr){4-5}\cmidrule(lr){6-7}\cmidrule(lr){8-9}\cmidrule(lr){10-11}
Configuration
& \entoxx & \xxtoen
& \entoxx & \xxtoen
& \entoxx & \xxtoen
& \entoxx & \xxtoen
& \entoxx & \xxtoen \\
\midrule
Zero-Shot        & 20.09 & 16.80 & 26.43 & 27.97 & 7.18  & 17.78 & -7.06 & 7.58  & 1.37  & 6.35 \\
\onesft~         & 42.05 & 45.24 & 43.33 & 64.13 & 25.13 & 46.58 & 28.87 & 56.56 & 10.68 & 24.34 \\
\cptsft{} (100 pairs) & 47.71 & 52.13 & 49.74 & 71.61 & 28.50 & 48.88 & 36.02 & 61.00 & 12.27 & 25.61 \\
\BTtechnique{} (100 + 40K)  & 57.66 & 46.50 & 58.12 & 65.46 & 36.23 & 48.67 & 46.28 & 58.71 & 17.56 & 27.59 \\
\BTtechnique{} (1K + 20K)   & 61.00 & 51.16 & 59.59 & 69.64 & 36.65 & 52.23 & 46.83 & 62.32 & 17.57 & 30.95 \\
\BTtechnique{} (1K + 40K)   & 66.34 & 54.94 & 64.25 & 73.29 & 41.32 & 55.19 & 52.18 & 65.72 & 20.43 & 33.70 \\
\BTtechnique{} (1K + 80K)   & 67.90 & 56.16 & 66.50 & 73.46 & 42.79 & 56.71 & 54.12 & 66.18 & 21.96 & 34.79 \\
\tensft~         & 64.33 & 61.51 & 61.05 & 78.21 & 40.91 & 59.70 & 50.32 & 69.50 & 20.03 & 38.03 \\
\BTtechnique{} (10K + 40K)  & 72.04 & 57.48 & 69.09 & 73.96 & 46.40 & 58.54 & 57.90 & 67.35 & \underline{24.81} & 37.97 \\
\BTtechnique{} (10K + 80K)  & \underline{74.17} & \underline{64.39} & \underline{70.92} & \underline{81.10}
                 & \underline{47.00} & \underline{62.59} & \underline{59.00} & \underline{72.04}
                 & 24.24 & \underline{41.37} \\
\fiftysft~       & \textbf{76.89} & \textbf{71.04} & \textbf{72.36} & \textbf{84.92}
                 & \textbf{50.36} & \textbf{68.71} & \textbf{62.13} & \textbf{77.14}
                 & \textbf{27.21} & \textbf{48.58} \\
\bottomrule
\end{tabular}%
}
\caption{Held-out (2K) scaling (Odia, \llamai, sampled pivots). \BTtechnique{} rows give gold pairs + monolingual sentences per language. \textbf{Bold} and \underline{underline}: best and second-best per column.}
\label{tab:scaling-heldout}
\end{table*}

\begin{table*}[tp]
\centering
\scriptsize
\setlength{\tabcolsep}{2.5pt}
\renewcommand{\arraystretch}{1.0}
\resizebox{\textwidth}{!}{%
\begin{tabular}{@{}l cc cc cc cc cc@{}}
\toprule
& \multicolumn{2}{c}{\xxl} & \multicolumn{2}{c}{\xl}
& \multicolumn{2}{c}{\chrf} & \multicolumn{2}{c}{\bert~F1}
& \multicolumn{2}{c}{BLEU} \\
\cmidrule(lr){2-3}\cmidrule(lr){4-5}\cmidrule(lr){6-7}\cmidrule(lr){8-9}\cmidrule(lr){10-11}
Configuration
& \entoxx & \xxtoen
& \entoxx & \xxtoen
& \entoxx & \xxtoen
& \entoxx & \xxtoen
& \entoxx & \xxtoen \\
\midrule
Zero-Shot        & 20.81 & 16.96 & 23.61 & 24.76 & 4.95  & 19.41 & -11.11 & 4.18  & 0.65 & 5.54 \\
\onesft~         & 30.60 & 40.94 & 29.41 & 54.67 & 18.03 & 37.41 & 17.80  & 46.19 & 5.84 & 14.27 \\
\cptsft{} (100 pairs) & 33.55 & \underline{48.27} & 31.26 & \underline{61.70} & 19.06 & 38.98 & 21.14 & 49.63 & 5.96 & 15.05 \\
\BTtechnique{} (100 + 40K)  & 33.95 & 29.61 & 32.14 & 44.77 & 20.66 & 29.87 & 24.68  & 38.03 & 7.10 & 11.33 \\
\BTtechnique{} (1K + 20K)   & 35.77 & 35.62 & 33.61 & 51.95 & 20.09 & 33.60 & 23.27  & 43.28 & 6.44 & 13.32 \\
\BTtechnique{} (1K + 40K)   & 39.04 & 38.87 & 35.91 & 54.56 & 22.25 & 34.92 & 26.61  & 45.29 & 7.70 & 13.79 \\
\BTtechnique{} (1K + 80K)   & 38.80 & 37.86 & 35.29 & 52.22 & 22.35 & 36.40 & 27.98  & 44.11 & 7.75 & 13.99 \\
\tensft~         & 37.87 & 47.07 & 33.48 & 61.54 & 21.73 & 39.26 & 26.17 & 49.77 & 7.20 & \underline{16.19} \\
\BTtechnique{} (10K + 40K)  & 40.96 & 35.93 & \underline{37.09} & 49.00 & 22.46 & 31.84 & 28.32 & 35.14 & 7.70 & 11.72 \\
\BTtechnique{} (10K + 80K)  & \underline{42.00} & 46.27 & 36.18 & 61.35 & \underline{23.82} & \textbf{40.03}
                 & \underline{29.13} & \underline{50.32} & \underline{7.95} & \textbf{16.71} \\
\fiftysft~       & \textbf{49.39} & \textbf{48.80} & \textbf{41.16} & \textbf{64.02}
                 & \textbf{25.29} & \underline{39.36} & \textbf{33.46} & \textbf{50.94}
                 & \textbf{8.65} & 15.77 \\
\bottomrule
\end{tabular}%
}
\caption{Flores-200 (out-of-domain) scaling (Odia, \llamai, sampled pivots). \BTtechnique{} rows give gold pairs + monolingual sentences per language. \textbf{Bold} and \underline{underline}: best and second-best per column.}
\label{tab:scaling-flores}
\end{table*}

\section{Languages and Backbones: Full Held-Out Results}
\label{app:heldout}

This section gives the full held-out results behind \S\ref{sec:crosslang} (2K test sentences per language, 486 for Amharic). Table~\ref{tab:heldout-full} reports all five metrics in the layout of Table~\ref{tab:main-crosslang}, in two blocks of three languages. The table also reports \tensft{}, trained on ten times more parallel data, as a reference. Into the low-resource language, \BTtechnique{} improves over \onesft{} on all five metrics for both backbones and all six languages. Into English, it does so in nine of the twelve language and backbone combinations; the exceptions are \llama{} Hindi (\xl{} and \bert{}, by at most 0.33), \qwen{} Sanskrit (all five metrics) and \qwen{} Amharic (\bert{} and BLEU). With one tenth of the parallel data, \BTtechnique{} exceeds \tensft{} into the low-resource language on \xxl{} and \chrf{} in eight of the ten combinations that have a \tensft{} run, so its gain over \onesft{} is the larger of the two; the exceptions are \qwen{} Sanskrit and Swahili, by at most 0.98 points. Into English, \tensft{} has the higher \xxl{} in all ten. Appendix~\ref{app:flores} gives the same comparison on Flores-200.

\begin{table*}[tp]
\centering
\tiny
\setlength{\tabcolsep}{3.0pt}
\renewcommand{\arraystretch}{1.1}
\resizebox{\textwidth}{!}{%
\begin{tabular}{@{}c c l *{15}{S[table-format=2.2]}@{}}
\toprule
& &
& \multicolumn{5}{c}{Odia} & \multicolumn{5}{c}{Assamese} & \multicolumn{5}{c}{Sanskrit} \\
\cmidrule(lr){4-8}\cmidrule(lr){9-13}\cmidrule(lr){14-18}
& & \multicolumn{1}{l}{Method}
& {X-XXL} & {\chrf} & {X-XL} & {BS} & {BLEU} & {X-XXL} & {\chrf} & {X-XL} & {BS} & {BLEU} & {X-XXL} & {\chrf} & {X-XL} & {BS} & {BLEU} \\
\midrule
\multirow{10}{*}{\rotatebox[origin=c]{90}{\textbf{LLaMA-3.2-3B}}}
& \multirow{5}{*}{\rotatebox[origin=c]{90}{\textit{\entoxx}}}
& Zero-Shot & {20.09} & {7.18} & {26.43} & {$-$7.06} & {1.37} & {38.81} & {11.61} & {45.99} & {15.71} & {3.48} & {21.72} & {10.91} & {25.16} & {$-$3.56} & {0.94} \\
& & \onesft & {42.05} & {25.13} & {43.33} & {28.87} & {10.68} & {36.59} & {18.99} & {41.92} & {24.65} & {9.74} & {24.21} & {14.57} & {26.21} & {11.14} & {6.38} \\
& & \textit{\tensft} & {\textit{64.33}} & {\textit{40.91}} & {\textit{61.05}} & {\textit{50.32}} & {\textit{20.03}} & {\textit{44.67}} & {\textit{25.77}} & {\textit{48.50}} & {\textit{34.65}} & {\textit{14.64}} & {\textit{27.43}} & {\textit{16.44}} & {\textit{27.71}} & {\textit{13.20}} & {\textit{7.76}} \\
& & \sftbt & {\gaincell{\textbf{66.34}}{\up{58}}} & {\gaincell{\textbf{41.32}}{\up{64}}} & {\gaincell{\textbf{64.25}}{\up{48}}} & {\gaincell{\textbf{52.18}}{\up{81}}} & {\gaincell{\textbf{20.43}}{\up{91}}} & {\gaincell{\textbf{46.41}}{\up{27}}} & {\gaincell{\textbf{27.02}}{\up{42}}} & {\gaincell{\textbf{49.95}}{\up{19}}} & {\gaincell{\textbf{35.76}}{\up{45}}} & {\gaincell{\textbf{16.01}}{\up{64}}} & {\gaincell{\textbf{28.07}}{\up{16}}} & {\gaincell{\textbf{18.09}}{\up{24}}} & {\gaincell{\textbf{28.23}}{\up{8}}} & {\gaincell{\textbf{17.61}}{\up{58}}} & {\gaincell{\textbf{8.32}}{\up{30}}} \\[0.5ex]
\cdashline{3-18}\addlinespace[0.6ex]
& \multirow{5}{*}{\rotatebox[origin=c]{90}{\textit{\xxtoen}}}
& Zero-Shot & {16.80} & {17.78} & {27.97} & {7.58} & {6.35} & {31.34} & {23.58} & {49.10} & {26.98} & {10.80} & {22.07} & {18.52} & {25.75} & {7.01} & {5.08} \\
& & \onesft & {45.24} & {46.58} & {64.13} & {56.56} & {24.34} & {36.49} & {27.84} & {56.40} & {42.12} & {15.97} & {23.60} & {20.72} & {27.55} & {13.32} & {5.92} \\
& & \textit{\tensft} & {\textit{61.51}} & {\textit{59.70}} & {\textit{78.21}} & {\textit{69.50}} & {\textit{38.03}} & {\textit{41.08}} & {\textit{33.36}} & {\textit{60.77}} & {\textit{46.82}} & {\textit{20.20}} & {\textit{29.09}} & {\textit{23.75}} & {\textit{32.35}} & {\textit{20.77}} & {\textit{7.49}} \\
& & \sftbt & {\gaincell{\textbf{54.94}}{\up{21}}} & {\gaincell{\textbf{55.19}}{\up{18}}} & {\gaincell{\textbf{73.29}}{\up{14}}} & {\gaincell{\textbf{65.72}}{\up{16}}} & {\gaincell{\textbf{33.70}}{\up{38}}} & {\gaincell{\textbf{39.69}}{\up{9}}} & {\gaincell{\textbf{33.54}}{\up{20}}} & {\gaincell{\textbf{59.63}}{\up{6}}} & {\gaincell{\textbf{45.38}}{\up{8}}} & {\gaincell{\textbf{19.98}}{\up{25}}} & {\gaincell{\textbf{25.67}}{\up{9}}} & {\gaincell{\textbf{23.33}}{\up{13}}} & {\gaincell{\textbf{28.57}}{\up{4}}} & {\gaincell{\textbf{13.93}}{\up{5}}} & {\gaincell{\textbf{6.66}}{\up{13}}} \\[0.5ex]
\midrule
\multirow{10}{*}{\rotatebox[origin=c]{90}{\textbf{Qwen-2.5-3B}}}
& \multirow{5}{*}{\rotatebox[origin=c]{90}{\textit{\entoxx}}}
& Zero-Shot & {16.96} & {7.05} & {24.16} & {$-$1.12} & {1.33} & {24.04} & {9.62} & {32.86} & {5.24} & {2.85} & {20.60} & {11.82} & {23.38} & {$-$0.16} & {2.06} \\
& & \onesft & {25.55} & {21.20} & {31.08} & {16.79} & {8.37} & {27.14} & {14.75} & {33.33} & {16.43} & {7.27} & {19.44} & {11.41} & {21.82} & {$-$3.90} & {4.24} \\
& & \textit{\tensft} & {\textit{66.21}} & {\textit{42.69}} & {\textit{63.93}} & {\textit{51.94}} & {\textit{20.84}} & {\textit{39.44}} & {\textit{23.11}} & {\textit{43.63}} & {\textit{30.12}} & {\textit{12.99}} & {\textit{26.89}} & {\textit{17.35}} & {\textit{26.99}} & {\textit{16.49}} & {\textit{9.17}} \\
& & \sftbt & {\gaincell{\textbf{69.73}}{\up{173}}} & {\gaincell{\textbf{46.51}}{\up{119}}} & {\gaincell{\textbf{67.96}}{\up{119}}} & {\gaincell{\textbf{56.97}}{\up{239}}} & {\gaincell{\textbf{24.86}}{\up{197}}} & {\gaincell{\textbf{41.58}}{\up{53}}} & {\gaincell{\textbf{24.12}}{\up{64}}} & {\gaincell{\textbf{46.41}}{\up{39}}} & {\gaincell{\textbf{32.26}}{\up{96}}} & {\gaincell{\textbf{14.28}}{\up{96}}} & {\gaincell{\textbf{25.91}}{\up{33}}} & {\gaincell{\textbf{16.84}}{\up{48}}} & {\gaincell{\textbf{26.62}}{\up{22}}} & {\gaincell{\textbf{14.62}}{\textcolor{gaingreen}{+19\,pt}}} & {\gaincell{\textbf{8.22}}{\up{94}}} \\[0.5ex]
\cdashline{3-18}\addlinespace[0.6ex]
& \multirow{5}{*}{\rotatebox[origin=c]{90}{\textit{\xxtoen}}}
& Zero-Shot & {22.70} & {28.33} & {36.73} & {30.22} & {10.51} & {31.08} & {24.06} & {47.52} & {28.11} & {10.11} & {24.33} & {19.56} & {28.30} & {12.57} & {5.35} \\
& & \onesft & {41.05} & {43.31} & {57.70} & {51.91} & {21.11} & {35.43} & {26.87} & {54.13} & {39.96} & {15.24} & {\textbf{25.80}} & {\textbf{21.88}} & {\textbf{28.99}} & {\textbf{15.83}} & {\textbf{6.47}} \\
& & \textit{\tensft} & {\textit{60.77}} & {\textit{59.37}} & {\textit{76.27}} & {\textit{68.10}} & {\textit{37.53}} & {\textit{38.62}} & {\textit{31.64}} & {\textit{56.94}} & {\textit{43.39}} & {\textit{18.81}} & {\textit{27.87}} & {\textit{24.52}} & {\textit{30.21}} & {\textit{19.99}} & {\textit{7.13}} \\
& & \sftbt & {\gaincell{\textbf{51.72}}{\up{26}}} & {\gaincell{\textbf{54.86}}{\up{27}}} & {\gaincell{\textbf{68.32}}{\up{18}}} & {\gaincell{\textbf{63.25}}{\up{22}}} & {\gaincell{\textbf{34.61}}{\up{64}}} & {\gaincell{\textbf{36.87}}{\up{4}}} & {\gaincell{\textbf{31.27}}{\up{16}}} & {\gaincell{\textbf{55.93}}{\up{3}}} & {\gaincell{\textbf{42.57}}{\up{7}}} & {\gaincell{\textbf{19.15}}{\up{26}}} & {\gaincell{23.77}{\down{8}}} & {\gaincell{21.74}{\down{1}}} & {\gaincell{27.09}{\down{7}}} & {\gaincell{12.04}{\down{24}}} & {\gaincell{6.46}{\textcolor{black!55}{0\%}}} \\[0.5ex]
\bottomrule
\end{tabular}%
}
\par\vspace{6pt}
\resizebox{\textwidth}{!}{%
\begin{tabular}{@{}c c l *{15}{S[table-format=2.2]}@{}}
\toprule
& &
& \multicolumn{5}{c}{Hindi} & \multicolumn{5}{c}{Amharic} & \multicolumn{5}{c}{Swahili} \\
\cmidrule(lr){4-8}\cmidrule(lr){9-13}\cmidrule(lr){14-18}
& & \multicolumn{1}{l}{Method}
& {X-XXL} & {\chrf} & {X-XL} & {BS} & {BLEU} & {X-XXL} & {\chrf} & {X-XL} & {BS} & {BLEU} & {X-XXL} & {\chrf} & {X-XL} & {BS} & {BLEU} \\
\midrule
\multirow{10}{*}{\rotatebox[origin=c]{90}{\textbf{LLaMA-3.2-3B}}}
& \multirow{5}{*}{\rotatebox[origin=c]{90}{\textit{\entoxx}}}
& Zero-Shot & {56.65} & {31.85} & {49.52} & {44.66} & {13.84} & {13.03} & {3.08} & {22.69} & {$-$13.45} & {0.26} & {38.83} & {36.77} & {49.82} & {39.17} & {15.61} \\
& & \onesft & {63.44} & {32.22} & {54.13} & {46.82} & {14.72} & {14.71} & {6.26} & {22.25} & {$-$6.75} & {2.12} & {42.12} & {39.10} & {52.19} & {41.66} & {17.08} \\
& & \textit{\tensft} & {\textit{64.82}} & {\textit{34.36}} & {\textit{54.56}} & {\textit{49.26}} & {\textit{16.04}} & {--} & {--} & {--} & {--} & {--} & {\textit{57.38}} & {\textit{45.27}} & {\textit{66.24}} & {\textit{49.18}} & {\textit{21.45}} \\
& & \sftbt & {\gaincell{\textbf{65.27}}{\up{3}}} & {\gaincell{\textbf{35.89}}{\up{11}}} & {\gaincell{\textbf{54.96}}{\up{2}}} & {\gaincell{\textbf{49.99}}{\up{7}}} & {\gaincell{\textbf{17.10}}{\up{16}}} & {\gaincell{\textbf{19.34}}{\up{31}}} & {\gaincell{\textbf{10.27}}{\up{64}}} & {\gaincell{\textbf{24.74}}{\up{11}}} & {\gaincell{\textbf{15.13}}{\textcolor{gaingreen}{+22\,pt}}} & {\gaincell{\textbf{4.40}}{\up{108}}} & {\gaincell{\textbf{62.92}}{\up{49}}} & {\gaincell{\textbf{47.17}}{\up{21}}} & {\gaincell{\textbf{71.23}}{\up{36}}} & {\gaincell{\textbf{51.79}}{\up{24}}} & {\gaincell{\textbf{23.32}}{\up{37}}} \\[0.5ex]
\cdashline{3-18}\addlinespace[0.6ex]
& \multirow{5}{*}{\rotatebox[origin=c]{90}{\textit{\xxtoen}}}
& Zero-Shot & {64.45} & {38.74} & {67.98} & {45.98} & {16.88} & {17.81} & {14.73} & {23.21} & {0.29} & {4.19} & {57.56} & {44.26} & {72.52} & {55.30} & {22.38} \\
& & \onesft & {67.78} & {41.65} & {\textbf{72.10}} & {\textbf{51.78}} & {19.83} & {19.37} & {21.92} & {29.20} & {25.98} & {7.97} & {60.05} & {46.13} & {76.08} & {60.37} & {25.23} \\
& & \textit{\tensft} & {\textit{70.48}} & {\textit{43.69}} & {\textit{73.49}} & {\textit{53.68}} & {\textit{21.68}} & {--} & {--} & {--} & {--} & {--} & {\textit{66.11}} & {\textit{49.13}} & {\textit{80.20}} & {\textit{63.59}} & {\textit{27.82}} \\
& & \sftbt & {\gaincell{\textbf{69.56}}{\up{3}}} & {\gaincell{\textbf{43.97}}{\up{6}}} & {\gaincell{71.77}{\textcolor{black!55}{0\%}}} & {\gaincell{51.77}{\textcolor{black!55}{0\%}}} & {\gaincell{\textbf{21.94}}{\up{11}}} & {\gaincell{\textbf{20.18}}{\up{4}}} & {\gaincell{\textbf{23.57}}{\up{8}}} & {\gaincell{\textbf{31.21}}{\up{7}}} & {\gaincell{\textbf{27.05}}{\up{4}}} & {\gaincell{\textbf{8.10}}{\up{2}}} & {\gaincell{\textbf{62.65}}{\up{4}}} & {\gaincell{\textbf{48.68}}{\up{6}}} & {\gaincell{\textbf{77.40}}{\up{2}}} & {\gaincell{\textbf{61.72}}{\up{2}}} & {\gaincell{\textbf{27.34}}{\up{8}}} \\[0.5ex]
\midrule
\multirow{10}{*}{\rotatebox[origin=c]{90}{\textbf{Qwen-2.5-3B}}}
& \multirow{5}{*}{\rotatebox[origin=c]{90}{\textit{\entoxx}}}
& Zero-Shot & {23.89} & {21.20} & {27.75} & {25.34} & {8.71} & {15.16} & {2.68} & {23.54} & {$-$18.27} & {0.52} & {17.97} & {20.44} & {26.45} & {3.22} & {7.35} \\
& & \onesft & {27.27} & {21.48} & {29.77} & {26.63} & {8.90} & {13.70} & {7.17} & {22.54} & {$-$5.17} & {2.63} & {21.14} & {26.99} & {30.73} & {19.50} & {10.67} \\
& & \textit{\tensft} & {\textit{44.72}} & {\textit{30.18}} & {\textit{40.26}} & {\textit{41.25}} & {\textit{13.82}} & {--} & {--} & {--} & {--} & {--} & {\textit{47.48}} & {\textit{43.59}} & {\textit{55.56}} & {\textit{45.90}} & {\textit{20.56}} \\
& & \sftbt & {\gaincell{\textbf{46.79}}{\up{72}}} & {\gaincell{\textbf{30.82}}{\up{43}}} & {\gaincell{\textbf{42.19}}{\up{42}}} & {\gaincell{\textbf{41.03}}{\up{54}}} & {\gaincell{\textbf{14.34}}{\up{61}}} & {\gaincell{\textbf{18.90}}{\up{38}}} & {\gaincell{\textbf{10.86}}{\up{51}}} & {\gaincell{\textbf{25.72}}{\up{14}}} & {\gaincell{\textbf{11.13}}{\textcolor{gaingreen}{+16\,pt}}} & {\gaincell{\textbf{4.21}}{\up{60}}} & {\gaincell{\textbf{46.62}}{\up{121}}} & {\gaincell{\textbf{43.16}}{\up{60}}} & {\gaincell{\textbf{55.61}}{\up{81}}} & {\gaincell{\textbf{46.53}}{\up{139}}} & {\gaincell{\textbf{20.19}}{\up{89}}} \\[0.5ex]
\cdashline{3-18}\addlinespace[0.6ex]
& \multirow{5}{*}{\rotatebox[origin=c]{90}{\textit{\xxtoen}}}
& Zero-Shot & {\textbf{59.35}} & {37.08} & {\textbf{62.51}} & {43.64} & {15.57} & {17.75} & {18.97} & {23.91} & {6.90} & {5.09} & {23.87} & {28.61} & {37.27} & {30.34} & {11.28} \\
& & \onesft & {56.00} & {35.95} & {60.59} & {42.41} & {16.00} & {19.75} & {22.64} & {30.06} & {\textbf{28.38}} & {\textbf{8.01}} & {32.94} & {34.77} & {49.06} & {44.45} & {17.05} \\
& & \textit{\tensft} & {\textit{60.08}} & {\textit{39.55}} & {\textit{63.47}} & {\textit{46.32}} & {\textit{18.37}} & {--} & {--} & {--} & {--} & {--} & {\textit{57.94}} & {\textit{46.45}} & {\textit{72.23}} & {\textit{59.00}} & {\textit{25.41}} \\
& & \sftbt & {\gaincell{56.48}{\up{1}}} & {\gaincell{\textbf{38.98}}{\up{8}}} & {\gaincell{61.48}{\up{1}}} & {\gaincell{\textbf{45.07}}{\up{6}}} & {\gaincell{\textbf{18.88}}{\up{18}}} & {\gaincell{\textbf{20.17}}{\up{2}}} & {\gaincell{\textbf{23.36}}{\up{3}}} & {\gaincell{\textbf{30.76}}{\up{2}}} & {\gaincell{26.29}{\down{7}}} & {\gaincell{7.71}{\down{4}}} & {\gaincell{\textbf{52.14}}{\up{58}}} & {\gaincell{\textbf{44.66}}{\up{28}}} & {\gaincell{\textbf{68.00}}{\up{39}}} & {\gaincell{\textbf{56.31}}{\up{27}}} & {\gaincell{\textbf{24.19}}{\up{42}}} \\[0.5ex]
\bottomrule
\end{tabular}%
}
\caption{Held-out results with all five metrics (2K test sentences per language, 486 for Amharic; 1K parallel pairs, 40K monolingual sentences, sampled pivots). Upper block: Odia, Assamese and Sanskrit; lower block: Hindi, Amharic and Swahili; \llamai{} above \qweni{} in each. Under each \BTtechnique{} score, \textcolor{gaingreen}{green}/\textcolor{lossred}{red} is the relative gain/loss over \onesft{}. Where either score is negative, the change is in absolute points (pt). \textbf{Bold}: best of Zero-Shot, \onesft{} and \BTtechnique{} per metric, model, language and direction. \tensft{} (italics) uses 10$\times$ more parallel data; it is shown for reference, is not ranked, and is not available for Amharic (--). X-XXL: XCOMET-XXL; X-XL: XCOMET-XL; BS: \bert{} F1.}
\label{tab:heldout-full}
\end{table*}

\section{Languages and Backbones: Full Flores-200 Results}
\label{app:flores}

This section gives the Flores-200 \citep{costajussa2022nllb} results (1,012 test sentences) for the systems of \S\ref{sec:crosslang}, all with sampled pivots, in the layout of Appendix~\ref{app:heldout} (Table~\ref{tab:flores-full}). Into the low-resource language, \BTtechnique{} improves over \onesft{} on all five metrics in ten of the twelve language and backbone combinations; the exceptions are \llama{} Hindi (all five metrics) and \llama{} Sanskrit (\xxl{} and \bert{}, by at most 0.40). Out of domain, it exceeds \tensft{} on \xxl{} in only five of the ten combinations that have a \tensft{} run, against eight of ten on the held-out test sets. Into English, \BTtechnique{} with sampled pivots stays below \onesft{} on \xxl{} in nine of the twelve combinations; only Amharic on both backbones and Swahili on \qwen{} improve, and they do so on all five metrics. \tensft{} has the higher \xxl{} into English in all ten. Beam-decoded pivots remove this gap on Odia and Swahili: on Flores-200 into English, \BTtechnique{} with beam pivots exceeds both \onesft{} and \tensft{} (Table~\ref{tab:recipes}). Finally, \llama{} zero-shot has the highest \xxl{} of all systems into Assamese and Sanskrit, although its \chrf{} into Assamese is below that of both trained systems.

\begin{table*}[tp]
\centering
\tiny
\setlength{\tabcolsep}{3.0pt}
\renewcommand{\arraystretch}{1.1}
\resizebox{\textwidth}{!}{%
\begin{tabular}{@{}c c l *{15}{S[table-format=2.2]}@{}}
\toprule
& &
& \multicolumn{5}{c}{Odia} & \multicolumn{5}{c}{Assamese} & \multicolumn{5}{c}{Sanskrit} \\
\cmidrule(lr){4-8}\cmidrule(lr){9-13}\cmidrule(lr){14-18}
& & \multicolumn{1}{l}{Method}
& {X-XXL} & {\chrf} & {X-XL} & {BS} & {BLEU} & {X-XXL} & {\chrf} & {X-XL} & {BS} & {BLEU} & {X-XXL} & {\chrf} & {X-XL} & {BS} & {BLEU} \\
\midrule
\multirow{10}{*}{\rotatebox[origin=c]{90}{\textbf{LLaMA-3.2-3B}}}
& \multirow{5}{*}{\rotatebox[origin=c]{90}{\textit{\entoxx}}}
& Zero-Shot & {20.81} & {4.95} & {23.61} & {$-$11.11} & {0.65} & {\textbf{41.97}} & {15.21} & {\textbf{34.81}} & {\textbf{30.13}} & {3.64} & {\textbf{35.43}} & {\textbf{15.82}} & {\textbf{31.56}} & {\textbf{17.87}} & {2.18} \\
& & \onesft & {30.60} & {18.03} & {29.41} & {17.80} & {5.84} & {26.62} & {17.07} & {24.66} & {18.89} & {5.23} & {19.93} & {10.47} & {22.26} & {$-$0.19} & {2.53} \\
& & \textit{\tensft} & {\textit{37.87}} & {\textit{21.73}} & {\textit{33.48}} & {\textit{26.17}} & {\textit{7.20}} & {\textit{32.35}} & {\textit{20.33}} & {\textit{27.70}} & {\textit{27.50}} & {\textit{6.62}} & {\textit{19.68}} & {\textit{9.71}} & {\textit{22.28}} & {\textit{$-$4.09}} & {\textit{2.17}} \\
& & \sftbt & {\gaincell{\textbf{39.04}}{\up{28}}} & {\gaincell{\textbf{22.25}}{\up{23}}} & {\gaincell{\textbf{35.91}}{\up{22}}} & {\gaincell{\textbf{26.61}}{\up{49}}} & {\gaincell{\textbf{7.70}}{\up{32}}} & {\gaincell{30.37}{\up{14}}} & {\gaincell{\textbf{20.54}}{\up{20}}} & {\gaincell{27.77}{\up{13}}} & {\gaincell{28.04}{\up{48}}} & {\gaincell{\textbf{6.88}}{\up{32}}} & {\gaincell{19.90}{\textcolor{black!55}{0\%}}} & {\gaincell{10.64}{\up{2}}} & {\gaincell{22.35}{\textcolor{black!55}{0\%}}} & {\gaincell{$-$0.59}{\textcolor{black!55}{0\,pt}}} & {\gaincell{\textbf{2.63}}{\up{4}}} \\[0.5ex]
\cdashline{3-18}\addlinespace[0.6ex]
& \multirow{5}{*}{\rotatebox[origin=c]{90}{\textit{\xxtoen}}}
& Zero-Shot & {16.96} & {19.41} & {24.76} & {4.18} & {5.54} & {\textbf{34.03}} & {\textbf{32.85}} & {39.51} & {33.41} & {\textbf{11.66}} & {\textbf{31.97}} & {\textbf{31.28}} & {\textbf{36.91}} & {\textbf{32.12}} & {\textbf{9.80}} \\
& & \onesft & {\textbf{40.94}} & {\textbf{37.41}} & {\textbf{54.67}} & {\textbf{46.19}} & {\textbf{14.27}} & {32.95} & {29.50} & {\textbf{41.87}} & {\textbf{40.51}} & {11.47} & {22.98} & {24.38} & {27.06} & {18.09} & {6.43} \\
& & \textit{\tensft} & {\textit{47.07}} & {\textit{39.26}} & {\textit{61.54}} & {\textit{49.77}} & {\textit{16.19}} & {\textit{38.20}} & {\textit{32.93}} & {\textit{47.02}} & {\textit{44.31}} & {\textit{12.71}} & {\textit{24.68}} & {\textit{24.66}} & {\textit{29.45}} & {\textit{19.20}} & {\textit{6.62}} \\
& & \sftbt & {\gaincell{38.87}{\down{5}}} & {\gaincell{34.92}{\down{7}}} & {\gaincell{54.56}{\textcolor{black!55}{0\%}}} & {\gaincell{45.29}{\down{2}}} & {\gaincell{13.79}{\down{3}}} & {\gaincell{30.66}{\down{7}}} & {\gaincell{29.97}{\up{2}}} & {\gaincell{39.65}{\down{5}}} & {\gaincell{36.19}{\down{11}}} & {\gaincell{11.13}{\down{3}}} & {\gaincell{21.53}{\down{6}}} & {\gaincell{22.43}{\down{8}}} & {\gaincell{23.42}{\down{13}}} & {\gaincell{4.37}{\down{76}}} & {\gaincell{4.57}{\down{29}}} \\[0.5ex]
\midrule
\multirow{10}{*}{\rotatebox[origin=c]{90}{\textbf{Qwen-2.5-3B}}}
& \multirow{5}{*}{\rotatebox[origin=c]{90}{\textit{\entoxx}}}
& Zero-Shot & {16.90} & {7.37} & {20.20} & {$-$0.32} & {1.60} & {20.10} & {11.29} & {21.06} & {6.74} & {2.33} & {\textbf{17.74}} & {\textbf{12.80}} & {\textbf{21.48}} & {\textbf{7.12}} & {1.44} \\
& & \onesft & {17.96} & {13.69} & {21.20} & {3.70} & {3.78} & {19.61} & {10.85} & {20.24} & {6.18} & {3.05} & {14.56} & {7.79} & {19.12} & {$-$16.99} & {1.16} \\
& & \textit{\tensft} & {\textit{33.47}} & {\textit{21.46}} & {\textit{31.88}} & {\textit{22.90}} & {\textit{7.13}} & {\textit{23.00}} & {\textit{14.64}} & {\textit{22.58}} & {\textit{15.32}} & {\textit{4.41}} & {\textit{18.36}} & {\textit{9.57}} & {\textit{21.16}} & {\textit{$-$3.43}} & {\textit{2.14}} \\
& & \sftbt & {\gaincell{\textbf{31.73}}{\up{77}}} & {\gaincell{\textbf{21.21}}{\up{55}}} & {\gaincell{\textbf{31.70}}{\up{50}}} & {\gaincell{\textbf{23.63}}{\up{539}}} & {\gaincell{\textbf{7.21}}{\up{91}}} & {\gaincell{\textbf{21.48}}{\up{10}}} & {\gaincell{\textbf{14.87}}{\up{37}}} & {\gaincell{\textbf{21.73}}{\up{7}}} & {\gaincell{\textbf{15.89}}{\up{157}}} & {\gaincell{\textbf{4.78}}{\up{57}}} & {\gaincell{16.99}{\up{17}}} & {\gaincell{9.43}{\up{21}}} & {\gaincell{20.31}{\up{6}}} & {\gaincell{$-$5.90}{\textcolor{gaingreen}{+11\,pt}}} & {\gaincell{\textbf{2.13}}{\up{84}}} \\[0.5ex]
\cdashline{3-18}\addlinespace[0.6ex]
& \multirow{5}{*}{\rotatebox[origin=c]{90}{\textit{\xxtoen}}}
& Zero-Shot & {24.65} & {29.77} & {34.49} & {30.94} & {9.07} & {\textbf{31.17}} & {\textbf{32.12}} & {\textbf{35.42}} & {\textbf{33.18}} & {\textbf{9.79}} & {\textbf{29.94}} & {\textbf{30.27}} & {\textbf{34.31}} & {\textbf{30.30}} & {\textbf{9.11}} \\
& & \onesft & {\textbf{30.99}} & {\textbf{30.90}} & {\textbf{43.55}} & {\textbf{37.34}} & {\textbf{11.02}} & {27.38} & {26.53} & {33.80} & {32.83} & {9.64} & {21.90} & {23.22} & {25.63} & {14.23} & {5.65} \\
& & \textit{\tensft} & {\textit{40.23}} & {\textit{35.51}} & {\textit{53.57}} & {\textit{44.30}} & {\textit{13.69}} & {\textit{30.97}} & {\textit{29.49}} & {\textit{38.65}} & {\textit{36.27}} & {\textit{10.49}} & {\textit{20.42}} & {\textit{22.65}} & {\textit{22.74}} & {\textit{9.82}} & {\textit{4.83}} \\
& & \sftbt & {\gaincell{24.55}{\down{21}}} & {\gaincell{28.66}{\down{7}}} & {\gaincell{37.28}{\down{14}}} & {\gaincell{33.36}{\down{11}}} & {\gaincell{10.67}{\down{3}}} & {\gaincell{24.38}{\down{11}}} & {\gaincell{24.42}{\down{8}}} & {\gaincell{30.54}{\down{10}}} & {\gaincell{27.29}{\down{17}}} & {\gaincell{8.86}{\down{8}}} & {\gaincell{18.64}{\down{15}}} & {\gaincell{17.85}{\down{23}}} & {\gaincell{19.81}{\down{23}}} & {\gaincell{$-$5.67}{\textcolor{lossred}{$-$20\,pt}}} & {\gaincell{3.88}{\down{31}}} \\[0.5ex]
\bottomrule
\end{tabular}%
}
\par\vspace{6pt}
\resizebox{\textwidth}{!}{%
\begin{tabular}{@{}c c l *{15}{S[table-format=2.2]}@{}}
\toprule
& &
& \multicolumn{5}{c}{Hindi} & \multicolumn{5}{c}{Amharic} & \multicolumn{5}{c}{Swahili} \\
\cmidrule(lr){4-8}\cmidrule(lr){9-13}\cmidrule(lr){14-18}
& & \multicolumn{1}{l}{Method}
& {X-XXL} & {\chrf} & {X-XL} & {BS} & {BLEU} & {X-XXL} & {\chrf} & {X-XL} & {BS} & {BLEU} & {X-XXL} & {\chrf} & {X-XL} & {BS} & {BLEU} \\
\midrule
\multirow{10}{*}{\rotatebox[origin=c]{90}{\textbf{LLaMA-3.2-3B}}}
& \multirow{5}{*}{\rotatebox[origin=c]{90}{\textit{\entoxx}}}
& Zero-Shot & {56.42} & {\textbf{41.88}} & {47.80} & {\textbf{56.91}} & {\textbf{18.69}} & {11.78} & {2.69} & {20.11} & {$-$22.23} & {0.07} & {35.05} & {38.86} & {39.94} & {43.63} & {14.05} \\
& & \onesft & {\textbf{63.87}} & {39.55} & {\textbf{52.67}} & {55.94} & {18.00} & {13.12} & {4.38} & {20.18} & {$-$11.54} & {1.29} & {35.41} & {37.97} & {40.21} & {41.76} & {13.61} \\
& & \textit{\tensft} & {\textit{63.00}} & {\textit{39.60}} & {\textit{51.78}} & {\textit{56.19}} & {\textit{17.88}} & {--} & {--} & {--} & {--} & {--} & {\textit{45.54}} & {\textit{40.56}} & {\textit{50.01}} & {\textit{46.87}} & {\textit{14.94}} \\
& & \sftbt & {\gaincell{62.25}{\down{3}}} & {\gaincell{39.49}{\textcolor{black!55}{0\%}}} & {\gaincell{50.20}{\down{5}}} & {\gaincell{55.02}{\down{2}}} & {\gaincell{17.37}{\down{3}}} & {\gaincell{\textbf{16.95}}{\up{29}}} & {\gaincell{\textbf{6.71}}{\up{53}}} & {\gaincell{\textbf{20.60}}{\up{2}}} & {\gaincell{\textbf{7.36}}{\textcolor{gaingreen}{+19\,pt}}} & {\gaincell{\textbf{2.90}}{\up{125}}} & {\gaincell{\textbf{53.18}}{\up{50}}} & {\gaincell{\textbf{43.17}}{\up{14}}} & {\gaincell{\textbf{56.80}}{\up{41}}} & {\gaincell{\textbf{50.67}}{\up{21}}} & {\gaincell{\textbf{17.49}}{\up{29}}} \\[0.5ex]
\cdashline{3-18}\addlinespace[0.6ex]
& \multirow{5}{*}{\rotatebox[origin=c]{90}{\textit{\xxtoen}}}
& Zero-Shot & {78.17} & {\textbf{52.64}} & {81.64} & {64.72} & {25.32} & {\textbf{15.94}} & {15.90} & {20.46} & {$-$1.38} & {4.78} & {\textbf{59.20}} & {\textbf{48.51}} & {70.09} & {60.34} & {\textbf{25.09}} \\
& & \onesft & {\textbf{78.38}} & {52.00} & {\textbf{83.71}} & {\textbf{66.27}} & {\textbf{25.71}} & {15.19} & {19.63} & {21.23} & {17.54} & {6.65} & {56.90} & {46.06} & {\textbf{71.21}} & {\textbf{60.55}} & {23.34} \\
& & \textit{\tensft} & {\textit{78.73}} & {\textit{51.79}} & {\textit{83.41}} & {\textit{66.02}} & {\textit{25.66}} & {--} & {--} & {--} & {--} & {--} & {\textit{58.46}} & {\textit{46.24}} & {\textit{72.20}} & {\textit{60.70}} & {\textit{23.86}} \\
& & \sftbt & {\gaincell{76.80}{\down{2}}} & {\gaincell{51.71}{\down{1}}} & {\gaincell{81.36}{\down{3}}} & {\gaincell{64.38}{\down{3}}} & {\gaincell{25.31}{\down{2}}} & {\gaincell{15.74}{\up{4}}} & {\gaincell{\textbf{20.55}}{\up{5}}} & {\gaincell{\textbf{22.95}}{\up{8}}} & {\gaincell{\textbf{17.96}}{\up{2}}} & {\gaincell{\textbf{6.66}}{\textcolor{black!55}{0\%}}} & {\gaincell{56.15}{\down{1}}} & {\gaincell{47.36}{\up{3}}} & {\gaincell{69.54}{\down{2}}} & {\gaincell{59.38}{\down{2}}} & {\gaincell{24.82}{\up{6}}} \\[0.5ex]
\midrule
\multirow{10}{*}{\rotatebox[origin=c]{90}{\textbf{Qwen-2.5-3B}}}
& \multirow{5}{*}{\rotatebox[origin=c]{90}{\textit{\entoxx}}}
& Zero-Shot & {17.78} & {24.28} & {21.46} & {28.89} & {8.64} & {14.03} & {1.97} & {\textbf{20.74}} & {$-$25.00} & {0.44} & {14.74} & {17.45} & {19.38} & {$-$1.29} & {4.56} \\
& & \onesft & {19.23} & {21.76} & {22.75} & {25.69} & {7.90} & {12.39} & {4.65} & {19.73} & {$-$13.80} & {1.45} & {14.50} & {21.82} & {19.85} & {12.01} & {6.26} \\
& & \textit{\tensft} & {\textit{31.25}} & {\textit{27.74}} & {\textit{28.45}} & {\textit{37.77}} & {\textit{10.82}} & {--} & {--} & {--} & {--} & {--} & {\textit{32.10}} & {\textit{36.74}} & {\textit{36.18}} & {\textit{40.51}} & {\textit{12.86}} \\
& & \sftbt & {\gaincell{\textbf{33.68}}{\up{75}}} & {\gaincell{\textbf{29.71}}{\up{37}}} & {\gaincell{\textbf{30.45}}{\up{34}}} & {\gaincell{\textbf{39.78}}{\up{55}}} & {\gaincell{\textbf{11.90}}{\up{51}}} & {\gaincell{\textbf{14.33}}{\up{16}}} & {\gaincell{\textbf{5.91}}{\up{27}}} & {\gaincell{20.53}{\up{4}}} & {\gaincell{\textbf{$-$5.28}}{\textcolor{gaingreen}{+9\,pt}}} & {\gaincell{\textbf{1.82}}{\up{26}}} & {\gaincell{\textbf{33.53}}{\up{131}}} & {\gaincell{\textbf{37.26}}{\up{71}}} & {\gaincell{\textbf{38.94}}{\up{96}}} & {\gaincell{\textbf{42.96}}{\up{258}}} & {\gaincell{\textbf{13.74}}{\up{119}}} \\[0.5ex]
\cdashline{3-18}\addlinespace[0.6ex]
& \multirow{5}{*}{\rotatebox[origin=c]{90}{\textit{\xxtoen}}}
& Zero-Shot & {\textbf{67.85}} & {\textbf{47.25}} & {\textbf{71.31}} & {\textbf{57.64}} & {\textbf{20.09}} & {16.10} & {20.27} & {20.82} & {7.97} & {5.38} & {19.78} & {27.23} & {29.89} & {27.43} & {9.78} \\
& & \onesft & {62.35} & {43.29} & {66.99} & {53.48} & {18.14} & {15.64} & {20.77} & {22.16} & {18.93} & {6.79} & {23.55} & {30.33} & {36.18} & {36.29} & {12.09} \\
& & \textit{\tensft} & {\textit{61.39}} & {\textit{43.62}} & {\textit{66.07}} & {\textit{53.39}} & {\textit{18.03}} & {--} & {--} & {--} & {--} & {--} & {\textit{42.81}} & {\textit{39.74}} & {\textit{57.22}} & {\textit{49.97}} & {\textit{18.19}} \\
& & \sftbt & {\gaincell{54.70}{\down{12}}} & {\gaincell{42.32}{\down{2}}} & {\gaincell{61.12}{\down{9}}} & {\gaincell{51.55}{\down{4}}} & {\gaincell{18.15}{\textcolor{black!55}{0\%}}} & {\gaincell{\textbf{16.24}}{\up{4}}} & {\gaincell{\textbf{21.97}}{\up{6}}} & {\gaincell{\textbf{23.83}}{\up{8}}} & {\gaincell{\textbf{19.40}}{\up{2}}} & {\gaincell{\textbf{6.84}}{\up{1}}} & {\gaincell{\textbf{38.25}}{\up{62}}} & {\gaincell{\textbf{39.62}}{\up{31}}} & {\gaincell{\textbf{54.34}}{\up{50}}} & {\gaincell{\textbf{48.71}}{\up{34}}} & {\gaincell{\textbf{18.36}}{\up{52}}} \\[0.5ex]
\bottomrule
\end{tabular}%
}
\caption{Flores-200 results with all five metrics (1,012 test sentences; 1K parallel pairs, 40K monolingual sentences, sampled pivots). Upper block: Odia, Assamese and Sanskrit; lower block: Hindi, Amharic and Swahili; \llamai{} above \qweni{} in each. Under each \BTtechnique{} score, \textcolor{gaingreen}{green}/\textcolor{lossred}{red} is the relative gain/loss over \onesft{}. Where either score is negative, the change is in absolute points (pt). \textbf{Bold}: best of Zero-Shot, \onesft{} and \BTtechnique{} per metric, model, language and direction. \tensft{} (italics) uses 10$\times$ more parallel data; it is shown for reference, is not ranked, and is not available for Amharic (--). X-XXL: XCOMET-XXL; X-XL: XCOMET-XL; BS: \bert{} F1.}
\label{tab:flores-full}
\end{table*}

\section{Seed Robustness and Significance (Odia)}
\label{app:seeds}

To check that the Odia results are not artifacts of one random initialisation, we train \onesft, \BTtechnique{} and \tensft~three times each: the original run and two more runs with seeds 123 and 456. Table~\ref{tab:seed-all} reports the mean $\pm$ standard deviation over the three seeds on the held-out (2K) set. Standard deviations are at most 1.85 points, which is larger than the gap between \BTtechnique{} and \tensft{} on \xxl{} into Odia (66.52 against 66.42).

We also test whether \BTtechnique{} and \tensft~differ per sentence. For each system, we first average each test sentence's score over the three seeds, so that the seeds are not counted as extra test sentences, and then apply a two-sided paired Wilcoxon signed-rank test \citep{wilcoxon1945individual} to the 2K sentence pairs. We correct the ten $p$-values (five metrics, two directions) with the Holm procedure \citep{holm1979simple}. The last row of each block in Table~\ref{tab:seed-all} gives the rank-biserial effect size, where a positive value means \BTtechnique{} is ahead. Into English, \tensft~is significantly ahead on every metric ($r$ from $-0.20$ to $-0.36$). Into Odia, \BTtechnique{} is significantly ahead on \xl, \bert{} and BLEU, and the two are not significantly different on \xxl~($p{=}0.67$) or \chrf~(Holm-adjusted $p{=}0.12$). A paired bootstrap over sentences \citep{koehn2004statistical} does favour \BTtechnique{} on mean \chrf~($+0.72$ points, 95\% CI $0.26$ to $1.18$), so the \chrf~result depends on the test. These tests use sampled pivots and treat the three seeds as fixed.

\begin{table*}[tp]
\centering
\setlength{\tabcolsep}{5pt}
\begin{tabular}{@{}l ccccc@{}}
\toprule
Configuration & \xxl~& \xl~& \chrf~& \bert~F1 & BLEU \\
\midrule
\multicolumn{6}{@{}l}{\textbf{Held-out (2K)}} \\
\addlinespace[0.2em]
\multicolumn{6}{@{}l}{\quad\textit{\entoxx}} \\
\addlinespace[0.1em]
\onesft~& 42.94 $\pm$ 0.79 & 43.94 $\pm$ 0.56 & 25.70 $\pm$ 0.51 & 30.08 $\pm$ 1.06 & 10.93 $\pm$ 0.22 \\
\tensft~& \underline{66.42 $\pm$ 1.85} & \underline{62.88 $\pm$ 1.59} & \underline{41.16 $\pm$ 0.21} & \underline{51.27 $\pm$ 0.85} & \underline{20.07 $\pm$ 0.07} \\
\sftbt~& \textbf{66.52 $\pm$ 0.68} & \textbf{64.52 $\pm$ 0.29} & \textbf{41.88 $\pm$ 0.80} & \textbf{52.71 $\pm$ 0.86} & \textbf{21.12 $\pm$ 0.68} \\
\emph{Significance} & $+0.01$\,n.s. & $+0.10^{***}$ & $+0.05$\,n.s. & $+0.12^{***}$ & $+0.12^{***}$ \\
\addlinespace[0.35em]
\multicolumn{6}{@{}l}{\quad\textit{\xxtoen}} \\
\addlinespace[0.1em]
\onesft~& 45.32 $\pm$ 0.09 & 63.99 $\pm$ 0.47 & 46.50 $\pm$ 0.09 & 56.53 $\pm$ 0.38 & 24.23 $\pm$ 0.10 \\
\tensft~& \textbf{60.65 $\pm$ 0.74} & \textbf{77.41 $\pm$ 0.76} & \textbf{58.88 $\pm$ 0.73} & \textbf{68.62 $\pm$ 0.76} & \textbf{36.97 $\pm$ 0.92} \\
\sftbt~& \underline{55.32 $\pm$ 1.36} & \underline{73.07 $\pm$ 0.63} & \underline{56.04 $\pm$ 1.24} & \underline{65.71 $\pm$ 0.63} & \underline{34.44 $\pm$ 0.93} \\
\emph{Significance} & $-0.36^{***}$ & $-0.32^{***}$ & $-0.27^{***}$ & $-0.30^{***}$ & $-0.20^{***}$ \\
\bottomrule
\end{tabular}
\caption{Seed robustness and per-sentence significance (Odia, \llamai): mean $\pm$ standard deviation over three seeds on the held-out (2K) set, \entoxx~above \xxtoen. Bold and underline mark the best and second-best mean per metric and direction. The \emph{Significance} row gives the rank-biserial effect size of a paired Wilcoxon test between \sftbt~and \tensft~on per-sentence scores averaged over the three seeds (2K pairs). A positive sign means \sftbt~is ahead, and a negative sign means it is behind. Stars use Holm-adjusted $p$-values over the ten tests: $^{*}p{<}0.05$, $^{**}p{<}0.01$, $^{***}p{<}0.001$, and n.s.\ = not significant.}
\label{tab:seed-all}
\end{table*}

\section{Qualitative Examples}
\label{app:examples}

\begin{figure*}[tp]
\centering
\includegraphics[width=\textwidth]{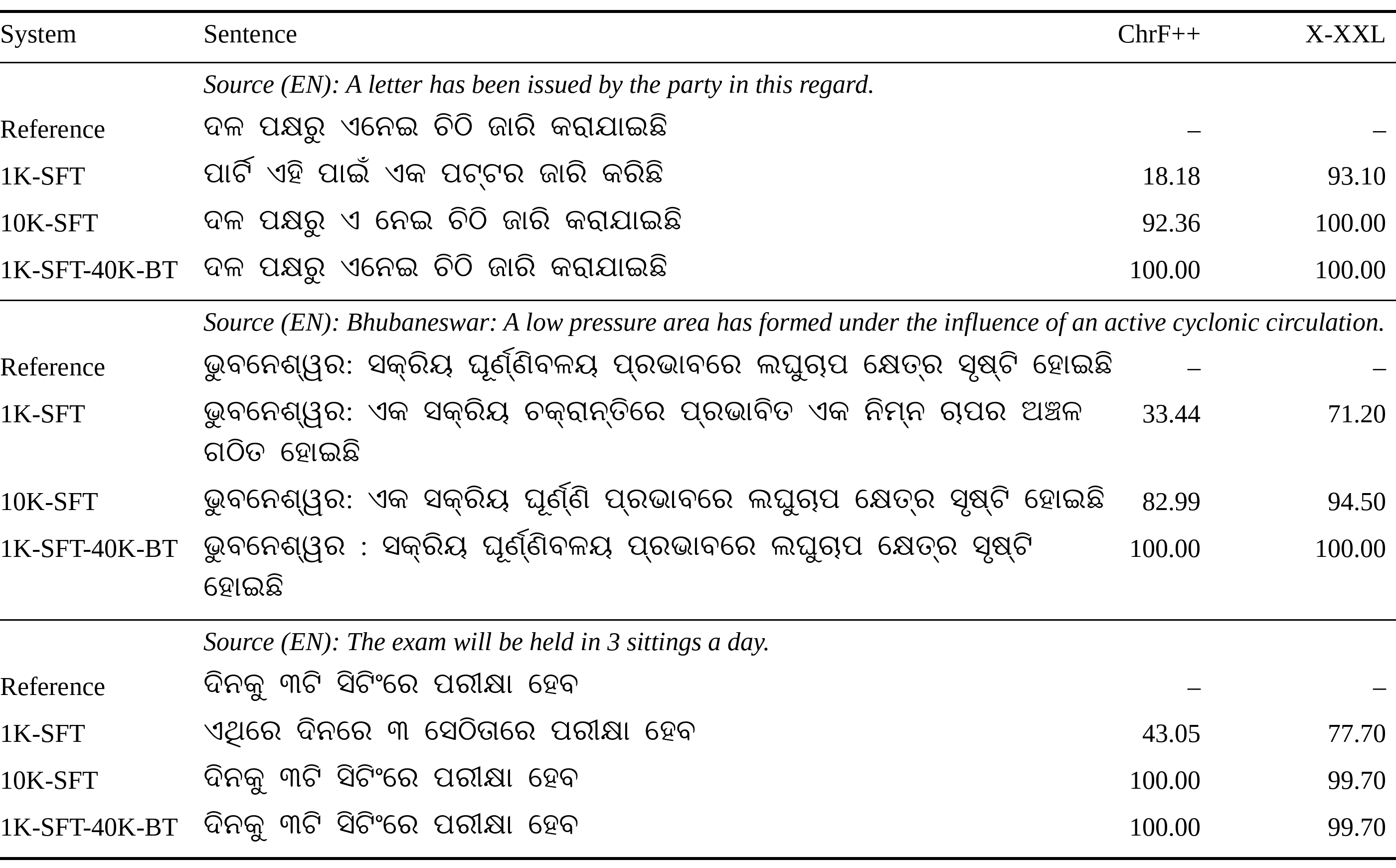}
\caption{EN$\rightarrow$OR (Odia) examples where \sftbt~and \tensft~both
improve over \onesft~on \chrf~\emph{and} \xxl~(X-XXL), \llamai. In the first
example the \onesft~output scores highly on \xxl~despite a low \chrf, one of
the cases motivating our use of a model-independent complement to the learned
metric.}
\label{fig:examples-or}

\bigskip
\includegraphics[width=\textwidth]{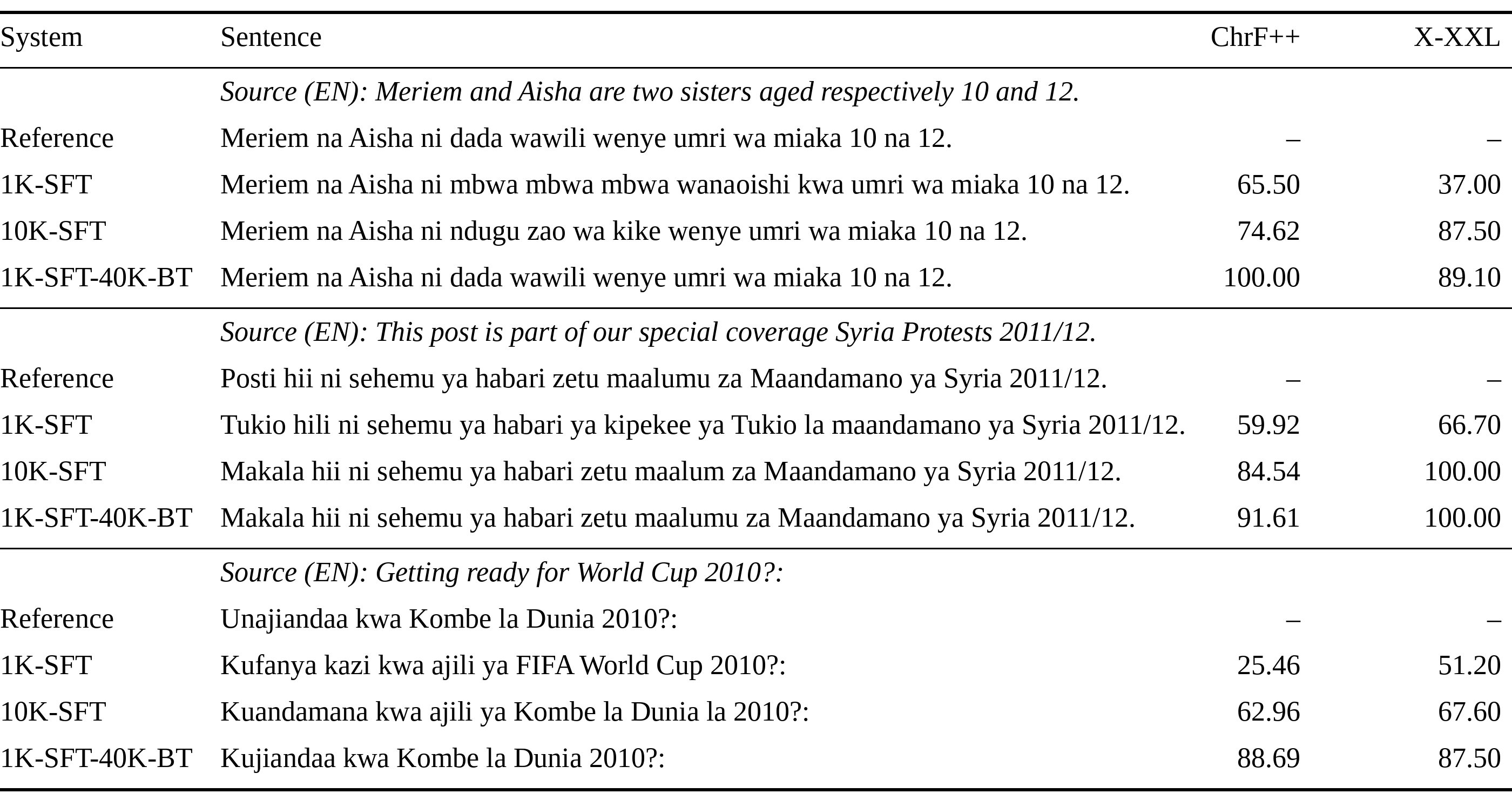}
\caption{EN$\rightarrow$SW (Swahili) examples where \sftbt~and \tensft~both
improve over \onesft~on \chrf~\emph{and} \xxl~(X-XXL), \llamai.}
\label{fig:examples-sw}
\end{figure*}

Figures~\ref{fig:examples-or} and~\ref{fig:examples-sw} show \entoxx~outputs on the held-out test set for the three configurations of Table~\ref{tab:main-crosslang}, alongside the source and reference, for Odia and Swahili. The examples are selected among sentences on which both \tensft~and \BTtechnique{} improve over \onesft~on \chrf~and \xxl, so they illustrate these gains rather than the average case. The \onesft~seed model is often fluent but inadequate or partly code-mixed. Both \tensft~and \BTtechnique{} recover the source meaning, and \BTtechnique{} is comparable to the model trained on ten times more gold data. 

\section{Human Evaluation Protocol}
\label{app:human}

\paragraph{Annotators.}
Each language was annotated by three research scholars. The Odia annotators are native speakers, and the Sanskrit annotators are language experts with ongoing research in Sanskrit computational linguistics. None of the six annotators is an author of this paper or a member of the authors' research group. Participation was voluntary and unpaid, and no further demographic information was collected.

\paragraph{Task and instructions.}
Each annotator ranked the \entoxx{} translations of 110 source sentences of their language, sampled at random from the held-out test set. On a web page, they saw the English source and the outputs of \onesft{}, \tensft{} and \BTtechnique{} (\llamai, sampled pivots, the checkpoints of Table~\ref{tab:main-crosslang}) in random order and without system names. Annotators did not see the reference translation. The instructions read as follows (a final note on reopening the page if it failed to load is omitted):
\begin{quote}
\small
You'll see an English sentence and three candidate translations below it. The job is to rank the translations from best to worst. \emph{Rank by dragging.} Drag each translation card up or down to set its priority --- best on top (\#1), worst at the bottom (\#3). \emph{Already in the right order?} Leave it as is. Ranking saves automatically when you move to the next sentence. \emph{Green tick = done.} A green tick next to a number on the left panel means that sentence is marked complete. You can click any sentence anytime to revisit it and change the order. Move between sentences using Previous / Next. Your progress is shown at the top.
\end{quote}
Every annotator ranked all 110 sentences of their language, each as a strict order.

\paragraph{Consent, data handling and ethics review:}
Before starting, annotators were told the purpose of the study and that their rankings would be reported only in aggregate in a publication, and all agreed verbally to take part. The task involved only ranking translations of sentences from public corpora. The interface recorded each annotator's login email only to save progress. We report annotators anonymously (A1 to A3 in Table~\ref{tab:human_eval}), and responses from accounts the authors used to test the interface were excluded. The study was not reviewed by an institutional ethics board. It posed minimal risk and involved no personal information beyond the login email.

\paragraph{Results and agreement:}
Table~\ref{tab:human_eval} reports the mean rank (1 = best, 3 = worst) per annotator and over all annotators. The annotators agree well above chance: the mean per-sentence Kendall's $W$ is 0.66 for Odia and 0.58 for Sanskrit (0.33 for random rankings), two annotators order a pair of systems the same way in 72.5\% and 66.7\% of cases (50\% by chance), and Fleiss' $\kappa$ on the best-ranked system is 0.37 and 0.25.

\section{Conditional Translation Perplexity}
\label{app:ppl}

As an intrinsic complement to the quality metrics, we report \emph{conditional} translation perplexity on the held-out (2K) test set. It is the perplexity of the reference translation given the source prompt (lower is better). Unlike unconditional language-model perplexity, it measures how well the model predicts the reference given the source, so it reflects translation modelling rather than target-language fluency alone.

Perplexity is aggregated at the corpus level as total negative log-likelihood divided by the total number of scored target tokens. Non-Latin scripts need many more target tokens than English under both tokenizers, so \entoxx~has more scored tokens than \xxtoen, from about $1.2\times$ (Sanskrit, \llama) to $10.2\times$ (Odia, \llama) (Table~\ref{tab:token_counts}). Absolute perplexities should therefore be compared only within a direction.

The perplexity trends broadly mirror translation quality (Table~\ref{tab:ppl}). Training lowers perplexity from the zero-shot model in every language, direction and backbone. \BTtechnique{} has the lowest perplexity into English in all twelve language and backbone combinations, and the lowest or second-lowest into the low-resource language in eleven of twelve. Amharic is the main exception. Despite better translation quality (Table~\ref{tab:main-crosslang}), \BTtechnique{}'s \entoxx~perplexity is higher than that of \onesft{} (28.60 against 18.37 with \llama, 82.09 against 18.71 with \qwen), and Swahili with \llama{} shows the same pattern mildly (5.31 against 4.98). Conditional perplexity therefore need not track reference-based quality.

\begin{table*}[tp]
\centering
{\small
\setlength{\tabcolsep}{4pt}
\begin{tabular}{@{}lrrrr@{}}
\toprule
 & \multicolumn{2}{c}{\textbf{LLaMA}} & \multicolumn{2}{c}{\textbf{Qwen}} \\
\cmidrule(lr){2-3}\cmidrule(lr){4-5}
Language & {\xxtoen} & {\entoxx} & {\xxtoen} & {\entoxx} \\
\midrule
Odia      & 41,245 & 421,964 & 45,190 & 347,728 \\
Assamese  & 31,837 & 160,459 & 32,288 & 144,232 \\
Sanskrit  & 87,583 & 107,315 & 87,012 & 177,406 \\
Hindi     & 52,087 & 117,775 & 52,706 & 205,350 \\
Amharic   & 11,449 &  88,049 & 13,050 &  48,881 \\
Swahili   & 49,476 &  94,210 & 50,376 &  95,846 \\
\bottomrule
\end{tabular}
}
\caption{Target tokens scored per direction over the held-out test set (2K sentences, 486 for Amharic). The \entoxx~count is larger for every language and backbone. The ratio ranges from $1.2\times$ (Sanskrit, \llama) to $10.2\times$ (Odia, \llama).}
\label{tab:token_counts}
\end{table*}

\begin{table*}[tp]
\centering
{\small
\setlength{\tabcolsep}{3pt}
\begin{tabular}{@{}lcccccc@{}}
\toprule
Method & Odia & Assamese & Sanskrit & Hindi & Amharic & Swahili \\
\midrule
\multicolumn{7}{@{}l}{\textbf{\llamai}} \\
\addlinespace[0.5pt]
\multicolumn{7}{@{}l}{\hspace{0.3em}\textit{\entoxx}} \\
Zero-Shot & 12.75 & 29.09 & 32.82 & 9.23 & 44.04 & 6.99 \\
\onesft~& 3.83 & 13.17 & 27.73 & 5.59 & \textbf{18.37} & \underline{4.98} \\
\tensft~& \underline{3.56} & \underline{12.16} & \textbf{13.21} & \textbf{5.00} & & \textbf{4.81} \\
\sftbt~& \textbf{2.68} & \textbf{9.93} & \underline{14.12} & \underline{5.19} & \underline{28.60} & 5.31 \\
\addlinespace[2pt]
\multicolumn{7}{@{}l}{\hspace{0.3em}\textit{\xxtoen}} \\
Zero-Shot & 2.48 & 4.77 & 26.96 & 4.83 & 4.38 & 6.10 \\
\onesft~& 1.39 & 2.36 & 17.04 & 3.72 & \underline{2.45} & 5.20 \\
\tensft~& \underline{1.32} & \underline{1.91} & \underline{8.23} & \underline{3.27} & & \underline{3.82} \\
\sftbt~& \textbf{1.21} & \textbf{1.81} & \textbf{7.68} & \textbf{3.15} & \textbf{2.23} & \textbf{3.38} \\
\midrule
\multicolumn{7}{@{}l}{\textbf{Qwen-2.5-3B-Instruct}} \\
\addlinespace[0.5pt]
\multicolumn{7}{@{}l}{\hspace{0.3em}\textit{\entoxx}} \\
Zero-Shot & 35.53 & 311.99 & 111.04 & 48.65 & 151.08 & 51.33 \\
\onesft~& 6.52 & \underline{16.81} & \underline{20.04} & \underline{10.49} & \textbf{18.71} & 10.04 \\
\tensft~& \underline{3.86} & 27.16 & 25.52 & 13.28 & & \textbf{6.46} \\
\sftbt~& \textbf{3.66} & \textbf{11.82} & \textbf{16.59} & \textbf{7.98} & \underline{82.09} & \underline{6.55} \\
\addlinespace[2pt]
\multicolumn{7}{@{}l}{\hspace{0.3em}\textit{\xxtoen}} \\
Zero-Shot & 3.15 & 8.19 & 11.54 & 4.39 & 17.57 & 59.28 \\
\onesft~& 1.74 & 3.08 & 5.23 & 2.81 & \underline{5.69} & 14.68 \\
\tensft~& \underline{1.35} & \underline{2.32} & \underline{4.38} & \underline{2.35} & & \underline{5.07} \\
\sftbt~& \textbf{1.26} & \textbf{2.02} & \textbf{3.64} & \textbf{2.09} & \textbf{4.37} & \textbf{4.09} \\
\bottomrule
\end{tabular}
}
\caption{Conditional translation perplexity (lower is better) on the held-out (2K) test set (\llamai~top, \qweni~bottom). \entoxx~columns report target-language perplexity, and \xxtoen~columns report English perplexity. Bold and underline denote the best and second-best per column within each model and direction. Blank cells were not evaluated (no \tensft~for Amharic).}
\label{tab:ppl}
\end{table*}

\end{document}